\documentclass{article} 
\usepackage{iclr2027_conference,times}
\iclrfinalcopy

\usepackage{amsmath,amsfonts,bm}

\def\eqref#1{equation~\ref{#1}}

\def\1{\bm{1}}

\DeclareMathAlphabet{\mathsfit}{\encodingdefault}{\sfdefault}{m}{sl}
\SetMathAlphabet{\mathsfit}{bold}{\encodingdefault}{\sfdefault}{bx}{n}

\usepackage{amsmath}
\usepackage{amssymb}

\usepackage{graphicx}
\usepackage{subcaption}
\usepackage{wrapfig}

\usepackage{booktabs}
\usepackage[table]{xcolor}
\usepackage{tcolorbox}
\tcbuselibrary{theorems}
\usepackage{empheq}
\usepackage{multirow}

\usepackage{enumitem}

\usepackage{microtype}
\usepackage{lineno}

\usepackage{hyperref}
\usepackage{url}

\definecolor{darkblue}{rgb}{0,0,0.5}
\hypersetup{
    colorlinks=true,
    citecolor=darkblue,
    linkcolor=darkblue,
    urlcolor=darkblue
}

\definecolor{groupgrey}{gray}{0.92}
\definecolor{childgrey}{gray}{0.97}
\definecolor{good}{rgb}{0.0,0.5,0.0}
\definecolor{cuteorange}{HTML}{F59E42}

\tcbset{highlight math style={colback=darkblue!3, colframe=darkblue!45,
    boxrule=0.5pt, arc=6pt, left=10pt, right=10pt, top=3pt, bottom=3pt}}

\newcommand{\paramdn}[1]{\textcolor{good}{$\downarrow$\,#1}}

\title{LayerRoPE: Dynamic Depth-wise Magnitude \& Angular Superposition}

\author{Shikhar Srivastava\thanks{Correspondence to \href{mailto:ssrivas9@ur.rochester.edu}
{\texttt{ssrivas9@ur.rochester.edu}}.\\
Project webpage:
\url{www.cs.rochester.edu/u/ssrivas9/layer-rope}.
}\;\;\& Christopher Kanan \\
University of Rochester
}

\begin{document}

\maketitle
\lhead{Preprint. Under review.}

\begin{abstract}
As data propagates through a Transformer, the norm of its hidden states grows by orders of magnitude with depth, a phenomenon framed as `curse of depth' and nearly universally treated as a pathology to be suppressed. We take the opposite view. Across 16 pre-trained LLMs from 9 families, spanning dense, mixture-of-experts and hybrid architectures and Pre-, Peri- and Post-Norm designs, we find that this growth reflects an emergent \textit{depth-positional encoding}, carried by the only learned per-layer gain on the residual stream, the normalization weight $\gamma$: with depth, $\gamma$ grows in magnitude and rotates in direction, jointly encoding the layer index. We make this depth-conditioned encoding explicit with \textbf{LayerRoPE}, an implicit analog of RoPE along the depth axis, which replaces all layerwise $\gamma$ vectors with a single shared vector, depth-conditioned scalars, at a net reduction in parameters and $<\!0.02\%$ change in FLOPs. Across a model ladder scaled upto $100$B+ tokens, LayerRoPE consistently outperforms Pre-, Post- and Peri-Norm, Layer-Norm Scaling, reaching Pre-Norm's 1.3B loss with $3.4\times$ less compute;  LayerRoPE is the only approach that shows strong convergence and improves near monotonically as depth scales to 512 layers. It improves learning-rate sensitivity by $3$-$10\times$, and transfers naively to and consistently improves looped latent models and Vision Transformers. Inspecting its learned schedule inverts the prevailing premise: LayerRoPE does not shrink the residual stream but \emph{widens} it, damping what each block reads while amplifying what it writes. Depth stability, our results suggest, calls not for suppressing the residual stream, but for depth-conditioned regulation of the computational blocks it feeds.
\end{abstract}

\begin{figure}[h]
  \centering
  \vspace{-15pt}
  \captionsetup[sub]{font=footnotesize, skip=2pt}
  \begin{subfigure}[b]{0.41\textwidth}
    \centering
    \includegraphics[width=\textwidth]{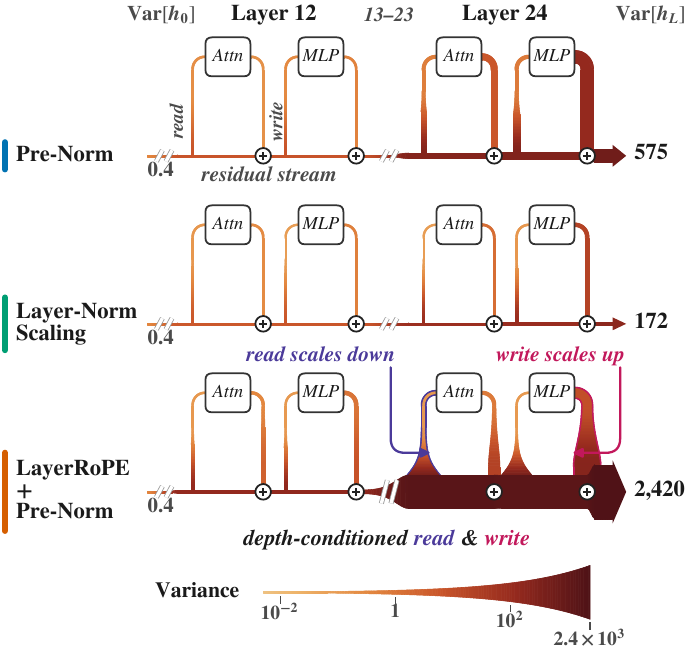}
    \caption{Illustration: LayerRoPE's regulation of computational block with \textit{depth-conditioned} I/O, while \textit{maximizing residual variance}.}
    \label{fig:residual-pipe}
  \end{subfigure}\hfill
  \begin{subfigure}[b]{0.575\textwidth}
    \centering
    \includegraphics[width=\textwidth]{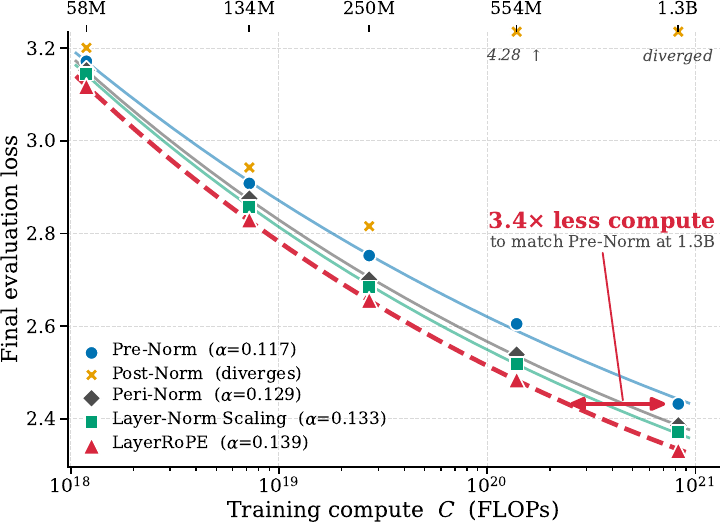}
    \caption{Compute-scaling curves on a \(4\times\) Chinchilla model-ladder}
    \label{fig:compute-scaling}
  \end{subfigure}
  \vspace{-5pt}
  \caption{\textbf{LayerRoPE offers a contrary premise:} \textit{Maximizing residual variance with depth-conditioned regulation of computational blocks}, allows better convergence, deeper networks, wider learning basins, efficiently ($<\!0.02\%$ FLOP increase), and across architectures and modalities.}
  \label{fig:highlight}

\end{figure}

\vspace{-10pt}
\section{Introduction}

Transformers underpin modern AI, from large language models (LLMs) to Vision Transformers (ViTs). All such systems share a persistent pathology: as the input propagates through the network, the $\ell_2$ norms of hidden states grow rapidly with depth, often by several orders of magnitude between the input embeddings and the final layers. Recent studies have characterized various distinct manifestations of this growth.
\citet{sun2024massive} identify ``massive activations" whose magnitudes exceed the mean by more than $3000$ times; \citet{darcet2024vision} observe high-norm ``register'' tokens that emerge in ViTs to absorb global context; and \citet{sun2025curse} frame the broader phenomenon of activation variance growth in Pre-Norm transformers as a ``curse of depth'' that degrades gradient flow and layer contributions in deeper LLMs.

The prevailing beliefs treat this growth as something to suppress or re-engineer away, along two lines. \textbf{One}, damps the residual stream directly: Layer-Norm Scaling~\citep{sun2025curse}, for instance, scales each block's input, the output of its layer normalization, by $1/\sqrt{\ell+1}$, and Peri-Norm~\citep{kim2025peri} re-normalizes each sub-layer's output before it re-enters the stream. \textbf{Second}, to rewrite the residual pathway itself: Attention Residuals~\citep{team2026attention} show that the unit-weight residual is equivalent to a low-rank linear attention over depth, whose uniform accumulation forces hidden-state norms to grow as $O(L)$, and replace it with a learned aggregation that holds magnitudes bounded; Hyper-Connections \citep{zhu2025hyper} and Manifold-Constrained Hyper-Connections \citep{xie2025mhc} instead widen the single stream into several parallel streams, constraining their mixing to preserve signal magnitude. A detailed discussion of related work is provided in Section~\ref{sec:related_work}. Nonetheless, across both lines, the unifying premise is that activation growth is a defect to be removed. 

\textit{We take the opposite view: } The residual growth is not in itself a failure mode, and the goal must not be to shrink the residual pipe that flows through the network. Rather, to regulate the computational blocks it feeds into, allowing us to further widen the residual pipe flowing through the network. We show through broad experimentation that this view leads to better depth scaling, than residual damping or unconstrained computation. In order to regulate computational blocks to read/write into a wide residual stream, it is first necessary to ask what the growth \emph{encodes}. We find, interestingly, that the systemic, per-layer growth of activation norms is indicative of an emergent \textit{depth-positional encoding} that the network learns to express through the only learned per-layer gain directly on the residual stream: the RMSNorm weight $\gamma$. Through analysis of a large breadth of LLMs, we show that the MLP block's $\gamma$ vectors grow and rotate {conditioned variously} on the network depth. Magnitude and rotation, we hypothesize, therefore, are conditioned on the residual stream's depth, and therefore jointly encode a form of \textit{layer-index encoding}: an implicit analog of RoPE~\citep{su2024roformer} along the depth axis rather than along sequence position.

We make this layer-index encoding explicit. \textbf{LayerRoPE} replaces the per-layer gains $\gamma_\ell$ with one shared $\gamma$ per site, rotated and scaled by a few depth-conditioned scalars. It is a drop-in addition to any computational block, at net fewer parameters and a $<\!0.02\%$ change in FLOPs. Our contributions are:
\begin{itemize}[leftmargin=*, itemsep=1pt, topsep=2pt]
  \item \textbf{An emergent depth code.} Across 16 LLMs with varying families and architectures, the norm weights $\gamma$ show depth-conditioned magnitude and rotation, with ${\sim}99\%$ of its dimensions following a shared layer-wise trajectory (Section~\ref{sec:norm_depth_code}).
  \item \textbf{Better scaling.} On a 58M-1.3B ladder trained at $4\times$ the Chinchilla budget (up to 107B tokens), LayerRoPE attains the lowest loss at every scale, reaches Pre-Norm's 1.3B loss with $3.4\times$ less compute, and achieves the best average zero-shot accuracy (Section~\ref{sec:scaling-ladder}).
  \item \textbf{Deeper, more stable networks.} It is the only method whose loss improves monotonically with strong convergence, up to 512 layers (Section~\ref{sec:abu_scaling}), and it cuts learning-rate sensitivity by $3$-$10\times$ while training stably at rates up to $8\times$ higher than its baseline tolerates (Section~\ref{sec:lr-robustness}).
  \item \textbf{Beyond language models.} Without tuning, it improves looped latent models on nearly every metric at every scale, and Vision Transformers by up to 0.9 points of top-1 accuracy (Section~\ref{sec:vit-looped}).
  \item \textbf{A wider stream, not a narrower one.} LayerRoPE's learned depth-conditioned schedule scales down reads and amplifies writes into the stream, widening the residual stream between $4$-$100\times$ its backbone's variance at 1.3B (Section~\ref{sec:abu_learned_dynamics}).
\end{itemize}

\begin{figure}[t]
    \centering
    \begin{subfigure}[b]{0.55\textwidth}
        \centering
        \includegraphics[width=\linewidth]{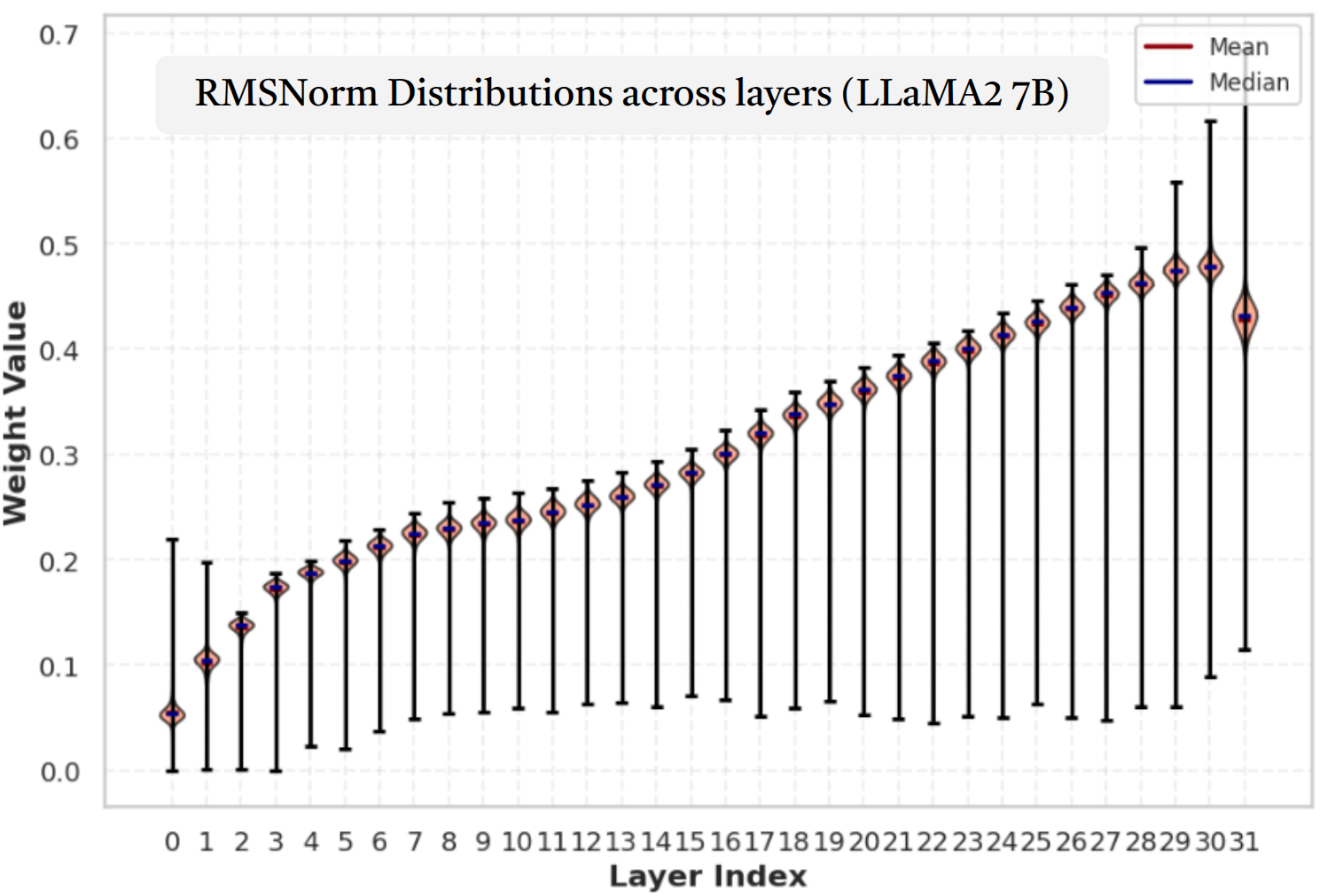}
        \caption{Distribution of MLP block RMSNorm $\gamma$ values. Means and upper percentiles grow nearly monotonically with depth.}
        \label{fig:abu_rmsnorm_dist}
    \end{subfigure}
    \hfill
    \begin{subfigure}[b]{0.42\textwidth}
        \centering
        \includegraphics[width=\linewidth]{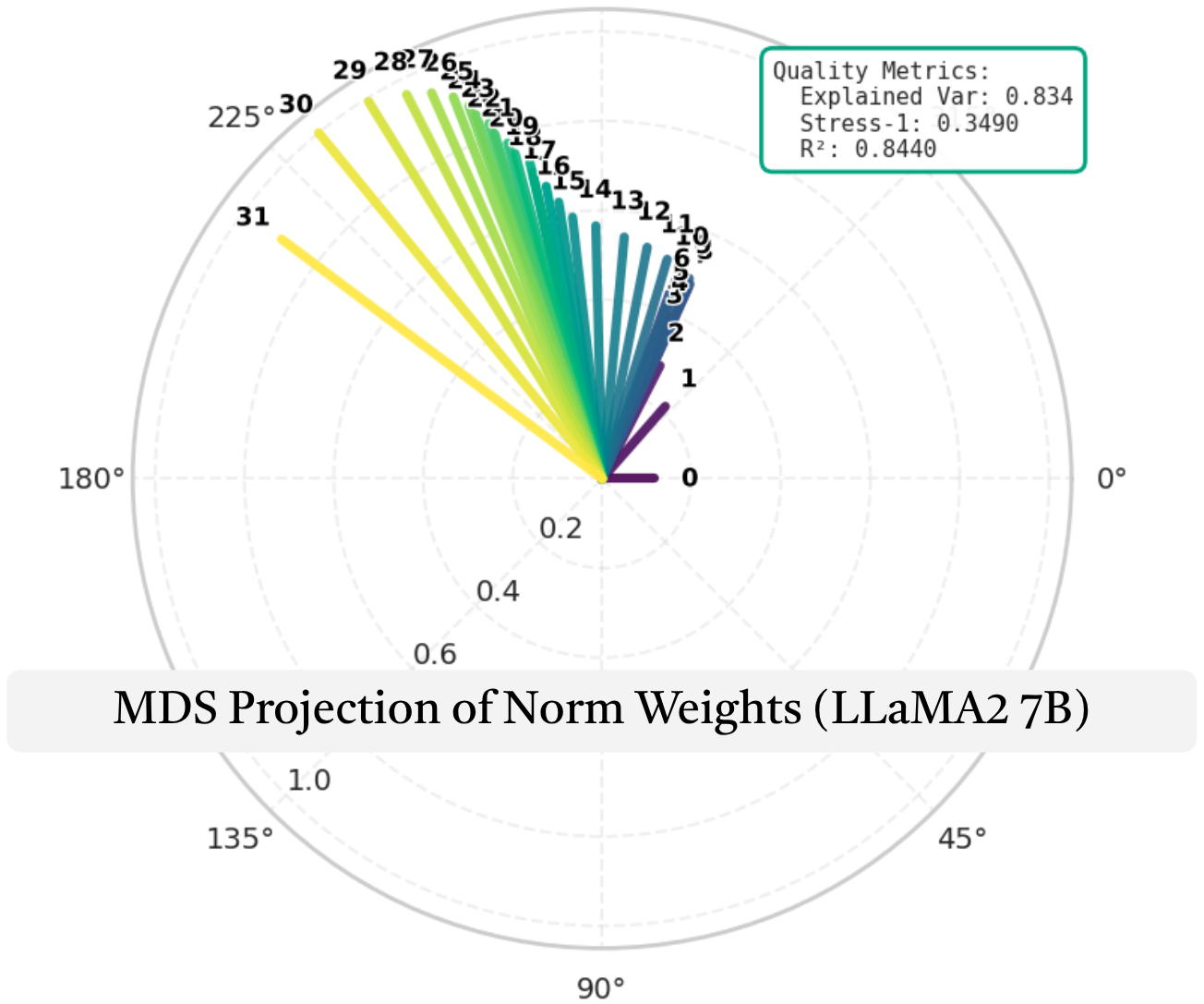}
        \caption{MDS projection of the same per-layer $\gamma$ vectors. Successive layers trace a near-monotone arc spanning $\sim\!143^\circ$.}
        \label{fig:abu_mds_projection}
    \end{subfigure}
    \vspace{-5pt}
    \caption{Norm weights encode depth in LLAMA-like LLMs (LLaMA-2 7B shown here). Magnitude (left) and angular position (right) jointly form a depth-positional code along the residual stream.}
    \vspace{-10pt}
    \label{fig:abu_norm_evidence}
\end{figure}

\section{Norm Weights Encode Depth: A LayerRoPE View\label{sec:norm_depth_code}}

A Transformer exposes exactly one learned gain per layer directly on the residual path: the RMSNorm weights $\gamma$. It is therefore the most straightforward site for any depth-conditional code the network may wish to learn. Analyzing pre-trained LLMs, we find that the MLP Norm weights $\gamma$ carry such a code, expressed jointly through two depth-varying quantities: its \emph{magnitude} and its \emph{direction}.

We observe (see Appendix~\ref{app:norm_depth_all_llms}) this emergent phenomena is pervasive across LLMs, independent of the model family, architecture or norm formulation. We provide a broad study analyzing 16 pre-trained open-source LLMs \textit{across 9 families, with varying architectures (Dense, Mixture-of-Experts, hybrid linear attention) and norm formulations (Pre/Peri/Post Norm)} in the Appendix~\ref{app:norm_depth_all_llms}. Below, we consider LLAMA-2~7B for our analysis.

\textbf{Magnitude grows with depth.}
Figure~\ref{fig:abu_norm_evidence}(a) shows the per-layer distribution of LLaMA-2~7B's MLP block norm weights $\gamma$. Its mean and upper percentiles rise near monotonically across layers, with its distribution mean, upper quantiles, showing a clear depth-conditioned trajectory. Note that the \textit{upper bounds of the gamma distributions} themselves are bounded by the respective subsequent layers.\\
\noindent \textbf{Direction rotates with depth.}
We perform a multi-dimensional scaling (MDS) projection of the same vectors across layers. Figure~\ref{fig:abu_norm_evidence}(b) shows how the $\gamma$ vectors trace a near-monotone arc of $\sim\!143^\circ$ across depth. Successive $\gamma$ rotate coherently rather than merely rescaling.

Together, magnitude and direction appear to be consistent with a learned depth-positional code: each layer is identified by the length and orientation of its norm weights $\gamma$. As we find broadly, this phenomenon is most pronounced in the MLP block's norm (post-attention norm) compared to the Attention block's norms (input norms). We make this analogy explicit in Section~\ref{sec:abu_layerrope}.

\begin{figure}[t]
    \centering
    \includegraphics[width=0.85\linewidth]{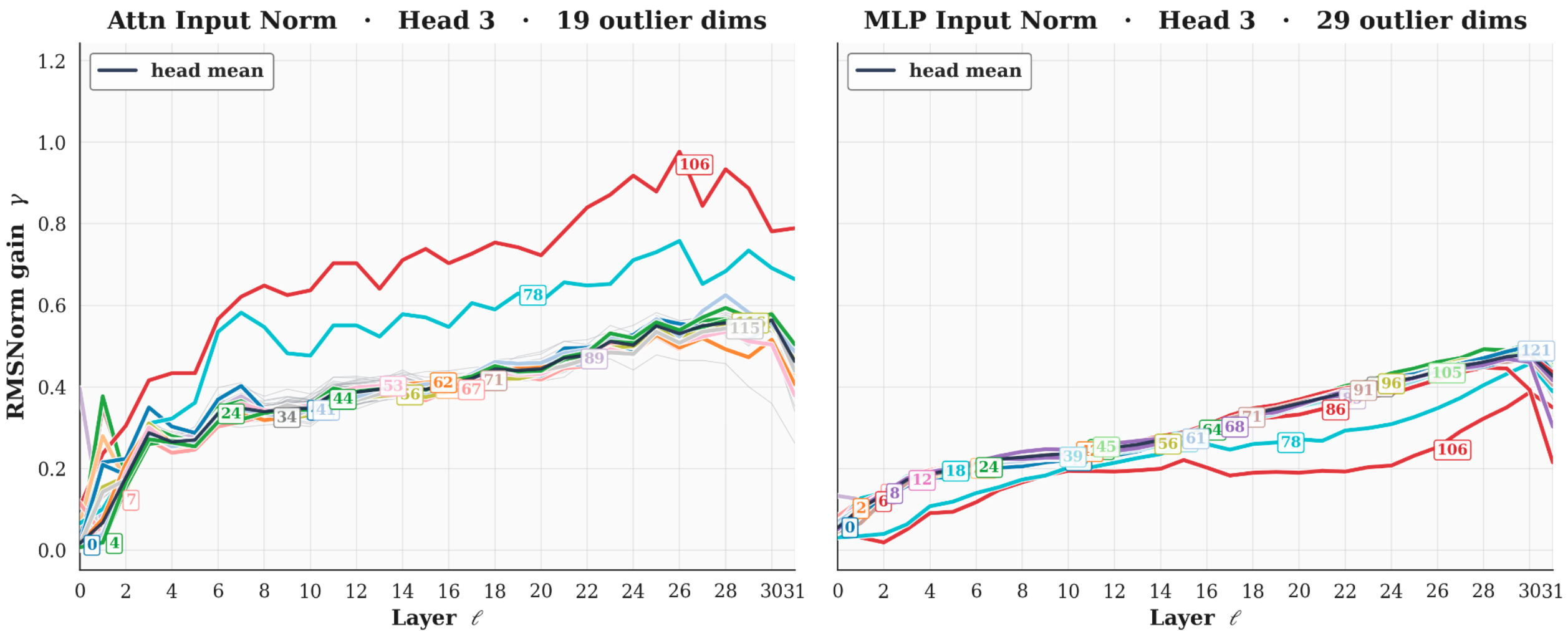}
    \vspace{-5pt}
    \caption{Per-dimension RMSNorm gain $\gamma$ across layers, for the Attention-input (left panel) and MLP-input (right panel) norms of a particular head (Head 3), in LLaMA-2 7B. Almost all dimensions share a single, near-monotone depth-conditioned trajectory.}
    \label{fig:abu_head3}
\end{figure}

\textbf{Head-wise Decomposition.\label{sec:headwise}} Decomposing $\gamma$ across dimensions for each head of the hidden dimensions (Figure~\ref{fig:abu_head3}) reveals that $\sim\!99\%$ of dimensions follow a single near-monotone trajectory; this shared schedule drives the magnitude growth, with the rotation coming from a sparse $\sim\!1\%$ of outlier dimensions. The depth encoding is therefore almost \textit{entirely a per-layer scalar depth-conditioning of the same magnitude trajectory} shared across thousands of dimensions, with direction supplied by a handful of outliers. This decomposition motivates our LayerRoPE parameterization (Section~\ref{sec:abu_layerrope}): replace the bulk per-layer $\gamma$ with a single shared vector plus a small set of layer-conditioning scalars, and let an explicit complex factor carry the magnitude and rotation that the
network would otherwise distribute across $L$ independent vectors. See Appendix~\ref{app:headwise_decomposition} for a more detailed analysis of the head-wise decomposition.

\vspace{-5pt}
\section{LayerRoPE}
\label{sec:abu_methods}\label{sec:abu_layerrope}

Intuitively, \textit{LayerRoPE is RoPE along the depth axis}: where RoPE~\citep{su2024roformer} encodes token position by rotating feature pairs, LayerRoPE encodes the \textit{layer index} by rotating and scaling a global per-feature gain $\gamma$ with depth, to get layer-wise gains $\gamma_\ell$. Applied at the input and output of every block, this explicit depth code regulates how each block reads from and writes to the residual stream, rather than suppressing the stream itself. 

Transformer normalization layers commonly apply such a learned per-feature gain $\gamma_\ell$ to the normalized activations $\hat{x}$, yielding $\gamma_\ell \odot \hat{x}$, along with an optional additive bias. A standard Pre-Norm transformer stores $L$ such independent $\gamma_\ell$ across the $L$ layers; LayerRoPE instead generates every $\gamma_\ell$ from a site-specific global shared $\gamma \in \mathbb{R}^{d}$ and a depth-conditioned complex factor, with one shared $\gamma$ for the input and output of each block. For each of the four sites (input, output of the Attention, MLP blocks) therefore, one $\gamma$ is shared across all layers. LayerRoPE additionally conditions the layer-wise $\gamma_\ell$, using just four site-specific scalars $(\alpha^{\text{mag}}, \beta^{\text{mag}}), (\alpha^{\text{rot}}, \beta^{\text{rot}})$; a reduction from $O(L\,d)$ per-layer gain parameters, to $O(d)$.

\vspace{-10pt}

\paragraph{Construction.} LayerRoPE multiplies global $\gamma$ by a single depth-conditioned complex number, with both magnitude and rotational components,
\begin{equation}
\gamma_\ell \;=\; \gamma \odot_c c(\ell),
\qquad
c(\ell, j) \;=\; \underbrace{e^{\,r(\ell)}}_{\text{magnitude}} \, \underbrace{e^{\,i\,\theta(\ell, j)}}_{\text{rotation}}
\label{eq:layerrope}
\end{equation}
where $\odot_c$ is the complex Hadamard product, $\ell \in [0,L-1]$ the layers index, and $j$ the vector dimension. Each layer's gain is thus the shared $\gamma$ scaled and rotated by depth conditioned magnitude $e^{r(\ell)}$ and rotation $e^{\theta(\ell,\cdot)}$ terms. Note: $\odot_c$ can be applied by simply reading the real gain $\gamma \in \mathbb{R}^{d}$, as $d/2$ complex pairs $\tilde\gamma_j := \gamma_{2j} + i\,\gamma_{2j+1}$.

\textbf{Magnitude Dept-conditioning.} The \emph{magnitude} schedule scales all pairs equally through a log-linear function of depth, $r(\ell) = \alpha^{\text{mag}} + \beta^{\text{mag}}\log(\ell + 1)$, a single per-layer scalar, matching the shared magnitude trajectory of Section~\ref{sec:headwise}.\\
\textbf{Rotation Depth-conditioning.} The \emph{rotation} schedule spreads a per-layer base angle $\theta(\ell) = \exp\!\big(\alpha^{\text{rot}} + \beta^{\text{rot}}\log(\ell + 1)\big)$ across pairs by the fixed RoPE spectrum, $\theta(\ell, j) = \theta(\ell)\,b_0^{-2j/d}$, where $b_0$ is the RoPE analagous base-frequency hyperparameter.

\vspace{-10pt}

\paragraph{Computational cost.} Since LayerRoPE is input-independent and computed once per forward pass, it is  quite efficient; Adding no matrix multiplications, and with an at most $0.013\%$ training FLOPs change over Pre-Norm across 60M-1B scales, and effectively none over Peri-Norm. Analytical FLOP estimates are provided in Appendix~\ref{app:computational_analysis}. LayerRoPE also replaces the $2Ld$ (Pre-Norm) or $4Ld$ (Peri-Norm) per-layer gain parameters with just $4d + 16$ parameters; causing a net parameter \emph{reduction} compared to both Pre- and Peri-Norm backbones ($90$K and $188$K fewer at 1B).

\section{Experiments and Results}
\label{sec:abu_experiments}

We study LayerRoPE across an extensive set of experiments, studying parameter (Section~\ref{sec:scaling-ladder}) and depth scaling (Section~\ref{sec:abu_scaling}), learning rate robustness (Section~\ref{sec:lr-robustness}), as well as applying it to latent looped models and vision transformers (Section~\ref{sec:vit-looped}). We also carefully ablate and study the learning dynamics induced by LayerRoPE (Section~\ref{sec:analysis}).

\subsection{Model Scaling Ladder: Scaling LayerRoPE to \(100\text{B+}\) tokens}
\label{sec:scaling-ladder}
We begin by applying LayerRoPE to a model scaling ladder from 58M to 1.3B parameters, training every model on C4 at $4\times$ the Chinchilla-optimal budget~\citep{hoffmann2022training}, upto 107B tokens.

\vspace{-5pt}

\paragraph{Setup.} We pre-train LLaMA-style~\citep{touvron2023llama} backbones on C4~\citep{dodge2021documenting} with the same training data and ordering, and differ only in their specific method: Pre-Norm, Post-Norm,  Peri-Norm~\citep{kim2025peri}, Layer-Norm Scaling (LNS)~\citep{sun2025curse}, and LayerRoPE on the Pre-Norm backbone. Inspired by~\citet{bi2024deepseek}, we carefully tune every method by the same procedure: the peak learning rate is swept independently for each method at each scale from 58M to 554M, at the full budget (80 tokens per parameter) until the optimum is bracketed, i.e., the loss diverges on both sides of it, and then extrapolated to 1.3B with a per-method power-law fit (Appendix~\ref{app:lr-sweeps}). LayerRoPE's own hyperparameters are held fixed across all scales. The tuned baselines are then used to fit a joint power law scaling curve $L(C) = E + A_m C^{-\alpha_m}$, which we report in Figure~\ref{fig:compute-scaling}. We also report downstream evaluations on all the tuned baselines in Table~\ref{tab:downstream_scaling_layerrope}. To test stability and convergence at larger scale, we further train all methods at 6.7B on a reduced budget (Appendix~\ref{app:scaling-7b}). Details of the setup, compute accounting and fit are provided in Appendix~\ref{app:scaling}; a discussion of the analytical FLOP estimates in Appendix~\ref{app:computational_analysis}. Finally, to guard against multiple-seed variability, we ran three seeds of the 58M and 134M scales for all methods at their optimal learning rates, and report these in Appendix~\ref{app:multi-seed}.


\begin{table}[h]
  \vspace{-5pt}
  \centering
  \small
  \setlength{\tabcolsep}{6pt}
  \resizebox{\textwidth}{!}{%
  \begin{tabular}{c l cccccccc c c}
    \toprule
    & & \multicolumn{8}{c}{\textbf{Accuracy} (\%, $\uparrow$)} & & \\
    \cmidrule(lr){3-10}
    & \textbf{Method} & \textbf{ARC-e} & \textbf{HellaS.} & \textbf{PIQA} & \textbf{BoolQ} & \textbf{OBQA} & \textbf{Wino.} & \textbf{LMB.} & \textbf{MMLU} & \textbf{Avg} ($\uparrow$) & \textbf{LMB PPL} ($\downarrow$) \\
    \midrule
    \multirow{5}{*}{\rotatebox[origin=c]{90}{\shortstack[c]{\textbf{134M}\\[2pt]{\scriptsize 10.7\,B}}}} & Pre-Norm & 35.0 & 31.1 & \underline{63.9} & 56.2 & \textbf{28.8} & 48.2 & 22.6 & \underline{23.0} & 40.8 & 110.5 \\
     & Post-Norm & 36.2 & 30.1 & 62.1 & 58.8 & 27.2 & 51.1 & 21.1 & \textbf{23.0} & 41.0 & 125.1 \\
     & Peri-Norm & 36.6 & 31.5 & 63.0 & \textbf{60.6} & 27.8 & \textbf{51.3} & \underline{24.4} & 22.9 & \underline{42.2} & \phantom{0}\underline{83.1} \\
     & LNS & \underline{37.1} & \underline{32.0} & 63.2 & 57.3 & 27.4 & 48.8 & 23.1 & 22.9 & 41.3 & \phantom{0}93.8 \\
     & \textbf{LayerRoPE} & \textbf{37.8} & \textbf{32.6} & \textbf{64.1} & \underline{59.6} & \underline{28.6} & \underline{51.2} & \textbf{25.9} & 23.0 & \textbf{42.8} & \phantom{0}\textbf{68.5} \\
    \midrule
    \multirow{5}{*}{\rotatebox[origin=c]{90}{\shortstack[c]{\textbf{554M}\\[2pt]{\scriptsize 44.3\,B}}}} & Pre-Norm & 39.7 & 38.8 & 67.8 & \underline{59.6} & 30.0 & \textbf{54.2} & 31.5 & \underline{23.0} & 46.0 & \phantom{0}34.7 \\
     & Post-Norm & 28.7 & 26.1 & 53.6 & 45.8 & 25.2 & 51.3 & \phantom{0}0.3 & 22.9 & 33.0 & -$^{\dagger}$ \\
     & Peri-Norm & \underline{42.4} & 42.0 & 68.3 & \textbf{59.6} & 29.8 & 52.1 & \underline{35.3} & 22.9 & \underline{47.1} & \phantom{0}\underline{26.1} \\
     & LNS & 41.6 & \underline{42.7} & \underline{68.8} & 48.7 & \textbf{31.0} & 51.8 & 34.7 & \textbf{23.0} & 45.6 & \phantom{0}26.3 \\
     & \textbf{LayerRoPE} & \textbf{42.8} & \textbf{44.9} & \textbf{69.5} & 56.9 & \underline{30.6} & \underline{53.3} & \textbf{36.1} & 22.9 & \textbf{47.7} & \phantom{0}\textbf{22.5} \\
    \midrule
    \multirow{5}{*}{\rotatebox[origin=c]{90}{\shortstack[c]{\textbf{1.3B}\\[2pt]{\scriptsize 107.1\,B}}}} & Pre-Norm & 43.3 & 46.9 & 70.2 & 52.9 & 31.2 & 52.4 & 39.1 & 23.0 & 48.0 & \phantom{0}21.0 \\
     & Post-Norm & - & - & - & - & - & - & - & - & - & - \\
     & Peri-Norm & 45.0 & 50.2 & \underline{72.1} & \textbf{61.3} & \underline{34.0} & \textbf{55.8} & \underline{43.1} & \underline{23.4} & \underline{51.6} & \phantom{0}\underline{15.7} \\
     & LNS & \underline{45.7} & \underline{51.5} & 71.9 & 57.6 & 32.8 & 54.5 & 42.2 & 22.9 & 50.9 & \phantom{0}16.7 \\
     & \textbf{LayerRoPE} & \textbf{48.1} & \textbf{53.8} & \textbf{72.3} & \underline{58.6} & \textbf{35.2} & \underline{55.5} & \textbf{44.7} & \textbf{23.8} & \textbf{52.6} & \phantom{0}\textbf{14.0} \\
    \bottomrule
  \end{tabular}
  }
\vspace{-5pt}
\caption{\textbf{Downstream Evaluations on scaling ladder.} Zero-shot accuracy of all methods at their tuned learning rate. Within each scale, \textbf{best} and \underline{second-best} are marked per column. $^{\dagger}$ indicates outlier Lambada perplexity ($>10^{4}$). (-) indicate Post-Norm failed to converge well at 1.3B scale.}
\vspace{-15pt}
  \label{tab:downstream_scaling_layerrope}
\end{table}

\paragraph{Results.}\textit{LayerRoPE reaches Pre-Norm's 1.3B loss with an estimated \textit{\textbf{3.4x less training compute}} (Figure~\ref{fig:compute-scaling}), while using fewer parameters and at most $0.013\%$ more training FLOPs (Table~\ref{tab:flop-overhead}).} LayerRoPE attains the lowest loss at every scale (Figure~\ref{fig:compute-scaling}), and its margin over Pre-Norm \textit{roughly doubles} from 58M to 1.3B. The joint power-law fit also gives LayerRoPE the steepest compute exponent of all methods (Figure~\ref{fig:compute-scaling}). The gains carry over to downstream tasks (Table~\ref{tab:downstream_scaling_layerrope}): LayerRoPE has the lowest Lambada perplexity at every scale and the highest average accuracy at essentially every scale, and at 1.3B it is best on six of the eight tasks. At 6.7B, LayerRoPE trains stably and outperforms every baseline in both loss and average accuracy (Appendix~\ref{app:scaling-7b}). Over multi-seed runs at 58M and 134M scales, LayerRoPE retains the lowest loss under every seed (Appendix~\ref{app:multi-seed}).

\vspace{-10pt}

\begin{figure}[t]
\vspace{-5pt}
  \centering
  \includegraphics[width=\linewidth]{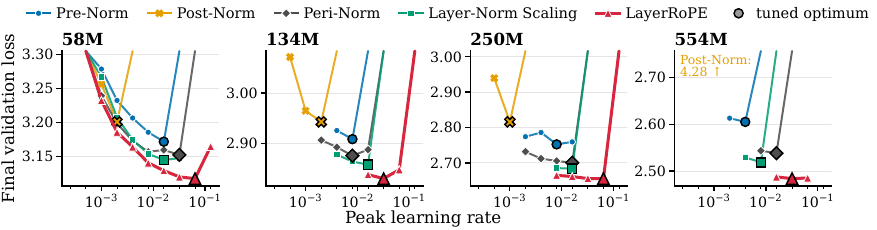}
  \vspace{-20pt}
  \caption{\textbf{Model Ladder Learning-rate Sweep.} Each method is swept on the shared $\times2$ grid until its optimum is interior; outlined markers indicate the tuned optima, whose rates are listed in Table~\ref{tab:scaling-ladder}.}
  \label{fig:lr-sweeps}
    \vspace{-15pt}
\end{figure}

\paragraph{Learning-rate sweeps.} Figure~\ref{fig:lr-sweeps} shows the learning-rate sweep behind the optimal LR model-ladder: the final validation loss of each run against its peak learning rate, for all five methods at the four swept scales. Because each method is compared at its own tuned optimum, no method is handicapped by a rate tuned for another. We see that: i) LayerRoPE's optimal learning rate \textit{is $4$-$8\times$ that of Pre-Norm}; ii) its loss is nearly flat over a wide range of rates around that optimum; and iii) it trains stably at rates at which every baseline diverges. The baselines, by contrast, diverge at progressively lower rates as the models grow, and at 554M Post-Norm collapses at all but its smallest rates. This motivates our analysis in Section~\ref{sec:lr-robustness} to investigate this learning rate robustness in detail.

\subsection{Depth Scaling: Scaling LayerRoPE to 512 Layers}
\label{sec:abu_scaling}

Scaling depth is the regime in which the ``curse of depth''~\citep{sun2025curse} is most severe. To isolate it, we fix a narrow LLaMA-style backbone and vary only the number of layers, $L \in \{48, 96, 192, 256, 384, 512\}$, from 18M to 111M parameters.

\begin{wrapfigure}{r}{0.6\columnwidth}
    \centering
    \vspace{-10pt}
    \includegraphics[width=\linewidth]
        {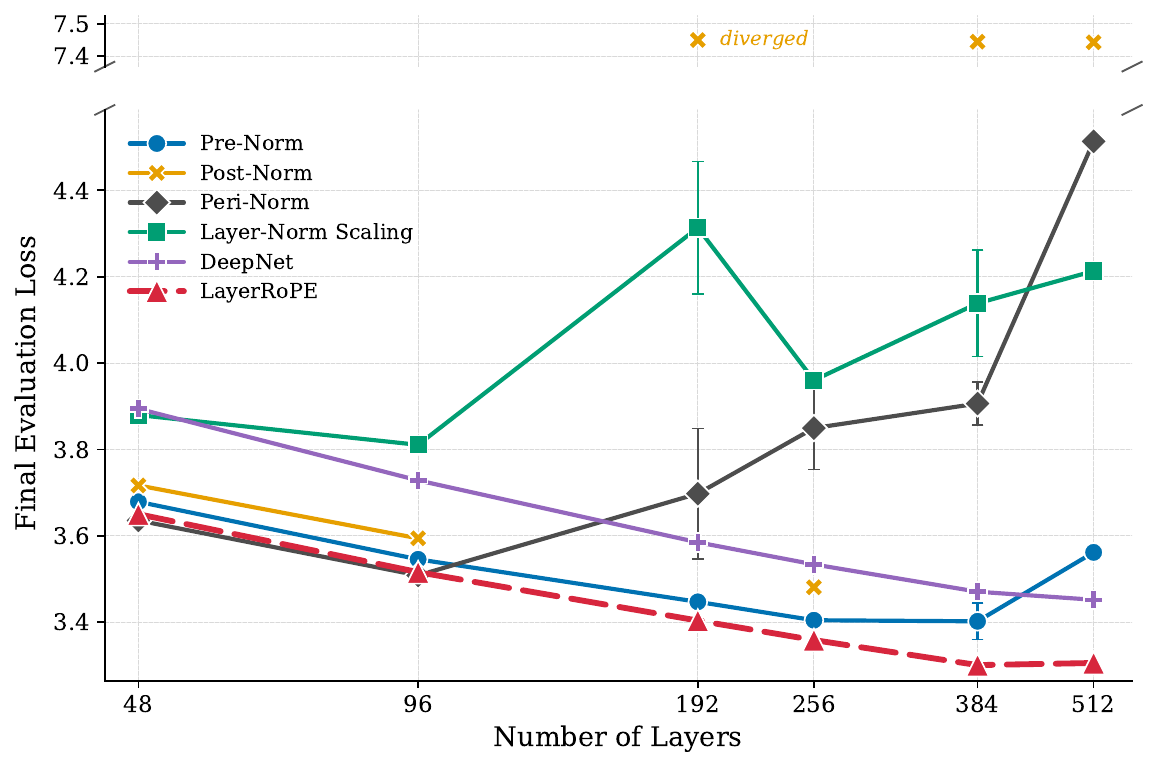}
    \caption{\textbf{Depth scaling.}
    Final validation loss against the number of layers.
    Points are means over two seeds and bars span both seeds
    (up to 384 layers). The upper strip shows diverged Post-Norm runs.}
    \label{fig:abu_depth_scaling}
    \vspace{-10pt}
\end{wrapfigure}

\paragraph{Setup.}
We compare against the standard depth-stabilizing normalizations, including DeepNet~\citep{wang2024deepnet}, which was designed to train very deep Transformers. 
Every model is trained on C4 for upto 20K steps (at least \(\sim\)1x Chinchilla optimal). 
First, we train with a shared peak learning rate of $10^{-3}$, with all methods trained with two seeds at every depth up to 384 layers; at 512 layers, we report a single seed. These depth scaling results are reported in Figure~\ref{fig:abu_depth_scaling}. Similar to the model scaling ladder, we also perform an exhaustive method-specific learning rate sweep, and report the equivalent depth scaling analysis with optimal LR in Appendix~\ref{app:depth-scaling}.

\paragraph{Results.} \textit{LayerRoPE is the only approach whose loss decreases \emph{near monotonically} with strong convergence, upto 512 layers, while also reaching the lowest loss (Figure~\ref{fig:abu_depth_scaling}) with a margin that grows with depth.} Every baseline degrades or fails with depth: Post-Norm diverges from 192 layers, Peri-Norm and Layer-Norm Scaling degrade beyond 96 layers, and Pre-Norm stops improving after 256 layers and regresses at 512. DeepNet is the only baseline that improves at every depth, but it remains well above LayerRoPE throughout. From Appendix~\ref{app:depth-scaling}, where we report an exhaustive learning-rate sweep for all methods at every depth, the equivalent depth scaling curves show that these results are method-specific (and unaffacted by exhaustive baseline tuning): with every method at its own best rate, LayerRoPE still attains the lowest loss at every depth, while Post-Norm, Peri-Norm and Layer-Norm Scaling still break down with depth. \textit{Widening rather than damping the residual stream therefore scales strictly better with depth. }

\vspace{-5pt}

\subsection{LayerRoPE Widens the Learning-Rate Basin and Delays Divergence}
\label{sec:lr-robustness}
\begin{figure}[htb]
  \centering
  \includegraphics[width=\linewidth]{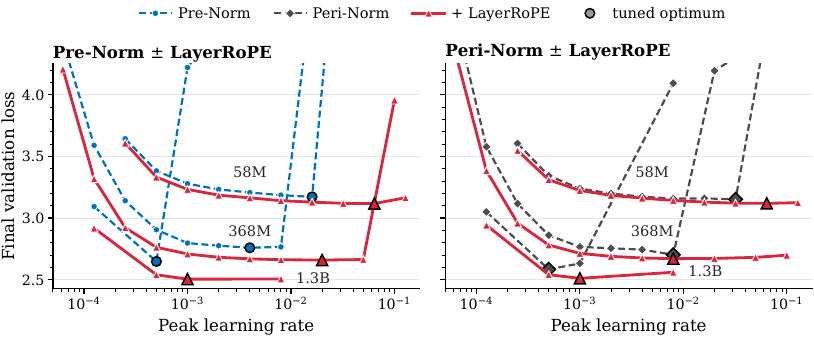}
  \caption{\textbf{Learning Rate Basins.} Final validation loss against peak learning rate for Pre-Norm (left) and Peri-Norm (right), each with and without LayerRoPE, at 58M, 368M and 1.3B scales. Runs with losses too high for the panel or diverged, are omitted.}
  \vspace{-20pt}
  \label{fig:lr-robustness}
\end{figure}

\vspace{-5pt}
\paragraph{Setup.}
We sweep the optimal learning rate over two to three orders of magnitude for Pre-Norm and Peri-Norm, each with and without LayerRoPE, at 58M, 368M and 1.3B scales (Figure~\ref{fig:lr-robustness}). The 58M  368M and 1.3B models are trained on C4 with each scale sharing the same data, recipe and seed, and LayerRoPE keeping the configuration of Section~\ref{sec:scaling-ladder} without any tuning. Figure~\ref{fig:lr-robustness} reports the final validation loss for all model scales across this sweep. We also measure the LR sensitivity~\citep{wortsman2024small} metric for LayerRoPE in Appendix~\ref{app:lr-dense}.

\vspace{-10pt}

\paragraph{Results.}
\textit{LayerRoPE promotes a wider basin and later divergence.} Learning-rate robustness is usually characterized in two ways. One, by how quickly the final loss degrades as the rate moves away from its optimum, summarized as LR sensitivity~\citep{wortsman2024small}; Two, by the largest rate at which training remains stable, which normalization placement is known to govern~\citep{xiong2020layer,kim2025peri}. We show that LayerRoPE improves both, on both backbones and at every scale (Figure~\ref{fig:lr-robustness}), and it changes the loss vs learning-rate curve in three distinct ways: \textbf{\emph{(i) A better optimum.} }LayerRoPE's best loss is lower than its baseline's in all six settings, and in five of them it is reached at a higher learning-rate. \textbf{\emph{(ii) A wider basin.}} Around its optimum, LayerRoPE's loss is nearly flat. At 368M on the Pre-Norm backbone, it varies negligibly from $4{\times}10^{-3}$ to $5{\times}10^{-2}$ (a $\sim10\times$ range of learning rates); a range over which Pre-Norm itself diverges. \textbf{\emph{(iii) Later divergence.}} LayerRoPE trains stably at rates up to $\sim8\times$ higher than those at which its baseline first fails, and at 58M and 1.3B on both backbones and at 368M on Peri-Norm it trains at every rate we swept. 
\textit{Together, these effects reduce LR sensitivity by $3\times$ to $10\times$ in every setting (Table~\ref{tab:lr-robustness}).}

The learning rate sweeps of the scaling ladder extend this picture to all five methods at the full $4\times$ Chinchilla budget (see Figure~\ref{fig:lr-sweeps}, Appendix~\ref{app:lr-sweeps}): the baselines diverge at progressively lower rates as the models grow, whereas LayerRoPE keeps a flat basin at $4\times$ to $8\times$ Pre-Norm's optimal rate at every scale. LayerRoPE appears to consistently improve the sensitivity of Pre- or Peri-Norm backbones near their optimum learning rates, and their stability away from it.

\vspace{-5pt}
\subsection{LayerRoPE for Looped Latent Models and Vision Transformers}
\label{sec:vit-looped}
\vspace{-5pt}
We next test whether LayerRoPE transfers out of the box to other architectures and modalities, by applying it without any tuning to looped latent language models and to Vision Transformers.

\begin{figure}[htb]
  \centering
  \begin{minipage}[b]{0.43\linewidth}
    \scriptsize
    \setlength{\tabcolsep}{4pt}
    \renewcommand{\arraystretch}{1.15}
    \begin{tabular*}{\linewidth}{@{\extracolsep{\fill}}c l ccc@{}}
      \toprule
      & & \multicolumn{2}{c}{PPL\,$\downarrow$} & Acc.\,(\%)\,$\uparrow$ \\
      \cmidrule(lr){3-4}\cmidrule(l){5-5}
      & Model & Val. & Lambada & Avg. \\
      \midrule
      \multirow{3}{*}{\rotatebox[origin=c]{90}{\shortstack[c]{\footnotesize\textbf{140M}\\[1pt]{\scriptsize 11.2\,B}}}} & GPT & 20.28 & 105.6 & 29.9 \\
      & Parcae & 18.66 & \phantom{0}81.2 & 33.0 \\
      & \hspace{0.5em}\textbf{+\,LayerRoPE} & \textbf{18.36} & \textbf{\phantom{0}67.3} & \textbf{34.9} \\
      \addlinespace[3pt]
      \multirow{3}{*}{\rotatebox[origin=c]{90}{\shortstack[c]{\footnotesize\textbf{370M}\\[1pt]{\scriptsize 29.6\,B}}}} & GPT & 15.01 & \phantom{0}37.2 & 40.6 \\
      & Parcae & 14.22 & \phantom{0}32.5 & 43.4 \\
      & \hspace{0.5em}\textbf{+\,LayerRoPE} & \textbf{14.05} & \textbf{\phantom{0}28.0} & \textbf{44.0} \\
      \addlinespace[3pt]
      \multirow{3}{*}{\rotatebox[origin=c]{90}{\shortstack[c]{\footnotesize\textbf{770M}\\[1pt]{\scriptsize 61.6\,B}}}} & GPT & 12.80 & \phantom{0}20.8 & 47.0 \\
      & Parcae & 12.49 & \phantom{0}18.6 & 50.1 \\
      & \hspace{0.5em}\textbf{+\,LayerRoPE} & \textbf{12.09} & \textbf{\phantom{0}18.4} & \textbf{50.6} \\
      \bottomrule
    \end{tabular*}
    \par\vspace{0pt}
  \end{minipage}\hfill
  \begin{minipage}[b]{0.54\linewidth}
    \includegraphics[width=\linewidth]{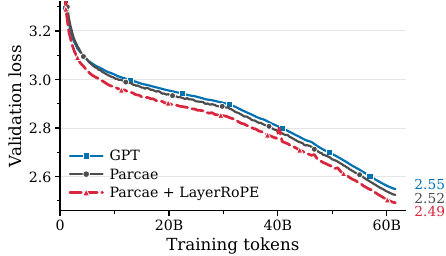}
    \par\vspace{0pt}
  \end{minipage}\par\nointerlineskip
  \captionsetup{position=bottom,skip=6pt,justification=raggedright,singlelinecheck=false}
  \begin{minipage}[t]{0.43\linewidth}
    \captionof{table}{\textbf{Looped latent models.} Validation and Lambada ppl. and downstream accuracy, with looped models at $T{=}8$.}
    \label{tab:parcae_layerrope_filtered}
  \end{minipage}\hfill
  \begin{minipage}[t]{0.54\linewidth}
    \caption{\textbf{Looped latent models at 770M.} Validation loss over training, with looped models at $T{=}8$; final losses at the right. The loss axis omits the first ${\sim}$1B tokens.}
    \label{fig:loss_770m_3way}
  \end{minipage}
\end{figure}

\paragraph{Looped latent models.}
Parcae~\citep{prairie2026parcae} is a looped language model: a prelude of Transformer layers embeds the input, a recurrent block of layers is applied $T$ times to a latent state, and a coda decodes the result. We train Parcae at 140M, 370M and 770M parameters with its official recipe, on 11.2B, 29.6B and 61.6B tokens of FineWeb-Edu~\citep{penedo2024fineweb}, alongside a non-looped GPT baseline with the same parameter count and recipe. We apply LayerRoPE at the input and output of every layer, as in Section~\ref{sec:abu_layerrope}: each layer of the recurrent block keeps its position in the network as its depth index at every iteration, and the magnitude slope is initialized at zero. Nothing else is changed; the official recipe, including its optimizer settings and stabilization, is used as is, without any tuning for LayerRoPE.

LayerRoPE improves Parcae at every scale (Table~\ref{tab:parcae_layerrope_filtered}): Parcae~$+$~LayerRoPE is the best model on nearly every metric, with lower validation perplexity, markedly lower Lambada perplexity at 140M and 370M, and higher average downstream accuracy. At 770M, its validation loss stays below both Parcae's and GPT's throughout training (Figure~\ref{fig:loss_770m_3way}), and its advantage over Parcae holds at every number of loop iterations, diminishing as loops are added (see Figure~\ref{fig:loss_770m_tsweep}). Critically, we observe that Parcae~$+$~LayerRoPE's recurrent state grows to norms of ${\sim}10^{9}$ at 770M against Parcae's ${\sim}10^{3}$ (see Figure~\ref{fig:parcae-recurrent-state}; Appendix~\ref{app:looped}), yet its token states remain markedly less collinear for most of training. LayerRoPE thus appears to make fuller use of the recurrent block's capacity, suggesting that a controlled growth in norm can aid, rather than hinder, the use of depth. See Appendix~\ref{app:looped} for details and further analysis.

\begin{wraptable}{r}{0.45\linewidth}
  \vspace{-13pt}
  \footnotesize
  \setlength{\tabcolsep}{4pt}
  \renewcommand{\arraystretch}{1.05}
  \begin{tabular*}{\linewidth}{@{\extracolsep{\fill}}l ccc@{}}
    \toprule
    & \multicolumn{2}{c}{Acc.\,(\%)\,$\uparrow$} & Loss\,$\downarrow$ \\
    \cmidrule(lr){2-3}\cmidrule(l){4-4}
    Model & Top-1 & Top-5 & Val. \\
    \midrule
    ViT-T & 72.21 & 91.26 & 1.211 \\
    \hspace{0.5em}\textbf{+\,LayerRoPE} & \textbf{73.08} & \textbf{91.62} & \textbf{1.171} \\
    \addlinespace[3pt]
    ViT-S & 79.80 & 94.89 & 0.897 \\
    \hspace{0.5em}\textbf{+\,LayerRoPE} & \textbf{80.38} & \textbf{95.26} & \textbf{0.865} \\
    \bottomrule
  \end{tabular*}
  \captionsetup{position=bottom,skip=4pt,justification=raggedright}
  \caption{\textbf{ViTs on ImageNet1k.} Each metric is measured as the best over training.}
  \label{tab:vit_layerrope_lns}
  \vspace{-15pt}
\end{wraptable}

\vspace{-10pt}

\paragraph{Vision Transformers.}
We train both \mbox{ViT-T} and \mbox{ViT-S} \citep{dosovitskiy2020image} on ImageNet-1k~\citep{russakovsky2015imagenet} for 300 epochs with the DeiT recipe~\citep{touvron2021training}. Similar to LLMs, we apply LayerRoPE naively at every block, conditioning the gains \(\gamma\) of the pre-norm LayerNorms (while keeping the default layer-wise centering and bias), with the magnitude slope initialized at zero; the recipe is otherwise unchanged. LayerRoPE improves both ViTs on every metric (see Table~\ref{tab:vit_layerrope_lns}), raising top-1 accuracy by 0.9 points for \mbox{ViT-T} and 0.6 points for \mbox{ViT-S}, at even slightly fewer parameters.

\vspace{-10pt}

\section{Analysis and Ablations}
\label{sec:analysis}
\vspace{-10pt}
We now examine how LayerRoPE achieves these gains: how it shapes the residual stream across depth (Section~\ref{sec:abu_learned_dynamics}) and which depth schedule it learns (Appendix~\ref{app:depth-conditioning}). We then ablate its components (Section~\ref{sec:abu_results_layerrope_130m}).

\begin{figure}[b]
  \centering
  \includegraphics[width=\linewidth]{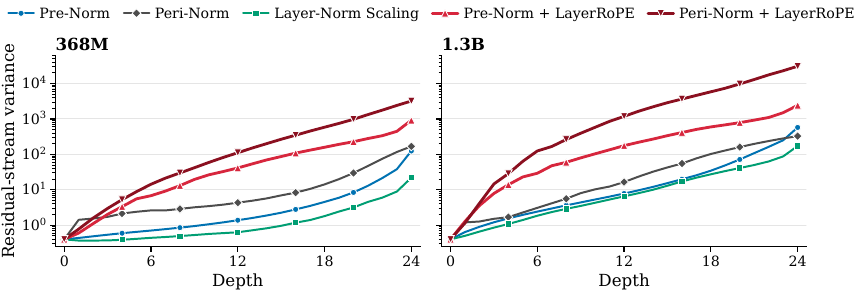}
  \caption{\textbf{LayerRoPE widens the residual stream.} Activation variance of the residual stream with model depth, across 368M and 1.3B models, measured at the output of each layer, on the C4 validation set.}
  \label{fig:variance-growth}
\end{figure}

\subsection{Analyzing Variance Growth: LayerRoPE Widens the Residual Stream}\label{sec:abu_learned_dynamics}

We measure the variance of the residual stream after every layer of the 368M and 1.3B models of Section~\ref{sec:lr-robustness}. We observe that while methods like Layer-Norm Scaling aim to suppress the variance growth observed in standard norm forms (Pre-Norm, and even more so in Peri-Norm), LayerRoPE in particular appears to do exactly the opposite and amplifies the variance growth further (Figure~\ref{fig:variance-growth}): at 1.3B, LayerRoPE on the Pre-Norm backbone ends with $4\times$ the variance of Pre-Norm and $14\times$ that of Layer-Norm Scaling, and on the Peri-Norm backbone with $94\times$ that of Peri-Norm. Rather than damping the residual stream, LayerRoPE widens it, attenuating what each block reads while learning to amplify what it writes. See Appendix~\ref{app:depth-conditioning} for more analysis of the learned depth schedule in LayerRoPE.

\subsection{Ablating LayerRoPE's Components}
\label{sec:abu_results_layerrope_130m}

We ablate LayerRoPE at 134M (Table~\ref{tab:abu_layerrope_mag_rot_combined}). Its magnitude and rotation schedules are complementary: on both backbones, each alone improves on the baseline, magnitude more than rotation, and their combination is best (Table~\ref{tab:abl-components}). This mirrors the depth code of Section~\ref{sec:norm_depth_code}, which is carried jointly by the magnitude and the direction of the gains.

\begin{table}[h]
  \centering
  \small
  \setlength{\tabcolsep}{4pt}

  \begin{subtable}[t]{0.49\linewidth}
    \centering
    \begin{tabular}{@{}l cc r@{}}
      \toprule
      & \multicolumn{3}{c}{\textbf{Peri-Norm}} \\
      \cmidrule(l){2-4}
      \textbf{Variant} & \textbf{Loss}\,$\downarrow$ & \textbf{PPL}\,$\downarrow$ & \textbf{Params} \\
      \midrule
      Baseline & 3.0720 & 21.585 & 134.12M \\
      \textbf{LayerRoPE} & \textbf{3.0252} & \textbf{20.599} & 134.09M \\
      \quad - Magnitude only & 3.0351 & 20.803 & - \\
      \quad - Rotation only & 3.0533 & 21.186 & - \\
      \bottomrule
    \end{tabular}
    \caption{Magnitude and rotation schedules.}
    \label{tab:abl-components}
  \end{subtable}
  \hfill
  \begin{subtable}[t]{0.49\linewidth}
    \centering
    \begin{tabular}{@{}l cc r@{}}
      \toprule
      & \multicolumn{3}{c}{\textbf{Pre-Norm}} \\
      \cmidrule(l){2-4}
      \textbf{Variant} & \textbf{Loss}\,$\downarrow$ & \textbf{PPL}\,$\downarrow$ & \textbf{Params} \\
      \midrule
      Baseline & 3.1003 & 22.205 & 134.11M \\
      \textbf{LayerRoPE} & \textbf{3.0317} & \textbf{20.732} & 134.09M \\
      \quad - Per-layer $\gamma$ & 3.0318 & 20.735 & 134.12M \\
      \quad - Uniform rotation & 3.0379 & 20.862 & 134.09M \\
      \bottomrule
    \end{tabular}
    \caption{Shared gains and rotation spectrum.}
    \label{tab:abl-design}
  \end{subtable}

  \caption{\textbf{Ablating LayerRoPE at 134M.}
  Validation loss and perplexity, best in \textbf{bold}; Baseline is the backbone
  without LayerRoPE. (a)~Removing either schedule degrades LayerRoPE on Peri-Norm,
  though each alone still improves on the baseline.
  (b)~Separate gains $\gamma$ per layer add parameters without improving on
  LayerRoPE's single shared $\gamma$, while one rotation angle for all feature
  pairs underperforms RoPE's spectrum of frequencies.}
  \label{tab:abu_layerrope_mag_rot_combined}
\end{table}

Two further design choices hold up (Table~\ref{tab:abl-design}). Learning a separate $\gamma$ for every layer, instead of generating all gains from one shared $\gamma$, adds 33.8K parameters without lowering the loss: as the head-wise analysis predicts (Section~\ref{sec:headwise}), per-layer gains are largely redundant. Rotating every feature pair by a single angle, instead of by RoPE's spectrum of frequencies, raises perplexity, so the multi-frequency structure like that of RoPE, matters along depth as well.

\section{Discussion and Conclusion}\label{sec:abu_discussion}

\noindent\textbf{Norm weights encode depth.} The growth of the residual stream has been read almost universally as a pathology to suppress. Across 16 pre-trained LLMs, we find that it reflects an emergent depth code in the per-layer norm weights, which grow in magnitude and rotate in direction layer by layer, with ${\sim}99\%$ of dimensions sharing a single trajectory. LayerRoPE makes this code explicit by applying RoPE along depth rather than along the sequence: every per-layer gain is generated from one shared $\gamma$ and four scalars per site, with fewer parameters than the gains it replaces.

\noindent\textbf{Regulate the blocks, not the stream.} With depth made explicit, LayerRoPE attains the lowest loss at every scale we train, is the only method whose loss keeps improving with strong convergence, up to 512 layers, reduces learning-rate sensitivity by $3$-$10\times$, and improves looped models and Vision Transformers without tuning. Its learned schedule inverts the premise behind prior remedies: rather than suppressing the residual stream, LayerRoPE \emph{widens} it, to between $4\times$ to $100\times$ its backbones' variance, by damping what each block reads while amplifying what it writes (Section~\ref{sec:abu_learned_dynamics}). The same preference surfaces elsewhere. Layer-Norm Scaling's learned gains largely undo its fixed down-scaling, and LayerRoPE lets a looped model's recurrent state grow six orders of magnitude beyond Parcae's while its token states stay less collinear for most of training (Appendices~\ref{app:looped} and~\ref{app:depth-conditioning}). Depth stability, our results suggest, depends less on how large the residual stream grows than on whether each block can scale what it reads and writes to its depth.

Transformers have long been told where each token sits in the sequence. Our results suggest they also benefit from knowing where each layer sits in depth.

\clearpage

\subsubsection*{Acknowledgments}

This work was partly supported by NSF EFRI Award 2317706 and NSF CAREER Awards 2047556/2326491. The views and conclusions contained herein are those of the authors and should not be interpreted as representing any sponsor’s official policies or endorsements. We gratefully acknowledge use of the research computing resources of the Empire AI Consortium, Inc, with support from Empire State Development of the State of New York, the Simons Foundation, and the Secunda Family Foundation.  This work was also supported in part by compute credits provided by Modal Labs through the Modal for Academics program.

\bibliography{iclr2027_conference}

@article{vaswani2017attention,
  title={Attention is all you need},
  author={Vaswani, Ashish and Shazeer, Noam and Parmar, Niki and Uszkoreit, Jakob and Jones, Llion and Gomez, Aidan N and Kaiser, {\L}ukasz and Polosukhin, Illia},
  journal={Advances in neural information processing systems},
  volume={30},
  year={2017}
}

@String(ICCV= {Int. Conf. Comput. Vis.})

@String(ICCV  = {ICCV})

@article{touvron2023llama,
  title={Llama: Open and efficient foundation language models},
  author={Touvron, Hugo and Lavril, Thibaut and Izacard, Gautier and Martinet, Xavier and Lachaux, Marie-Anne and Lacroix, Timoth{\'e}e and Rozi{\`e}re, Baptiste and Goyal, Naman and Hambro, Eric and Azhar, Faisal and others},
  journal={arXiv preprint arXiv:2302.13971},
  year={2023}
}

@article{touvron2023llama2,
  title={Llama 2: Open foundation and fine-tuned chat models},
  author={Touvron, Hugo and Martin, Louis and Stone, Kevin and Albert, Peter and Almahairi, Amjad and Babaei, Yasmine and Bashlykov, Nikolay and Batra, Soumya and Bhargava, Prajjwal and Bhosale, Shruti and others},
  journal={arXiv preprint arXiv:2307.09288},
  year={2023}
}

@article{russakovsky2015imagenet,
  title={Imagenet large scale visual recognition challenge},
  author={Russakovsky, Olga and Deng, Jia and Su, Hao and Krause, Jonathan and Satheesh, Sanjeev and Ma, Sean and Huang, Zhiheng and Karpathy, Andrej and Khosla, Aditya and Bernstein, Michael and others},
  journal={International journal of computer vision},
  volume={115},
  number={3},
  pages={211--252},
  year={2015},
  publisher={Springer}
}

@article{jiang2023mistral,
  title={Mistral 7B},
  author={Jiang, Albert Q and Sablayrolles, Alexandre and Mensch, Arthur and Bamford, Chris and Chaplot, Devendra Singh and Casas, Diego de las and Bressand, Florian and Lengyel, Gianna and Lample, Guillaume and Saulnier, Lucile and others},
  journal={arXiv preprint arXiv:2310.06825},
  year={2023}
}

@inproceedings{biderman2023pythia,
  title={Pythia: A suite for analyzing large language models across training and scaling},
  author={Biderman, Stella and Schoelkopf, Hailey and Anthony, Quentin Gregory and Bradley, Herbie and O’Brien, Kyle and Hallahan, Eric and Khan, Mohammad Aflah and Purohit, Shivanshu and Prashanth, USVSN Sai and Raff, Edward and others},
  booktitle={International Conference on Machine Learning},
  pages={2397--2430},
  year={2023},
  organization={PMLR}
}

@article{sun2025curse,
  title={The curse of depth in large language models},
  author={Sun, Wenfang and Song, Xinyuan and Li, Pengxiang and Yin, Lu and Zheng, Yefeng and Liu, Shiwei},
  journal={arXiv preprint arXiv:2502.05795},
  year={2025}
}

@article{sun2024massive,
  title={Massive activations in large language models},
  author={Sun, Mingjie and Chen, Xinlei and Kolter, J Zico and Liu, Zhuang},
  journal={arXiv preprint arXiv:2402.17762},
  year={2024}
}

@article{xiong2020layer,
title={On Layer Normalization in the Transformer Architecture},
author={Xiong, Ruibin and Yang, Yunchang and He, Di and Zheng, Kai and Zheng, Shuxin and Xing, Chen and Zhang, Huishuai and Lan, Yanyan and Wang, Liwei and Liu, Tieyan},
journal={arXiv preprint arXiv:2002.04745},
year={2020}
}

@article{xiao2023efficient,
  title={Efficient streaming language models with attention sinks},
  author={Xiao, Guangxuan and Tian, Yuandong and Chen, Beidi and Han, Song and Lewis, Mike},
  journal={arXiv preprint arXiv:2309.17453},
  year={2023}
}

@inproceedings{darcet2024vision,
  title={Vision transformers need registers},
  author={Darcet, Timoth{\'e}e and Oquab, Maxime and Mairal, Julien and Bojanowski, Piotr},
  booktitle={International conference on learning representations},
  volume={2024},
  pages={2632--2652},
  year={2024}
}

@inproceedings{dodge2021documenting,
  title={Documenting large webtext corpora: A case study on the colossal clean crawled corpus},
  author={Dodge, Jesse and Sap, Maarten and Marasovi{\'c}, Ana and Agnew, William and Ilharco, Gabriel and Groeneveld, Dirk and Mitchell, Margaret and Gardner, Matt},
  booktitle={Proceedings of the 2021 conference on empirical methods in natural language processing},
  pages={1286--1305},
  year={2021}
}

@article{penedo2024fineweb,
  title={The fineweb datasets: Decanting the web for the finest text data at scale},
  author={Penedo, Guilherme and Kydl{\'\i}{\v{c}}ek, Hynek and Lozhkov, Anton and Mitchell, Margaret and Raffel, Colin and Von Werra, Leandro and Wolf, Thomas and others},
  journal={Advances in Neural Information Processing Systems},
  volume={37},
  pages={30811--30849},
  year={2024}
}

@article{su2024roformer,
  title={Roformer: Enhanced transformer with rotary position embedding},
  author={Su, Jianlin and Ahmed, Murtadha and Lu, Yu and Pan, Shengfeng and Bo, Wen and Liu, Yunfeng},
  journal={Neurocomputing},
  volume={568},
  pages={127063},
  year={2024},
  publisher={Elsevier}
}

@article{xie2025mhc,
  title={mhc: Manifold-constrained hyper-connections},
  author={Xie, Zhenda and Wei, Yixuan and Cao, Huanqi and Zhao, Chenggang and Deng, Chengqi and Li, Jiashi and Dai, Damai and Gao, Huazuo and Chang, Jiang and Yu, Kuai and others},
  journal={arXiv preprint arXiv:2512.24880},
  year={2025}
}

@article{team2026attention,
  title={Attention residuals},
  author={Team, Kimi and Chen, Guangyu and Zhang, Yu and Su, Jianlin and Xu, Weixin and Pan, Siyuan and Wang, Yaoyu and Wang, Yucheng and Chen, Guanduo and Yin, Bohong and others},
  journal={arXiv preprint arXiv:2603.15031},
  year={2026}
}

@inproceedings{zhu2025hyper,
  title={Hyper-connections},
  author={Zhu, Defa and Huang, Hongzhi and Huang, Zihao and Zeng, Yutao and Mao, Yunyao and Wu, Banggu and Min, Qiyang and Zhou, Xun},
  booktitle={International Conference on Learning Representations},
  volume={2025},
  pages={97183--97219},
  year={2025}
}

@article{wang2024deepnet,
  title={Deepnet: Scaling transformers to 1,000 layers},
  author={Wang, Hongyu and Ma, Shuming and Dong, Li and Huang, Shaohan and Zhang, Dongdong and Wei, Furu},
  journal={IEEE Transactions on Pattern Analysis and Machine Intelligence},
  volume={46},
  number={10},
  pages={6761--6774},
  year={2024},
  publisher={IEEE}
}

@inproceedings{yang2024tensor,
  title={Tensor programs VI: Feature learning in infinite depth neural networks},
  author={Yang, Greg and Yu, Dingli and Zhu, Chen and Hayou, Soufiane},
  booktitle={International Conference on Learning Representations},
  volume={2024},
  pages={55099--55150},
  year={2024}
}

@article{kim2025peri,
  title={Peri-ln: Revisiting normalization layer in the transformer architecture},
  author={Kim, Jeonghoon and Lee, Byeongchan and Park, Cheonbok and Oh, Yeontaek and Kim, Beomjun and Yoo, Taehwan and Shin, Seongjin and Han, Dongyoon and Shin, Jinwoo and Yoo, Kang Min},
  journal={arXiv preprint arXiv:2502.02732},
  year={2025}
}

@article{kaplan2020scaling,
  title={Scaling laws for neural language models},
  author={Kaplan, Jared and McCandlish, Sam and Henighan, Tom and Brown, Tom B and Chess, Benjamin and Child, Rewon and Gray, Scott and Radford, Alec and Wu, Jeffrey and Amodei, Dario},
  journal={arXiv preprint arXiv:2001.08361},
  year={2020}
}

@article{hoffmann2022training,
  title={Training compute-optimal large language models},
  author={Hoffmann, Jordan and Borgeaud, Sebastian and Mensch, Arthur and Buchatskaya, Elena and Cai, Trevor and Rutherford, Eliza and Casas, Diego de Las and Hendricks, Lisa Anne and Welbl, Johannes and Clark, Aidan and others},
  journal={arXiv preprint arXiv:2203.15556},
  year={2022}
}

@article{grattafiori2024llama,
  title={The llama 3 herd of models},
  author={Grattafiori, Aaron and Dubey, Abhimanyu and Jauhri, Abhinav and Pandey, Abhinav and Kadian, Abhishek and Al-Dahle, Ahmad and Letman, Aiesha and Mathur, Akhil and Schelten, Alan and Vaughan, Alex and others},
  journal={arXiv preprint arXiv:2407.21783},
  year={2024}
}

@article{yang2025qwen3,
  title={Qwen3 technical report},
  author={Yang, An and Li, Anfeng and Yang, Baosong and Zhang, Beichen and Hui, Binyuan and Zheng, Bo and Yu, Bowen and Gao, Chang and Huang, Chengen and Lv, Chenxu and others},
  journal={arXiv preprint arXiv:2505.09388},
  year={2025}
}

@article{agarwal2025gpt,
  title={gpt-oss-120b \& gpt-oss-20b model card},
  author={Agarwal, Sandhini and Ahmad, Lama and Ai, Jason and Altman, Sam and Applebaum, Andy and Arbus, Edwin and Arora, Rahul K and Bai, Yu and Baker, Bowen and Bao, Haiming and others},
  journal={arXiv preprint arXiv:2508.10925},
  year={2025}
}

@article{team2024gemma,
  title={Gemma 2: Improving open language models at a practical size},
  author={Team, Gemma and Riviere, Morgane and Pathak, Shreya and Sessa, Pier Giuseppe and Hardin, Cassidy and Bhupatiraju, Surya and Hussenot, L{\'e}onard and Mesnard, Thomas and Shahriari, Bobak and Ram{\'e}, Alexandre and others},
  journal={arXiv preprint arXiv:2408.00118},
  year={2024}
}

@article{glm2024chatglm,
  title={Chatglm: A family of large language models from glm-130b to glm-4 all tools},
  author={Glm, Team and Zeng, Aohan and Xu, Bin and Wang, Bowen and Zhang, Chenhui and Yin, Da and Zhang, Dan and Rojas, Diego and Feng, Guanyu and Zhao, Hanlin and others},
  journal={arXiv preprint arXiv:2406.12793},
  year={2024}
}

@article{olmo20242,
  title={2 OLMo 2 Furious},
  author={OLMo, Team and Walsh, Pete and Soldaini, Luca and Groeneveld, Dirk and Lo, Kyle and Arora, Shane and Bhagia, Akshita and Gu, Yuling and Huang, Shengyi and Jordan, Matt and others},
  journal={arXiv preprint arXiv:2501.00656},
  year={2024}
}

@article{bi2024deepseek,
  title={Deepseek llm: Scaling open-source language models with longtermism},
  author={Bi, Xiao and Chen, Deli and Chen, Guanting and Chen, Shanhuang and Dai, Damai and Deng, Chengqi and Ding, Honghui and Dong, Kai and Du, Qiushi and Fu, Zhe and others},
  journal={arXiv preprint arXiv:2401.02954},
  year={2024}
}

@inproceedings{wortsman2024small,
  title={Small-scale proxies for large-scale transformer training instabilities},
  author={Wortsman, Mitchell and Liu, Peter and Xiao, Lechao and Everett, Katie and Alemi, Alexander and Adlam, Ben and Co-Reyes, John D and Gur, Izzeddin and Kumar, Abhishek and Novak, Roman and others},
  booktitle={International Conference on Learning Representations},
  volume={2024},
  pages={49844--49869},
  year={2024}
}

@article{prairie2026parcae,
  title={Parcae: Scaling laws for stable looped language models},
  author={Prairie, Hayden and Novack, Zachary and Berg-Kirkpatrick, Taylor and Fu, Daniel Y},
  journal={arXiv preprint arXiv:2604.12946},
  year={2026}
}

@article{dosovitskiy2020image,
  title={An image is worth 16x16 words: Transformers for image recognition at scale},
  author={Dosovitskiy, Alexey},
  journal={arXiv preprint arXiv:2010.11929},
  year={2020}
}

@inproceedings{touvron2021training,
  title={Training data-efficient image transformers \& distillation through attention},
  author={Touvron, Hugo and Cord, Matthieu and Douze, Matthijs and Massa, Francisco and Sablayrolles, Alexandre and J{\'e}gou, Herv{\'e}},
  booktitle={International conference on machine learning},
  pages={10347--10357},
  year={2021},
  organization={PMLR}
}

@inproceedings{bachlechner2021rezero,
  title={Rezero is all you need: Fast convergence at large depth},
  author={Bachlechner, Thomas and Majumder, Bodhisattwa Prasad and Mao, Henry and Cottrell, Gary and McAuley, Julian},
  booktitle={Uncertainty in artificial intelligence},
  pages={1352--1361},
  year={2021},
  organization={PMLR}
}

@inproceedings{touvron2021going,
  title={Going deeper with image transformers},
  author={Touvron, Hugo and Cord, Matthieu and Sablayrolles, Alexandre and Synnaeve, Gabriel and J{\'e}gou, Herv{\'e}},
  booktitle={2021 IEEE/CVF international conference on computer vision (ICCV)},
  pages={32--42},
  year={2021},
  organization={IEEE}
}

@article{wang2026negligible,
  title={Negligible in Size, Significant in Effect: On Scale Vectors in Large Language Models},
  author={Wang, Mingze and Zhu, Shuchen and Fang, Yuxin and Li, Binghui and Shen, Kai and Zhong, Shu},
  journal={arXiv preprint arXiv:2605.26895},
  year={2026}
}

@article{wang2026residual,
  title={On the Residual Scaling of Looped Transformers: Stability and Transferability},
  author={Wang, Shaowen and Li, Bingrui and Zhang, Ge and Huang, Wenhao and Yan, Shen and Li, Jian},
  journal={arXiv preprint arXiv:2606.18524},
  year={2026}
}

@article{wei2022outlier,
  title={Outlier suppression: Pushing the limit of low-bit transformer language models},
  author={Wei, Xiuying and Zhang, Yunchen and Zhang, Xiangguo and Gong, Ruihao and Zhang, Shanghang and Zhang, Qi and Yu, Fengwei and Liu, Xianglong},
  journal={Advances in Neural Information Processing Systems},
  volume={35},
  pages={17402--17414},
  year={2022}
}

@inproceedings{kovaleva2021bert,
  title={BERT busters: Outlier dimensions that disrupt transformers},
  author={Kovaleva, Olga and Kulshreshtha, Saurabh and Rogers, Anna and Rumshisky, Anna},
  booktitle={Findings of the Association for Computational Linguistics: ACL-IJCNLP 2021},
  pages={3392--3405},
  year={2021}
}

@article{dong2026dissecting,
  title={Dissecting Outlier Dynamics in LLM NVFP4 Pretraining},
  author={Dong, Peijie and Fan, Ruibo and Tao, Yuechen and Mou, Di and Hu, Wenhu and Tang, Zhenheng and Yu, Yinghao and Wang, Jiamang and Su, Wenbo and Yang, Guodong and others},
  journal={arXiv preprint arXiv:2602.02047},
  year={2026}
}

@article{dehghani2018universal,
  title={Universal transformers},
  author={Dehghani, Mostafa and Gouws, Stephan and Vinyals, Oriol and Uszkoreit, Jakob and Kaiser, {\L}ukasz},
  journal={arXiv preprint arXiv:1807.03819},
  year={2018}
}

@article{xu2024expressive,
  title={On expressive power of looped transformers: Theoretical analysis and enhancement via timestep encoding},
  author={Xu, Kevin and Sato, Issei},
  journal={arXiv preprint arXiv:2410.01405},
  year={2024}
}
\bibliographystyle{iclr2027_conference}

\clearpage

\appendix

\section{Computational Cost Analysis}
\label{app:computational_analysis}
We provide an analytical FLOP estimate and a parameter comparison of LayerRoPE to that of Pre and Peri-Norm backbones.

\noindent \textbf{Parameter Comparison. } LayerRoPE simply replaces the per-layer normalization weights with shared depth-conditioned norm weights that use 16 learned scalars in total (8 scalars each for magnitude and rotation, respectively). Relative to the corresponding backbone, it therefore uses fewer parameters: $2Ld-4d-16$ fewer for Pre-Norm and $4Ld-4d-16$ fewer for Peri-Norm. 

As Table~\ref{tab:flop-overhead} shows, LayerRoPE is slightly
smaller than its backbone at every scale (such as $90$K and $188$K fewer at 1B scale).

\noindent \textbf{Analytical FLOP Estimates. } LayerRoPE introduces no matrix multiplications. It's per-layer function at the input and output of the computational block can be generalized as
$\tilde{\gamma}_\ell = e^{\alpha + \beta\log(\ell+1)}\,R(\theta_\ell)\,\gamma$ depending
only on the layer index, a couple of learned scalars, and a shared $d$-dimensional $\gamma$ vector. Since it does not depend on the input, it is therefore \textit{simply computed once per forward pass} and applied as a per-channel scale. Note that the $R(\theta_\ell)$ is the RoPE rotation matrix with layer depth, instead of token position. 

Under the standard FLOP accounting, which counts dense matrix
multiplications, attention~\citep{kaplan2020scaling,hoffmann2022training} among other major operations, but omits secondary normalization and element-wise costs. LayerRoPE, therefore, adds no FLOP increase to the architecture it modifies (Pre or Peri Norm). For a more conservative estimate, we additionally count every elementwise operation introduced by LayerRoPE, assigning one FLOP per scalar addition or multiplication in both the forward and backward passes. At existing normalization layers, $\tilde{\gamma}_\ell$ therefore replaces the existing RMSNorm weight, and adds no per-token cost.

\begin{itemize}
    \item \textbf{For Pre-Norm backbones}, the only new per-token operation is the residual gate $\tilde{\gamma}_\ell \odot f(x)$ on each of the $2L$ sublayer outputs. Each application costs approximately $4d$ training FLOPs: $d$ in the forward pass and $3d$ in the backward pass (one multiplication for the activation gradient, and one multiplication plus one accumulation for the gate gradient). Thus, LayerRoPE adds $8Ld$ FLOPs per training token.
    \item \textbf{For Peri-Norm backbones}, the gate is absorbed into the weight of the output norm that Peri-Norm already applies, so the per-token cost is identical to Peri-Norm.
\end{itemize}

Following the standard convention~\citep{kaplan2020scaling,hoffmann2022training}, a LLaMA-style Transformer block with standard MHA and a SwiGLU MLP ($d_{\mathrm{ff}}=\tfrac{8}{3}d$) costs $72d^2$ matrix-multiplication FLOPs per training token. LayerRoPE's overhead over Pre-Norm is therefore at most
$8d/(72d^2)=1/(9d)$. Under the same elementwise accounting, the two additional RMSNorms of Peri-Norm cost approximately $24Ld$ training FLOPs, corresponding to an upper-bound of $1/(3d)$, three times LayerRoPE's overhead.

\begin{table}[h]
  \centering
  \small
  \setlength{\tabcolsep}{5.5pt}
  \begin{tabular}{lrrrrrrr}
    \toprule
     & & & \multicolumn{1}{c}{Pre-Norm} & \multicolumn{2}{c}{FLOPs overhead (\%)} & \multicolumn{2}{c}{LayerRoPE param reduction} \\
    \cmidrule(lr){5-6}\cmidrule(lr){7-8}
    Model & $d$ & $L$ & GFLOPs/token & vs.\ Pre-Norm & vs.\ Peri-Norm & vs.\ Pre-Norm & vs.\ Peri-Norm \\
    \midrule
    60M & 512 & 8 & 0.26 & 0.0128 & $\approx 0$ & \paramdn{6.128K} & \paramdn{14.320K} \\
    130M & 768 & 12 & 0.67 & 0.0110 & $\approx 0$ & \paramdn{15.344K} & \paramdn{33.776K} \\
    250M & 896 & 20 & 1.36 & 0.0106 & $\approx 0$ & \paramdn{32.240K} & \paramdn{68.080K} \\
    500M & 1280 & 24 & 3.12 & 0.0079 & $\approx 0$ & \paramdn{56.304K} & \paramdn{117.744K} \\
    1B & 2048 & 24 & 7.72 & 0.0051 & $\approx 0$ & \paramdn{90.096K} & \paramdn{188.400K} \\
    \bottomrule
    \end{tabular}
  \caption{Estimated Analytic training FLOPs of the Pre-Norm baseline, and the relative
  overhead (\%) and parameter reduction (\textcolor{good}{green}~$\downarrow$) of LayerRoPE over Pre-Norm and Peri-Norm. Over Peri-Norm, LayerRoPE adds only the negligible cost ($<10^{-4}\%$) of constructing the depth-conditioned $\gamma_{\ell}$, shown as $\approx 0$.}
  \label{tab:flop-overhead}
\end{table}

\clearpage

\section{Do Norm Weights Encode Depth across all LLMs?}
\label{app:norm_depth_all_llms}

We check whether this depth-conditioned encoding in the norm weights is a phenomenon observable across LLM families and architectures. We look at \textit{16 pre-trained open-weight LLMs across 9 families}, varying in Pre/Post/Peri-Norm formulations, and architectures (Dense, Mixture-of-Experts, hybrid linear attention, parallel MLP-Attention branches). As in Figure~\ref{fig:abu_norm_evidence}, we read the \emph{angle} of each layer's $\gamma$ from an MDS projection of the per-layer $\gamma$ vectors, and its \emph{magnitude} from the per-layer distribution of $\gamma$ values.
\noindent \textbf{Overview:} We start with Pre-Norm LLMs, their MLP Block norms (Appendix~\ref{app:norm_depth_prenorm_mlp}) which show clear angular and magnitude depth-conditioning, followed by their Attention Block norms (Appendix~\ref{app:norm_depth_prenorm_attn}). We then study Peri- and Post-Norm LLMs (Appendix~\ref{app:norm_depth_peri_post}) and observe the same depth-conditioning effects as Pre-Norm LLMs.

\subsection{Pre-Norm LLMs}
\label{app:norm_depth_prenorm_mlp}
\label{app:norm_depth_prenorm_attn}

We study open-weight Pre-Norm LLMs from seven model families, spanning various architectures. \textbf{Dense}: LLaMA-1/2~7B~\citep{touvron2023llama,touvron2023llama2}, LLaMA-3~8B~\citep{grattafiori2024llama}, Mistral~7B~v0.1~\citep{jiang2023mistral}, Qwen2/2.5~7B, Qwen3~8B~\citep{yang2025qwen3} and Gemma~7B~\citep{team2024gemma}; \textbf{Mixture-of-Experts} (GPT-OSS~20B~\citep{agarwal2025gpt}), \textbf{parallel branches: } Phi-2~2.7B and Pythia~2.8B~\citep{biderman2023pythia}; and \textbf{hybrid linear-attention: } Qwen3.5~9B~\citep{qwen35blog}.

\begin{figure}[p]
    \centering
    \captionsetup[figure]{position=top}
    \caption{\textbf{{(Pre-Norm, MLP Blocks)} Depth-conditioning of Norm Weights.} For each pre-trained model (\textit{\textbf{GPT-OSS~20B, Mistral~7B, Pythia~2.8B and LLaMA-3~8B}}), the distribution of MLP Block RMSNorm $\gamma$ values and MDS Projection of the same per-layer $\gamma$ vectors is shown. Note that in all cases, we see strong depth-conditioning in both layer-wise angular and magnitude properties of MLP norm weights.}
    \label{fig:app_mlp}
    \begin{subfigure}[b]{0.43\textwidth}
        \centering
        \includegraphics[width=\linewidth]{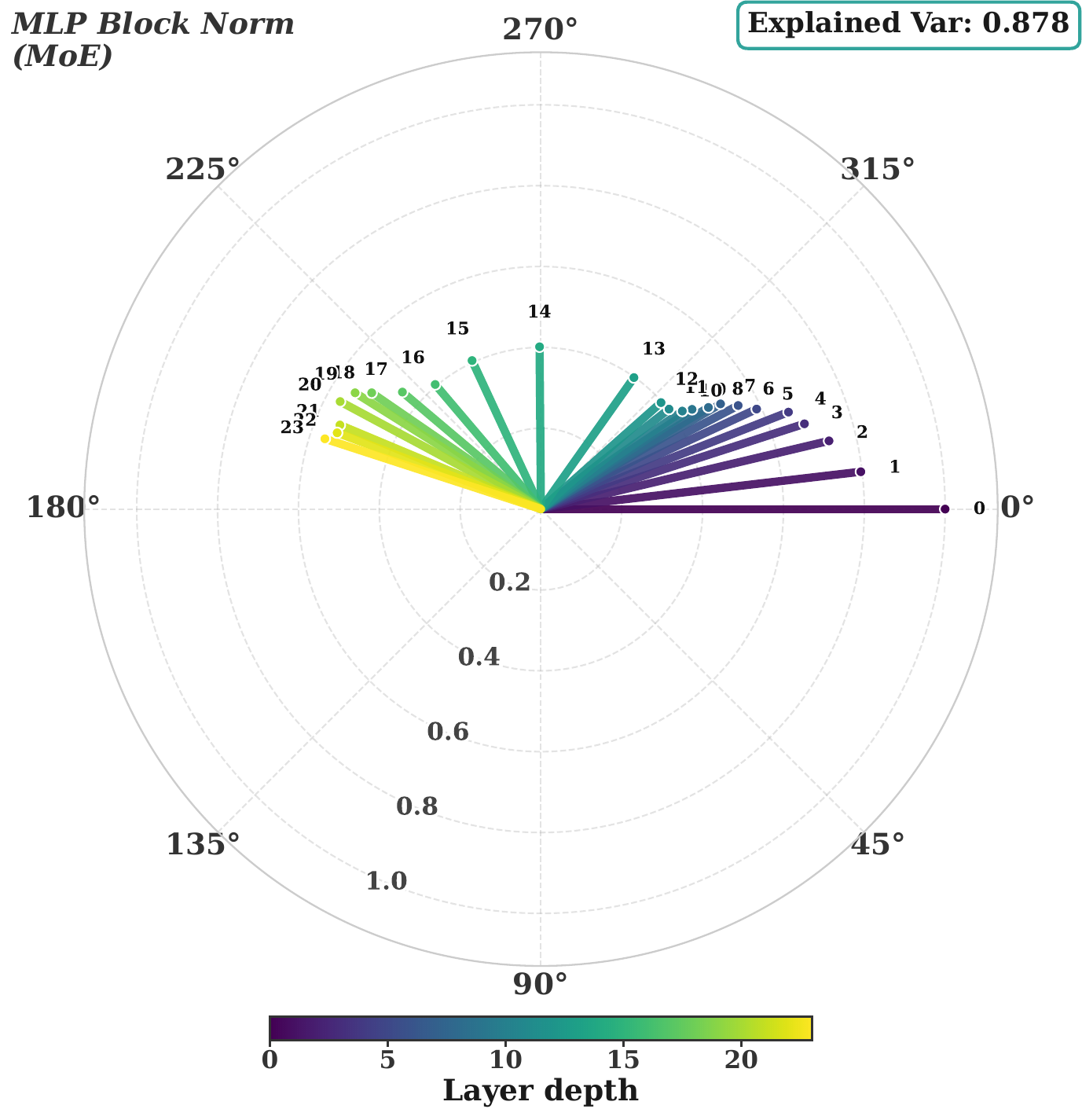}\\[2pt]
        \includegraphics[width=\linewidth]{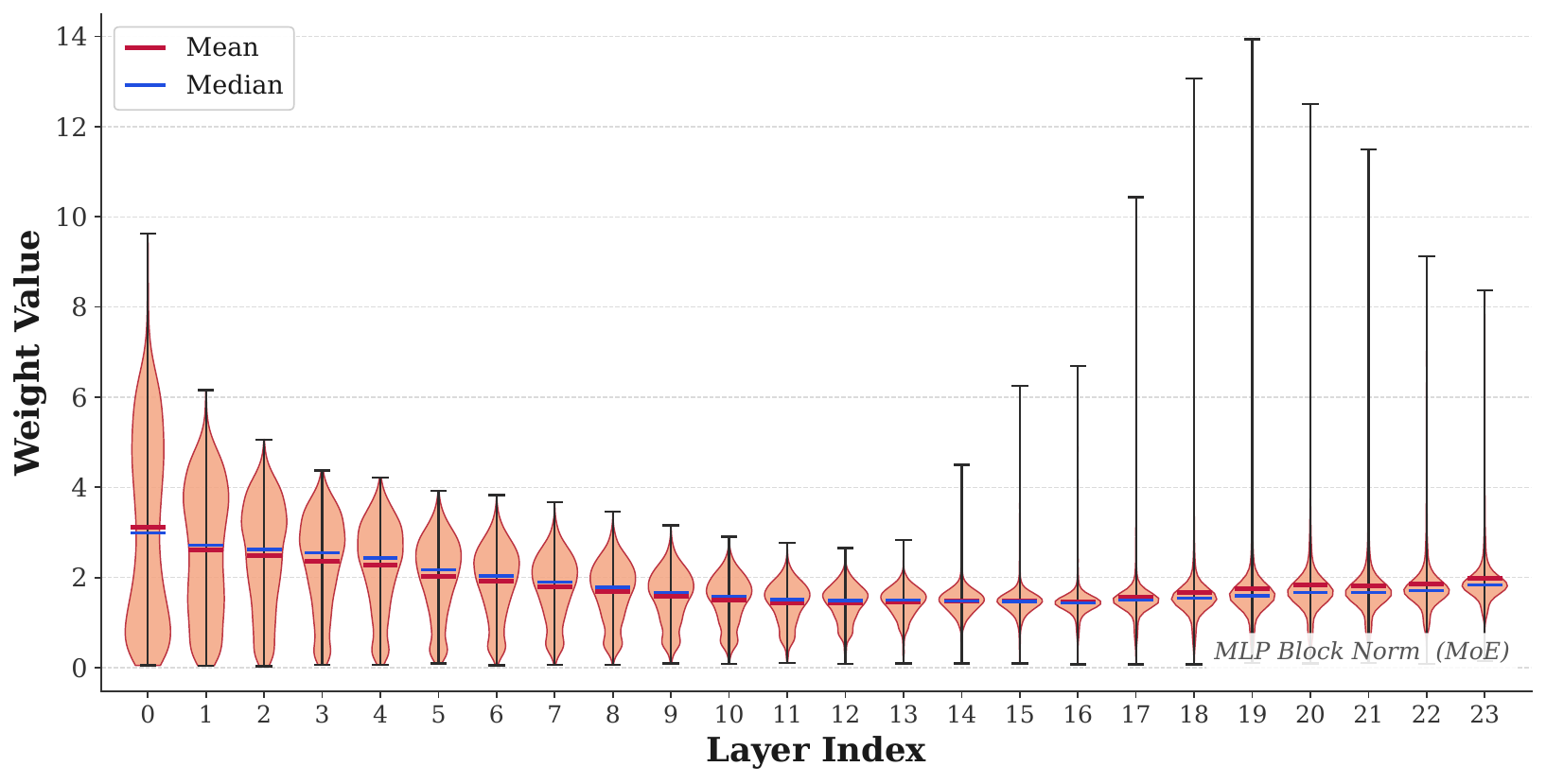}
        \caption{GPT-OSS~20B}
    \end{subfigure}\hspace{0.04\textwidth}%
    \begin{subfigure}[b]{0.43\textwidth}
        \centering
        \includegraphics[width=\linewidth]{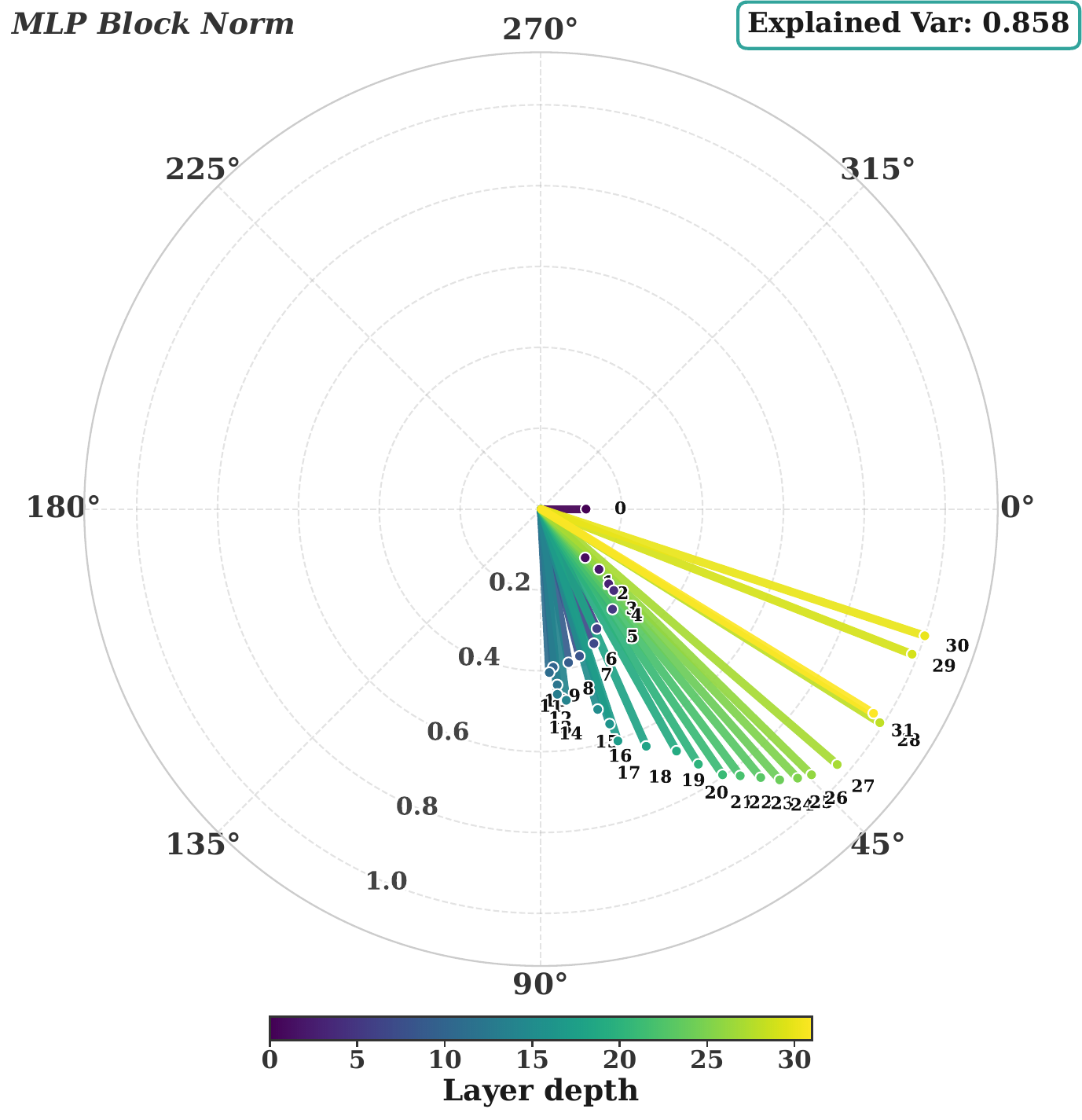}\\[2pt]
        \includegraphics[width=\linewidth]{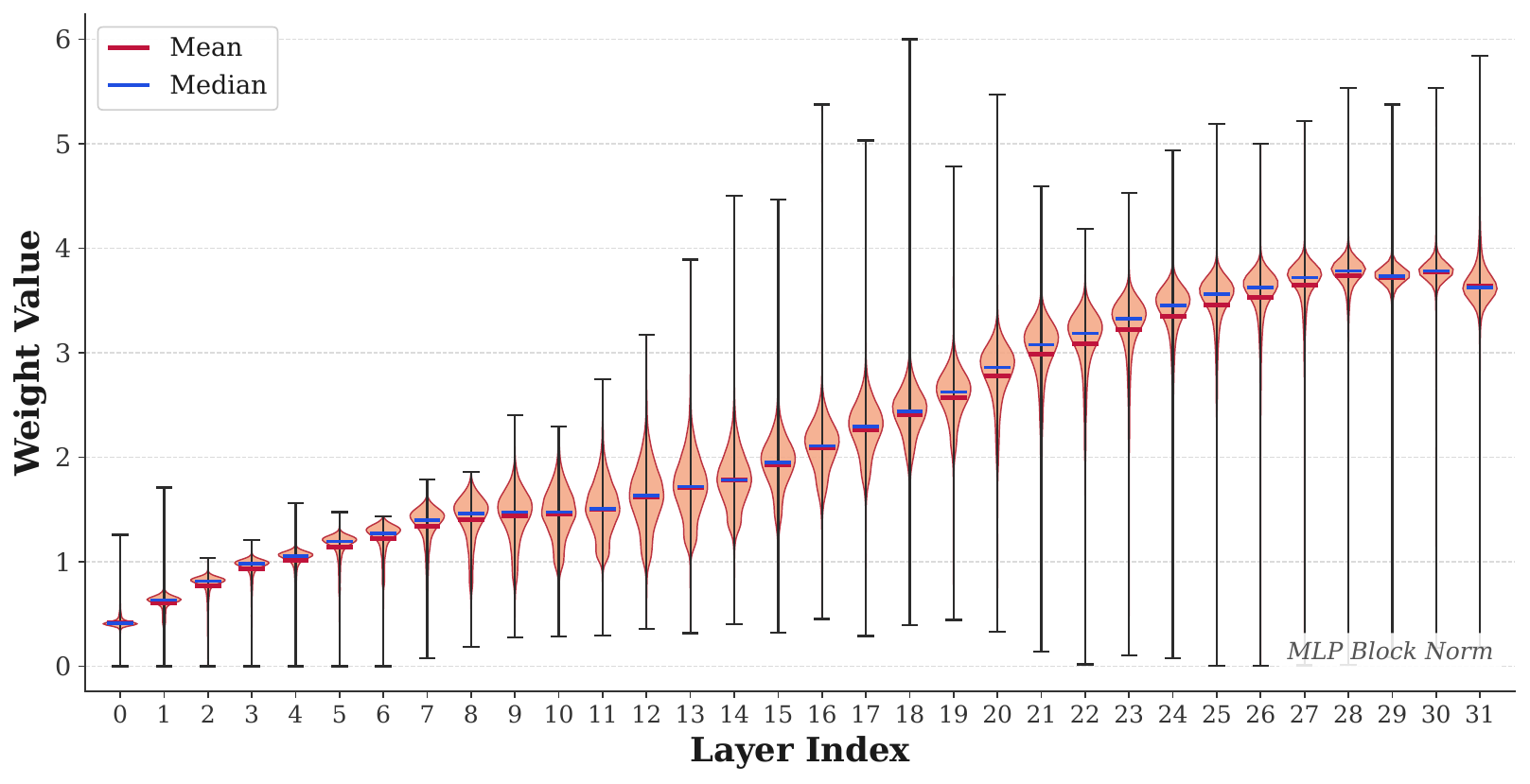}
        \caption{Mistral~7B}
    \end{subfigure}

    \medskip
    \begin{subfigure}[b]{0.43\textwidth}
        \centering
        \includegraphics[width=\linewidth]{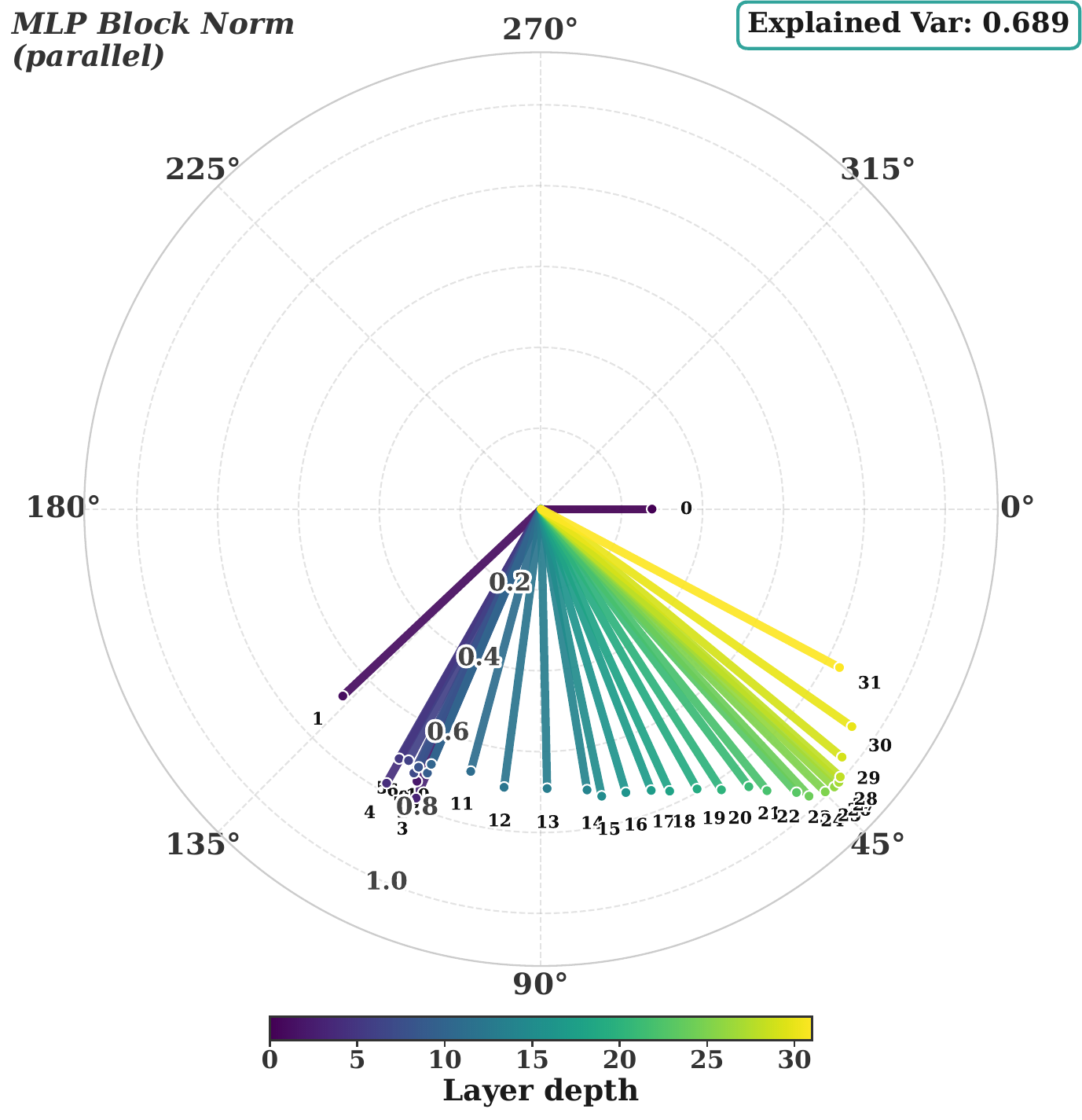}\\[2pt]
        \includegraphics[width=\linewidth]{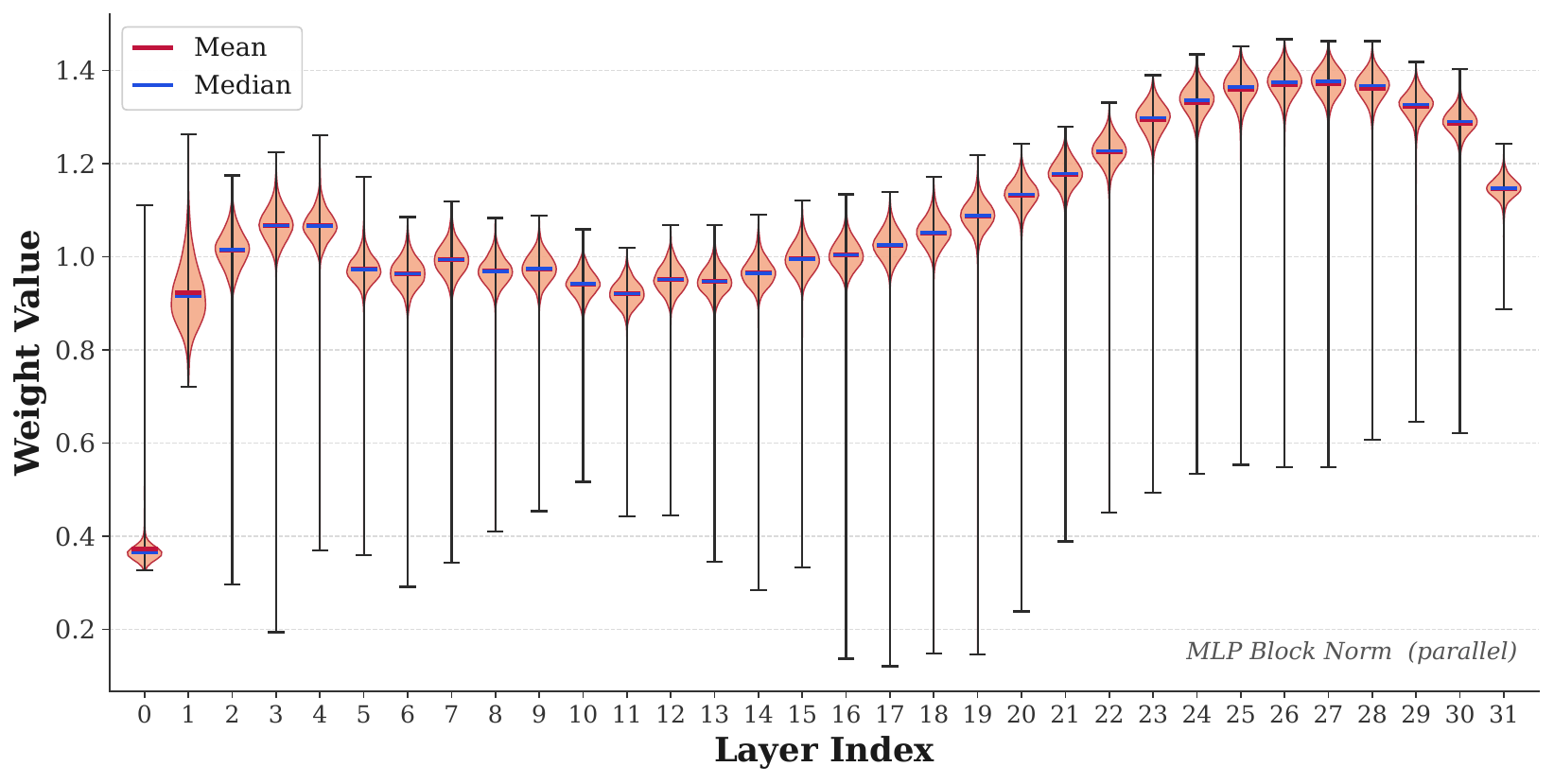}
        \caption{Pythia~2.8B}
    \end{subfigure}\hspace{0.04\textwidth}%
    \begin{subfigure}[b]{0.43\textwidth}
        \centering
        \includegraphics[width=\linewidth]{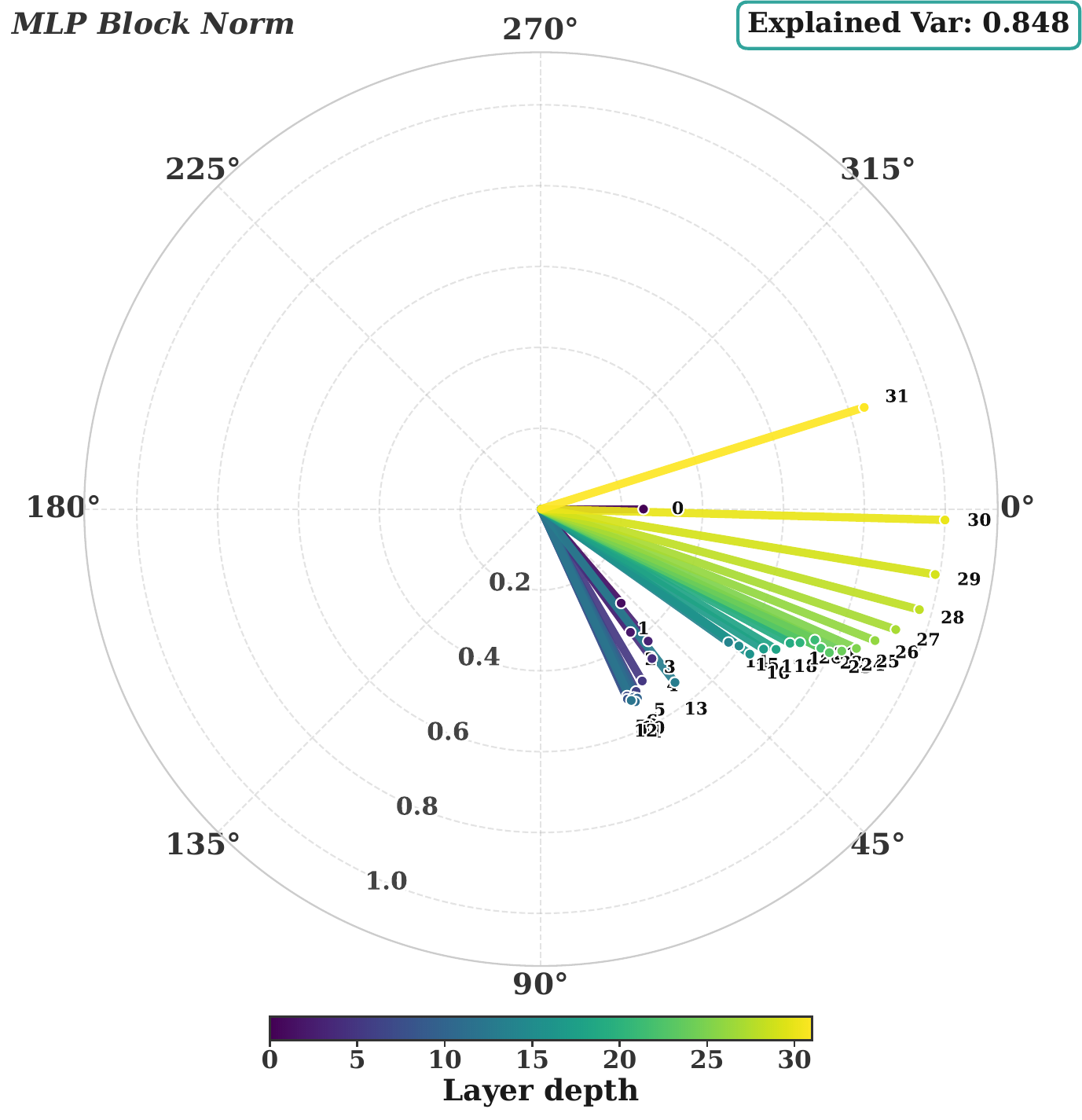}\\[2pt]
        \includegraphics[width=\linewidth]{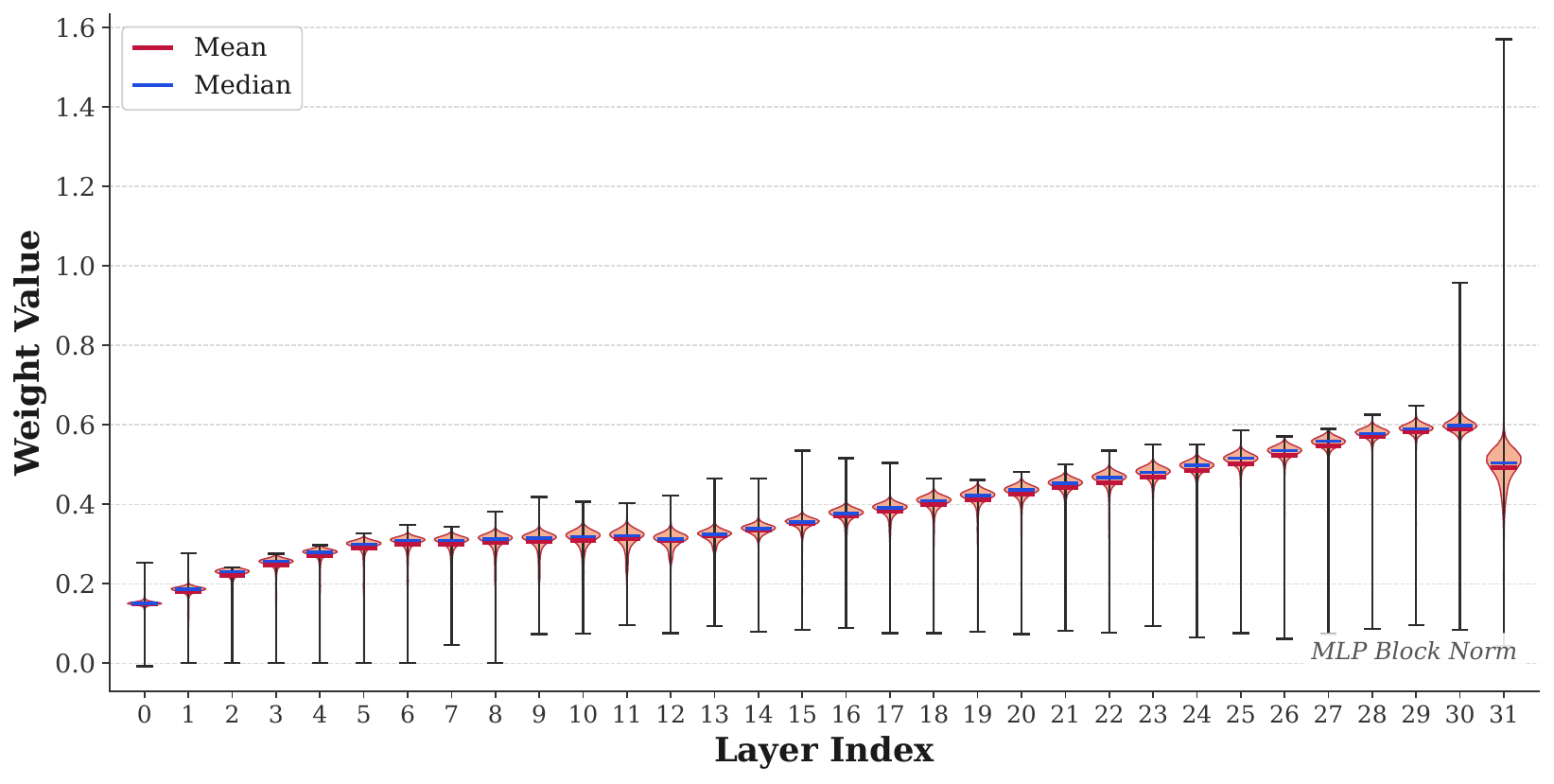}
        \caption{LLaMA-3~8B}
    \end{subfigure}
\end{figure}

\begin{figure}[t]
    \ContinuedFloat
    \centering
    \captionsetup[figure]{position=top}
    \caption[]{\textbf{{(Pre-Norm, MLP Blocks. (continued) )} Depth-conditioning of Norm Weights.} For each pre-trained model (\textit{\textbf{LLaMA-1~7B, Qwen3~8B and Phi-2~2.7B}}), we continue to see strong depth-conditioning in both layer-wise angular and magnitude properties of the MLP norm weights. We note Qwen2.5, Qwen3.5 and Gemma (highlighted in the {\color{orange}boxed} row), unlike all other Pre-Norm models, do not show a strong depth conditioning effect in their MLP Norm weights.}
    \begin{minipage}{\dimexpr\textwidth-10pt\relax}
        \begin{subfigure}[b]{0.31\linewidth}
            \centering
            \includegraphics[width=\linewidth]{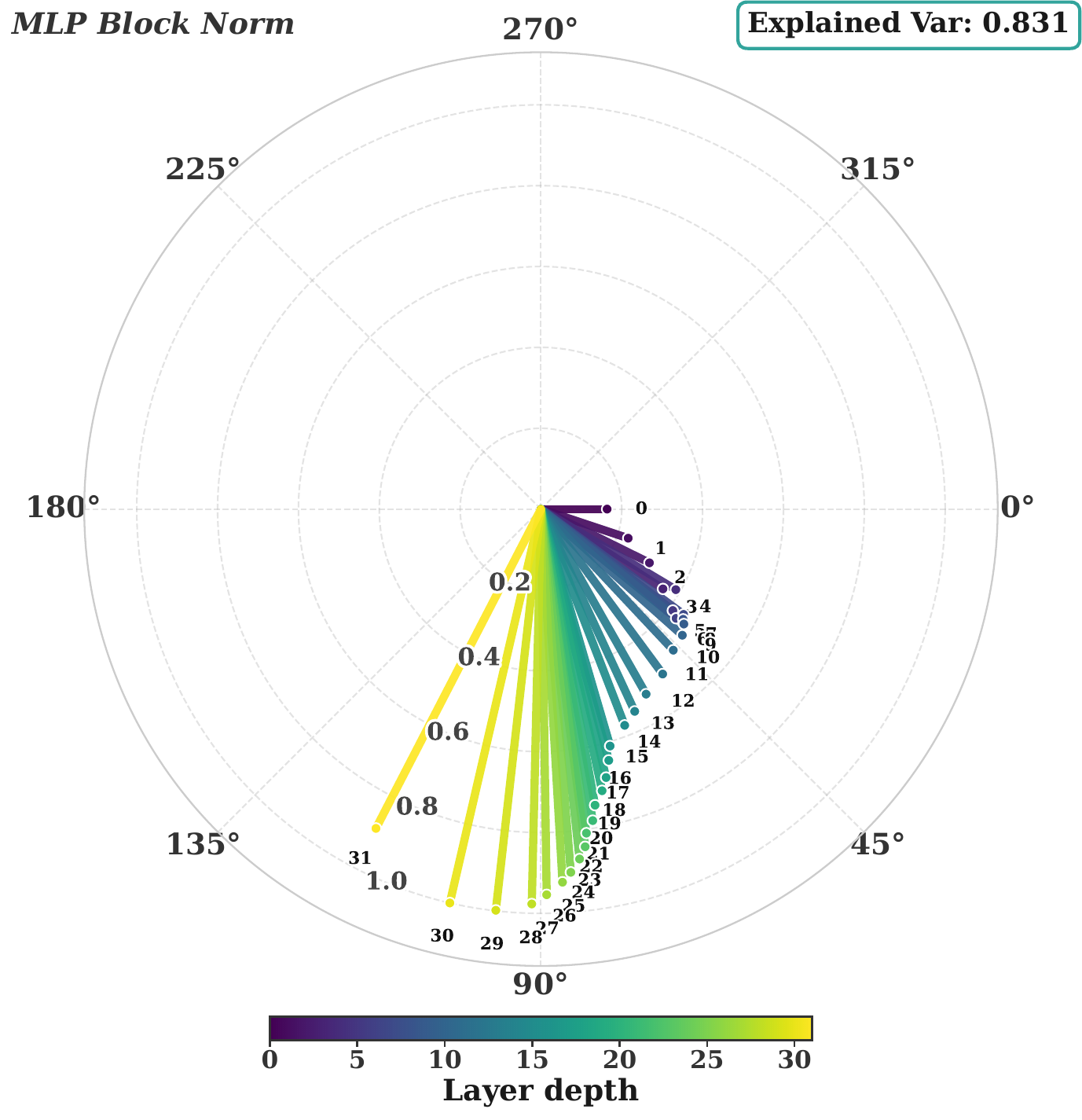}\\[2pt]
            \includegraphics[width=\linewidth]{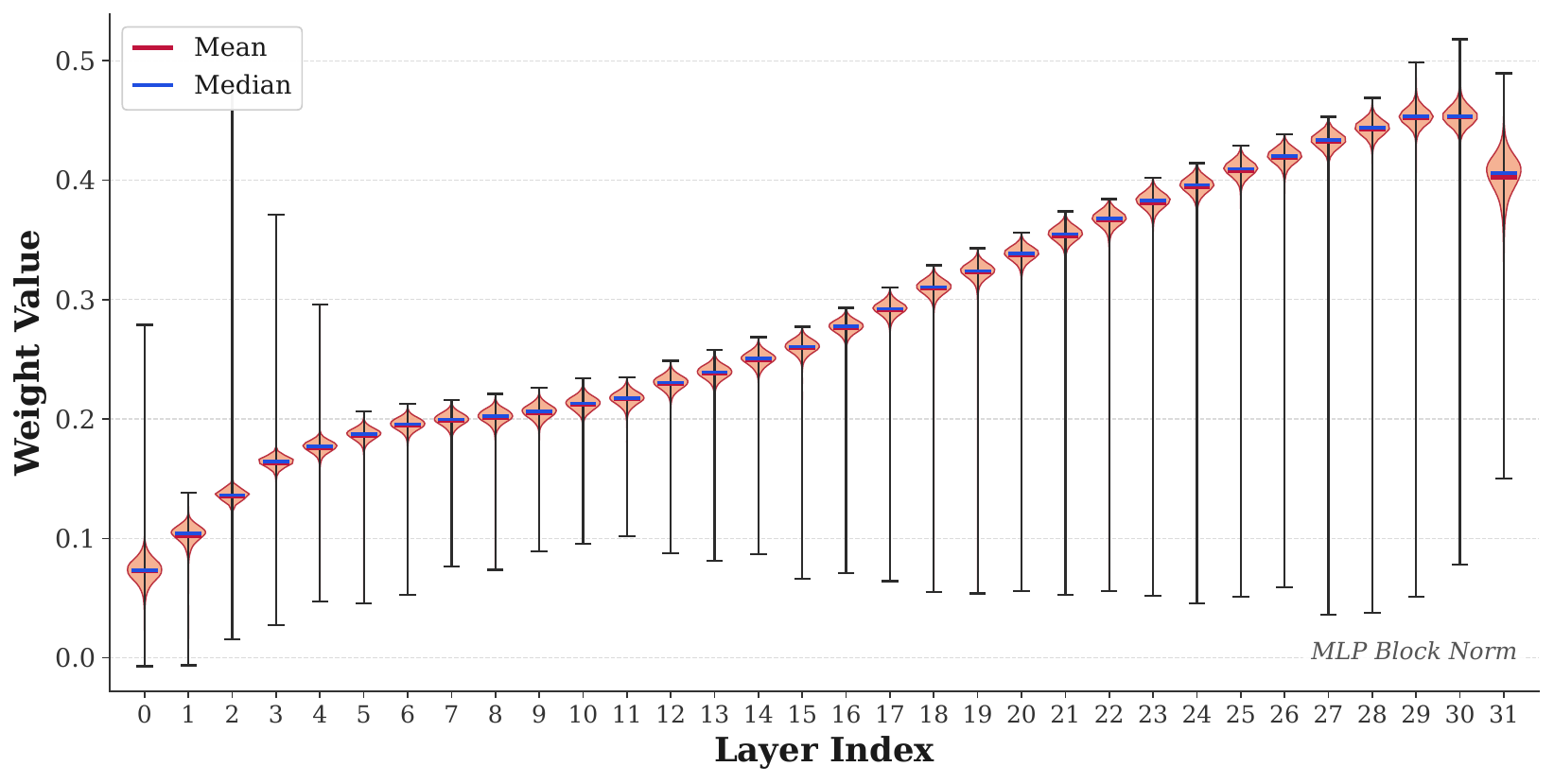}
            \caption{LLaMA-1~7B}
        \end{subfigure}\hfill
        \begin{subfigure}[b]{0.31\linewidth}
            \centering
            \includegraphics[width=\linewidth]{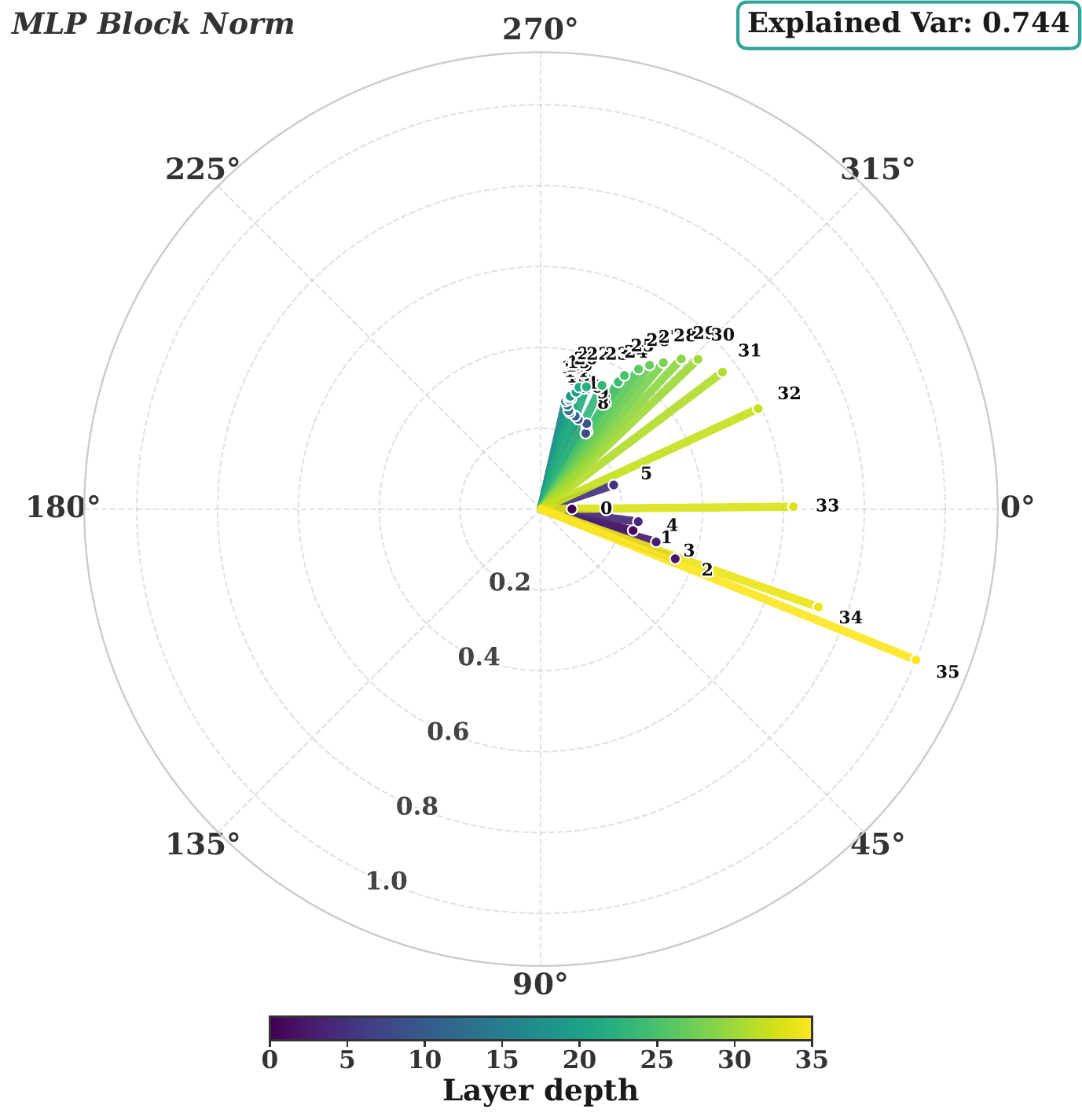}\\[2pt]
            \includegraphics[width=\linewidth]{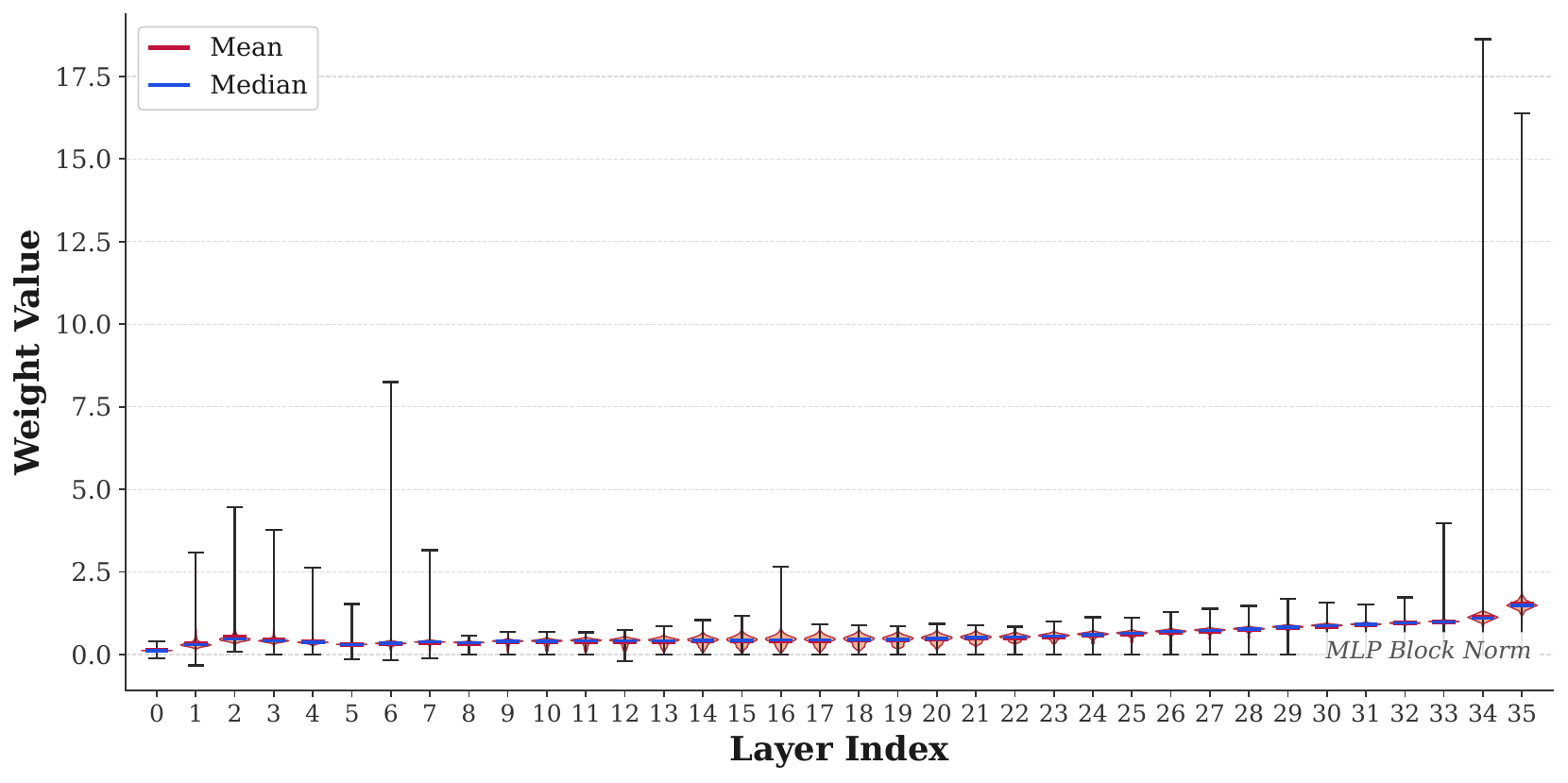}
            \caption{Qwen3~8B}
        \end{subfigure}\hfill
        \begin{subfigure}[b]{0.31\linewidth}
            \centering
            \includegraphics[width=\linewidth]{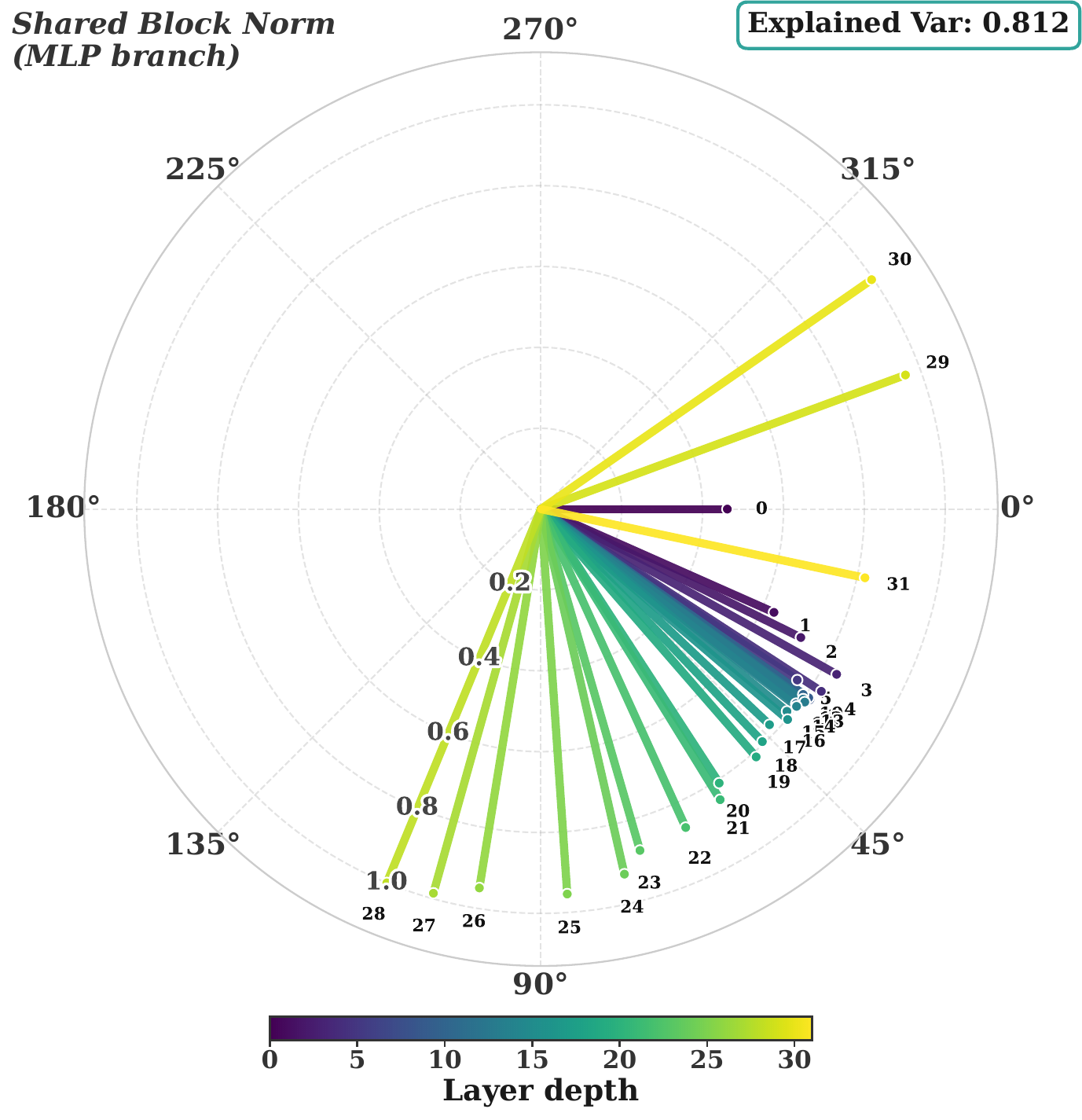}\\[2pt]
            \includegraphics[width=\linewidth]{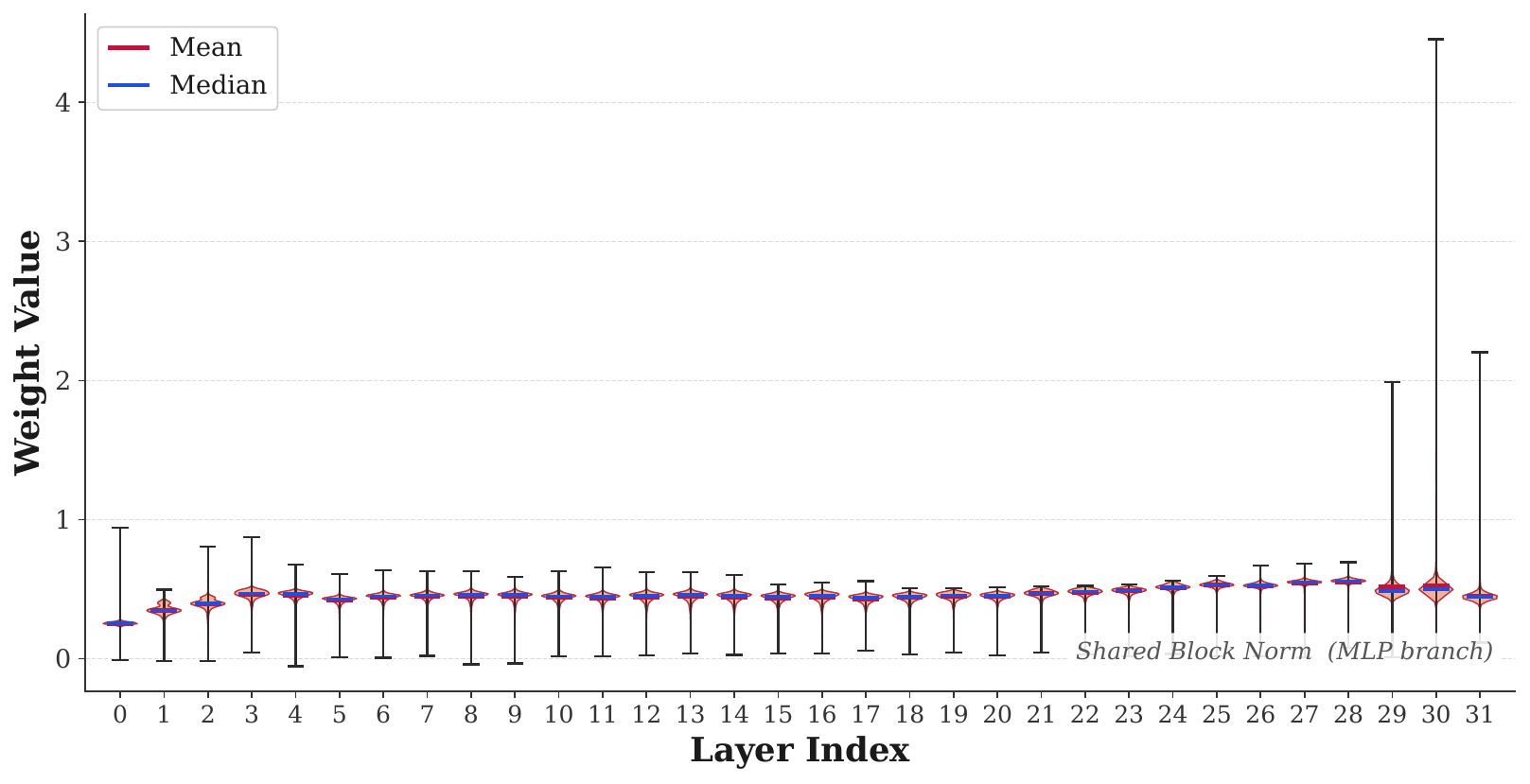}
            \caption{Phi-2~2.7B}
        \end{subfigure}
    \end{minipage}

    \medskip
    \begin{tcolorbox}[colframe=cuteorange, colback=white, arc=6pt, boxrule=1pt,
                      boxsep=0pt, left=4pt, right=4pt, top=4pt, bottom=2pt]
        \begin{subfigure}[b]{0.31\linewidth}
            \centering
            \includegraphics[width=\linewidth]{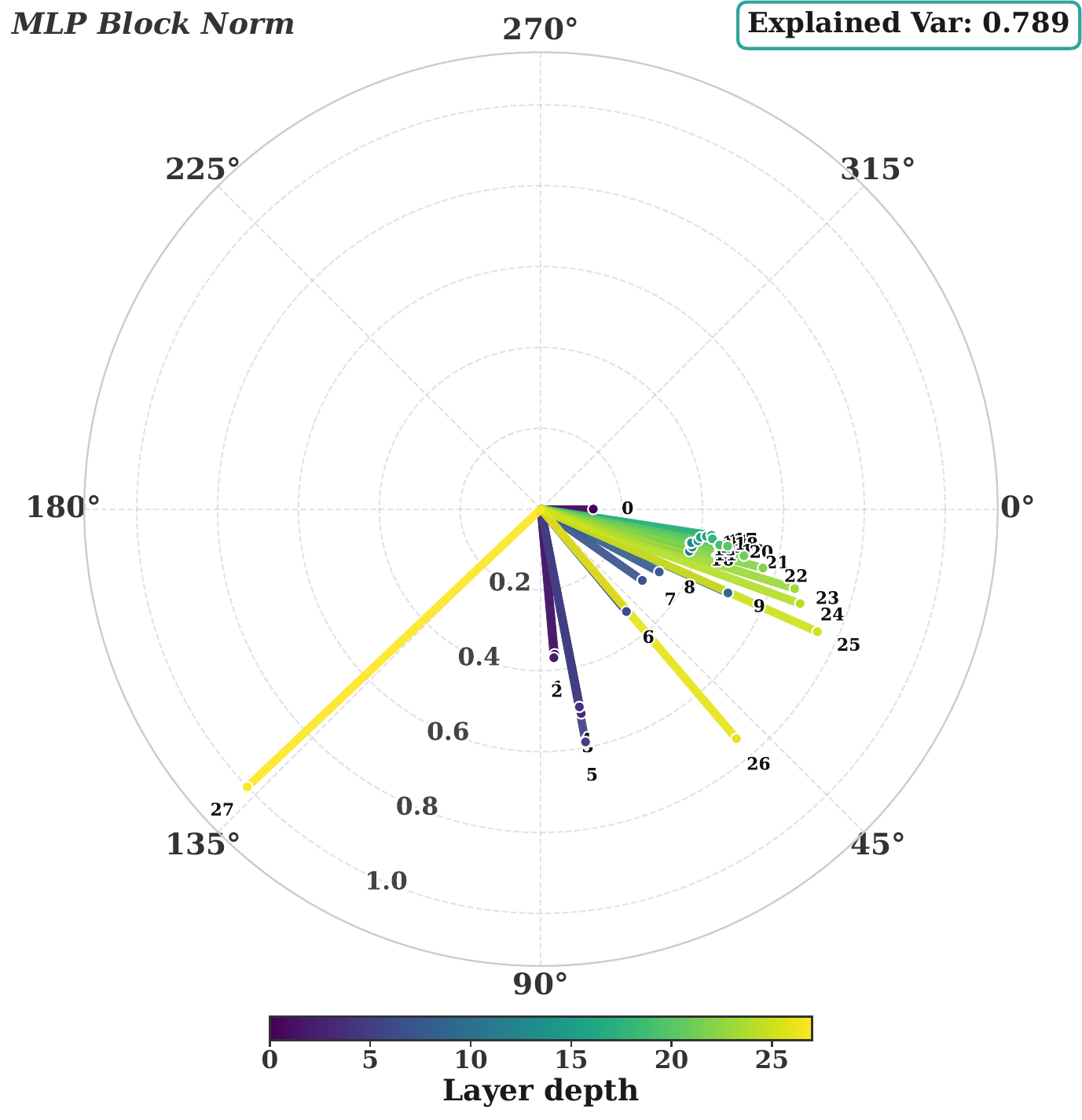}\\[2pt]
            \includegraphics[width=\linewidth]{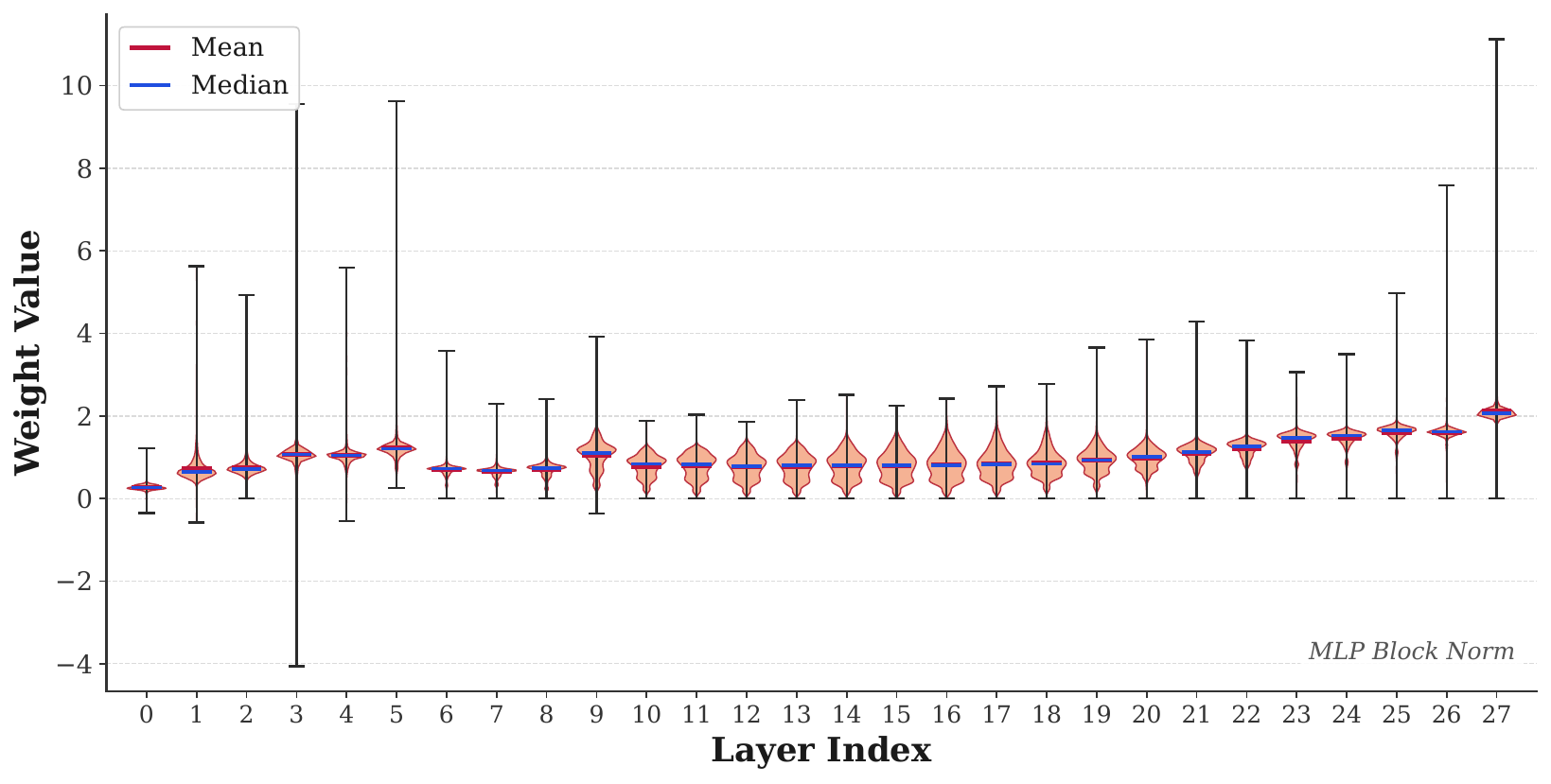}
            \caption{Qwen2.5~7B}
        \end{subfigure}\hfill
        \begin{subfigure}[b]{0.31\linewidth}
            \centering
            \includegraphics[width=\linewidth]{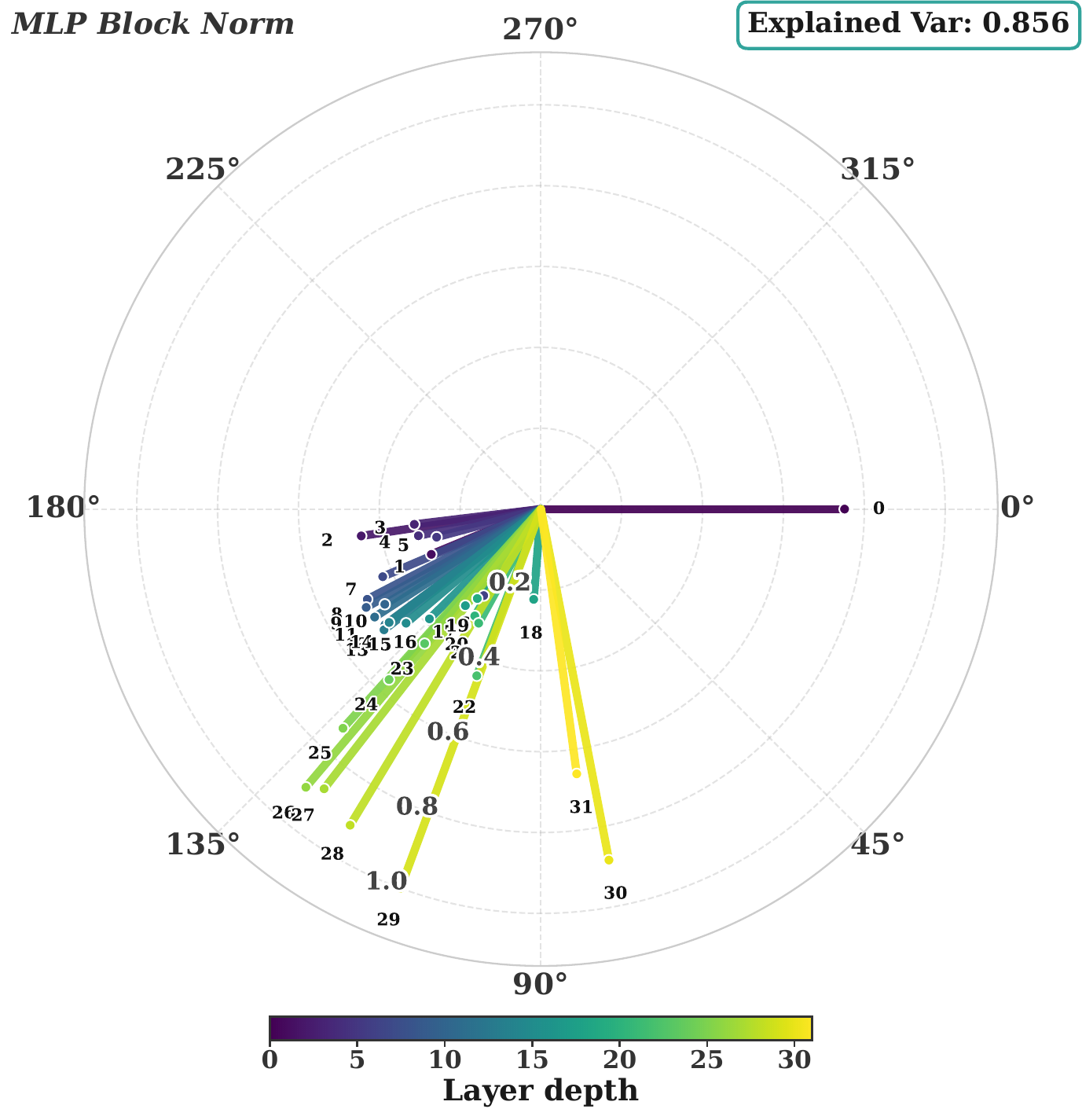}\\[2pt]
            \includegraphics[width=\linewidth]{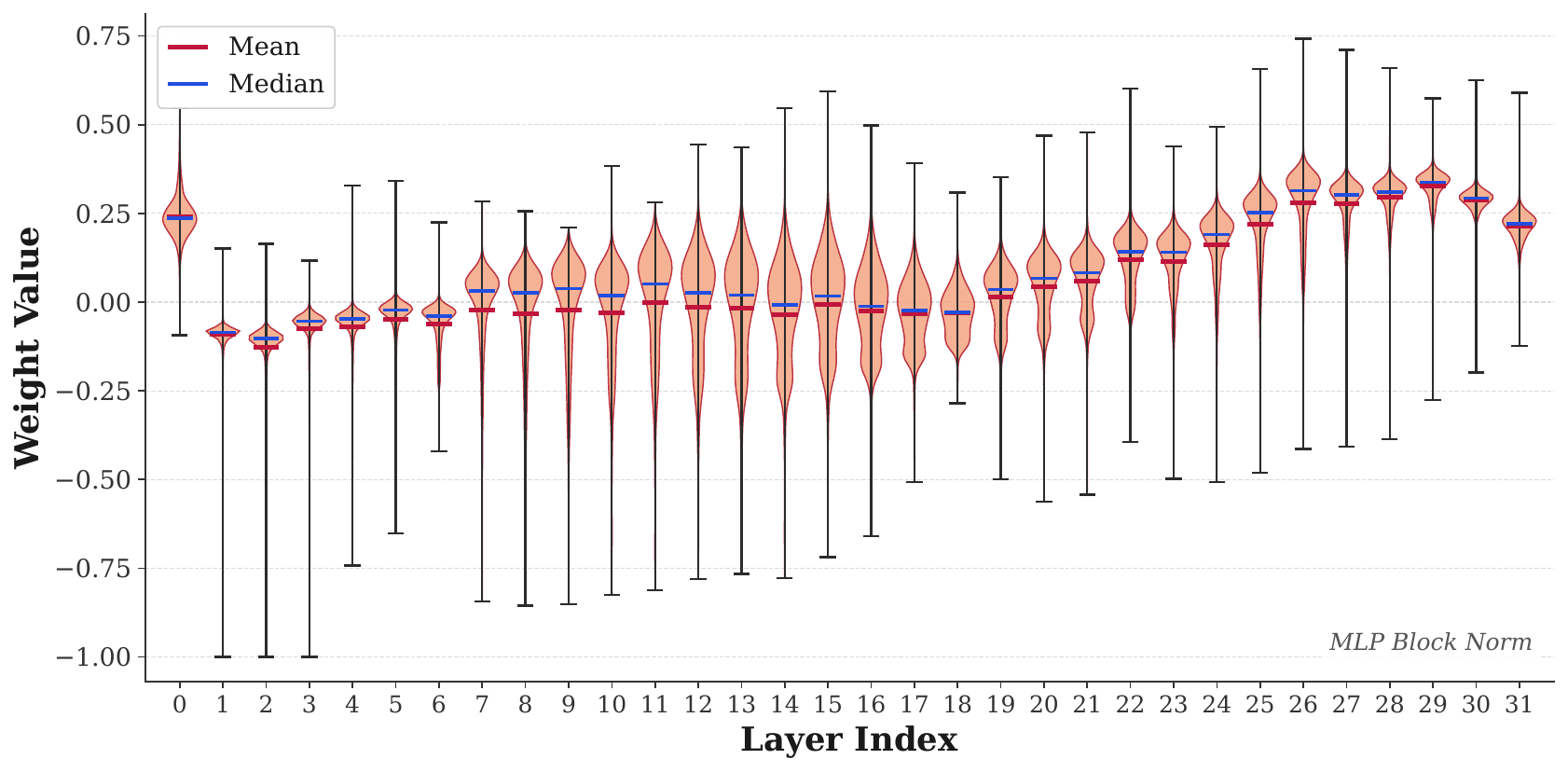}
            \caption{Qwen3.5~9B (offsets $\gamma$)}
        \end{subfigure}\hfill
        \begin{subfigure}[b]{0.31\linewidth}
            \centering
            \includegraphics[width=\linewidth]{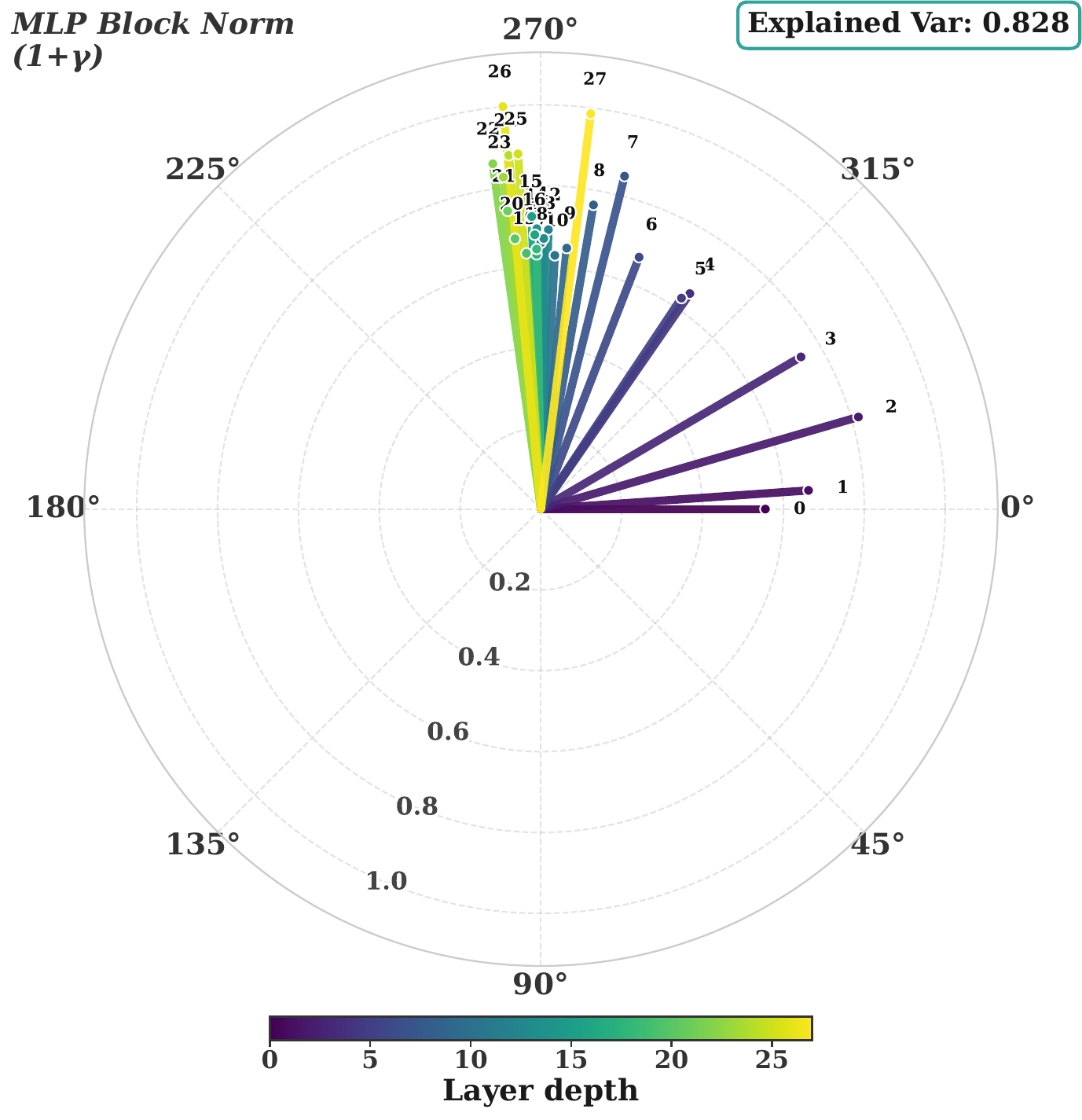}\\[2pt]
            \includegraphics[width=\linewidth]{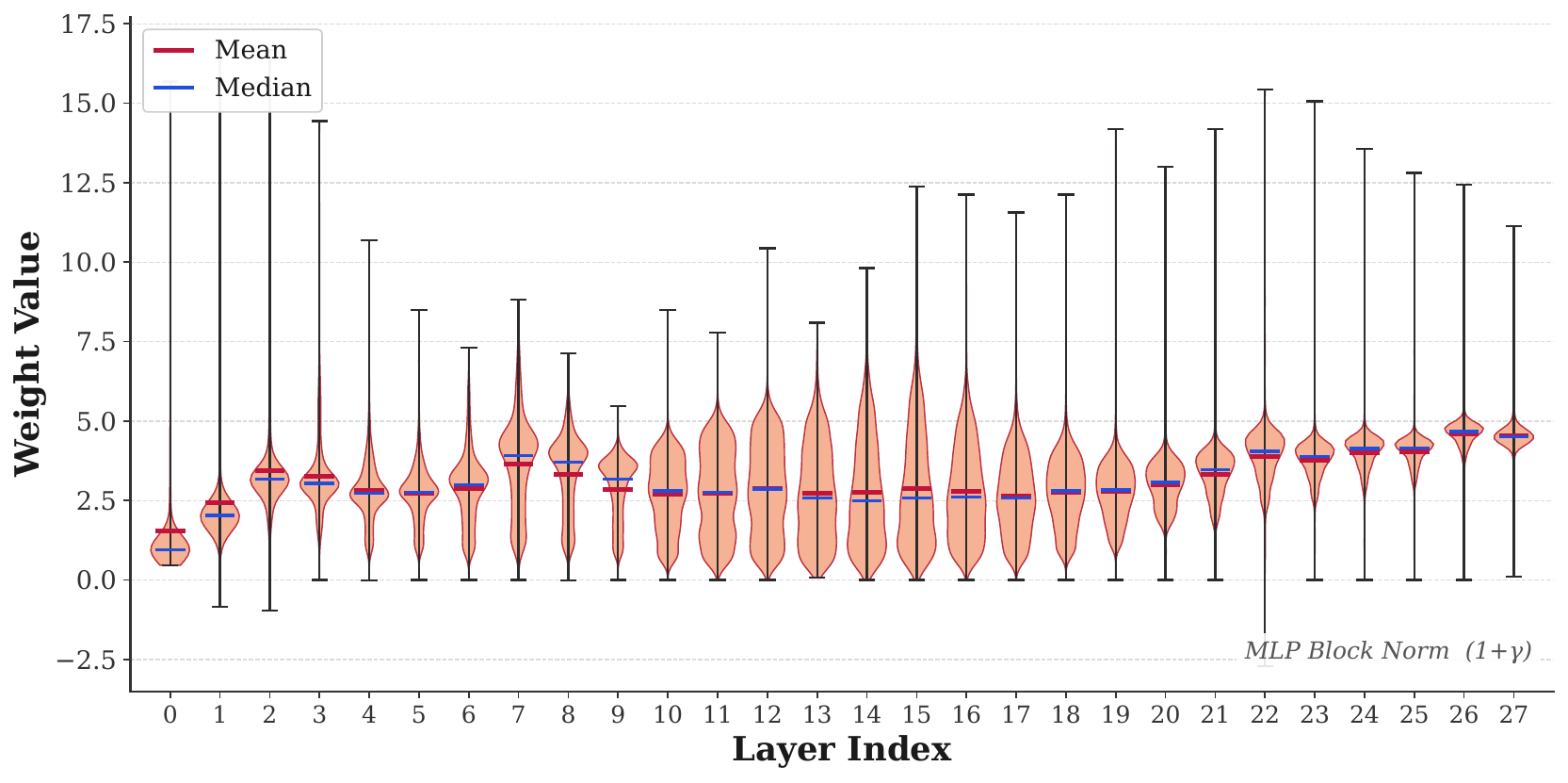}
            \caption{Gemma~7B ($1+\gamma$)}
        \end{subfigure}
    \end{tcolorbox}
\end{figure}

\begin{figure}[p]
    \centering
    \begin{subfigure}[b]{0.44\textwidth}
        \centering
        \includegraphics[width=\linewidth]{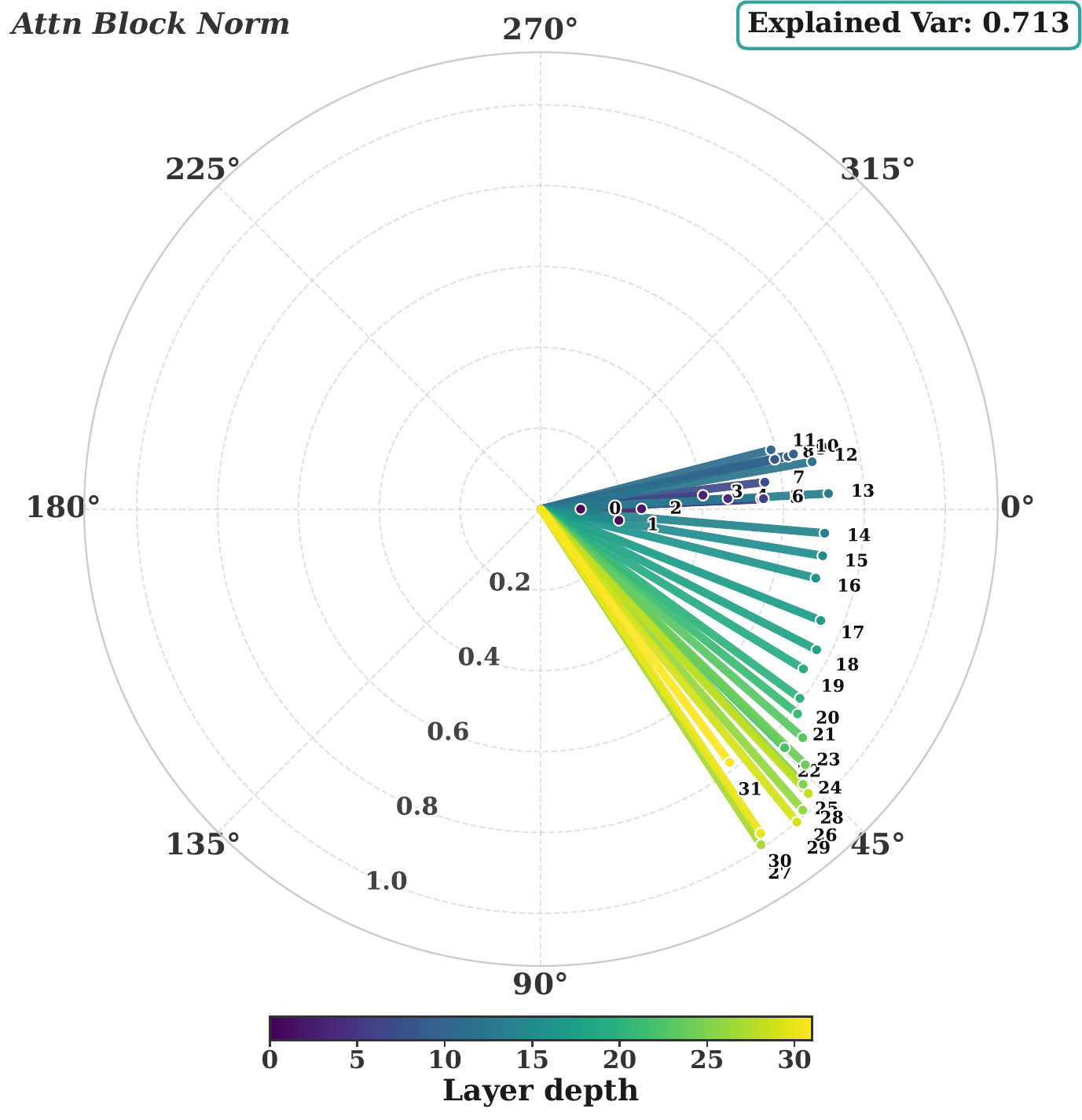}\\[2pt]
        \includegraphics[width=\linewidth]{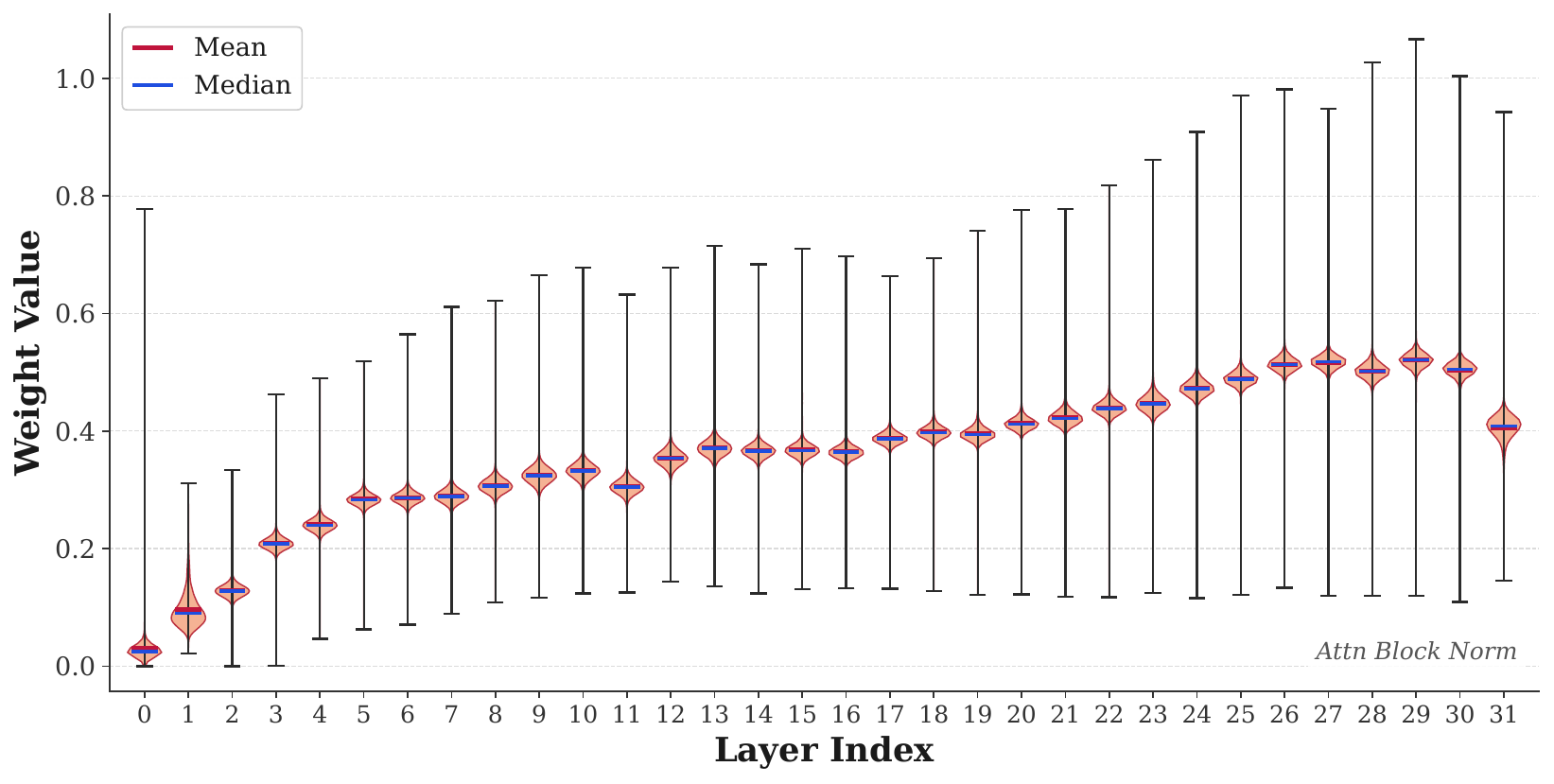}
        \caption{LLaMA-1~7B}
    \end{subfigure}\hspace{0.04\textwidth}%
    \begin{subfigure}[b]{0.44\textwidth}
        \centering
        \includegraphics[width=\linewidth]{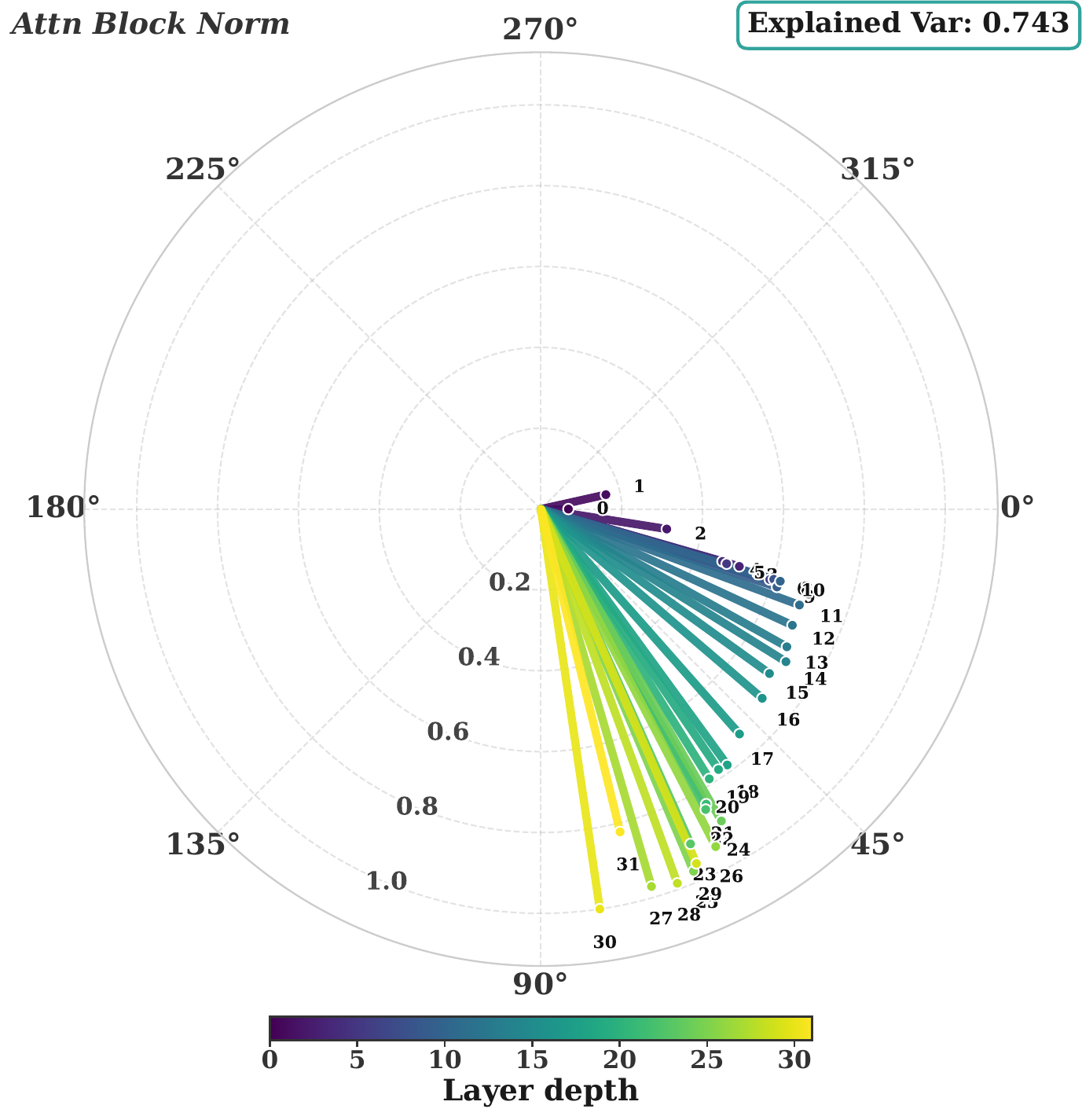}\\[2pt]
        \includegraphics[width=\linewidth]{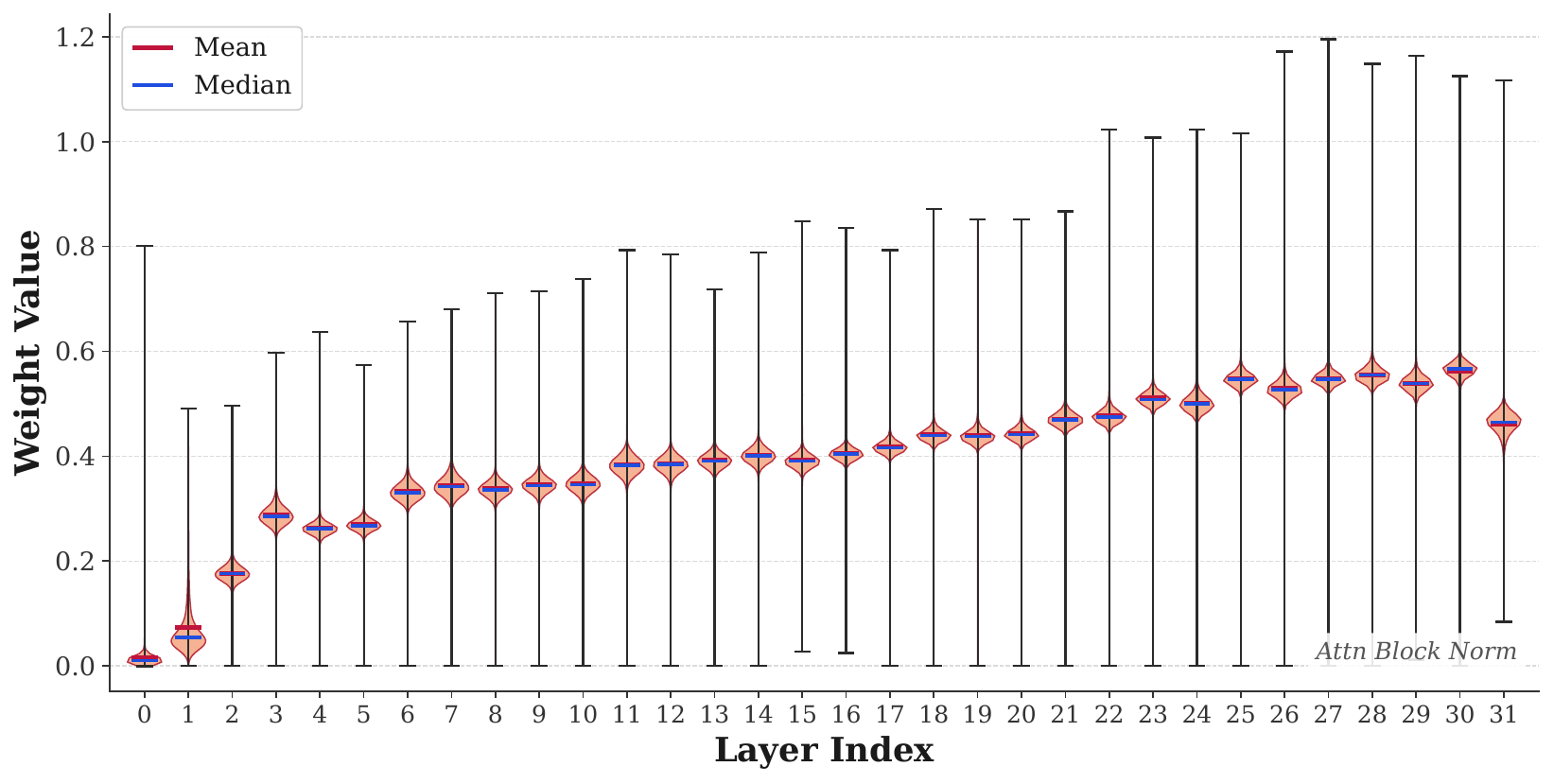}
        \caption{LLaMA-2~7B}
    \end{subfigure}

    \medskip
    \begin{subfigure}[b]{0.44\textwidth}
        \centering
        \includegraphics[width=\linewidth]{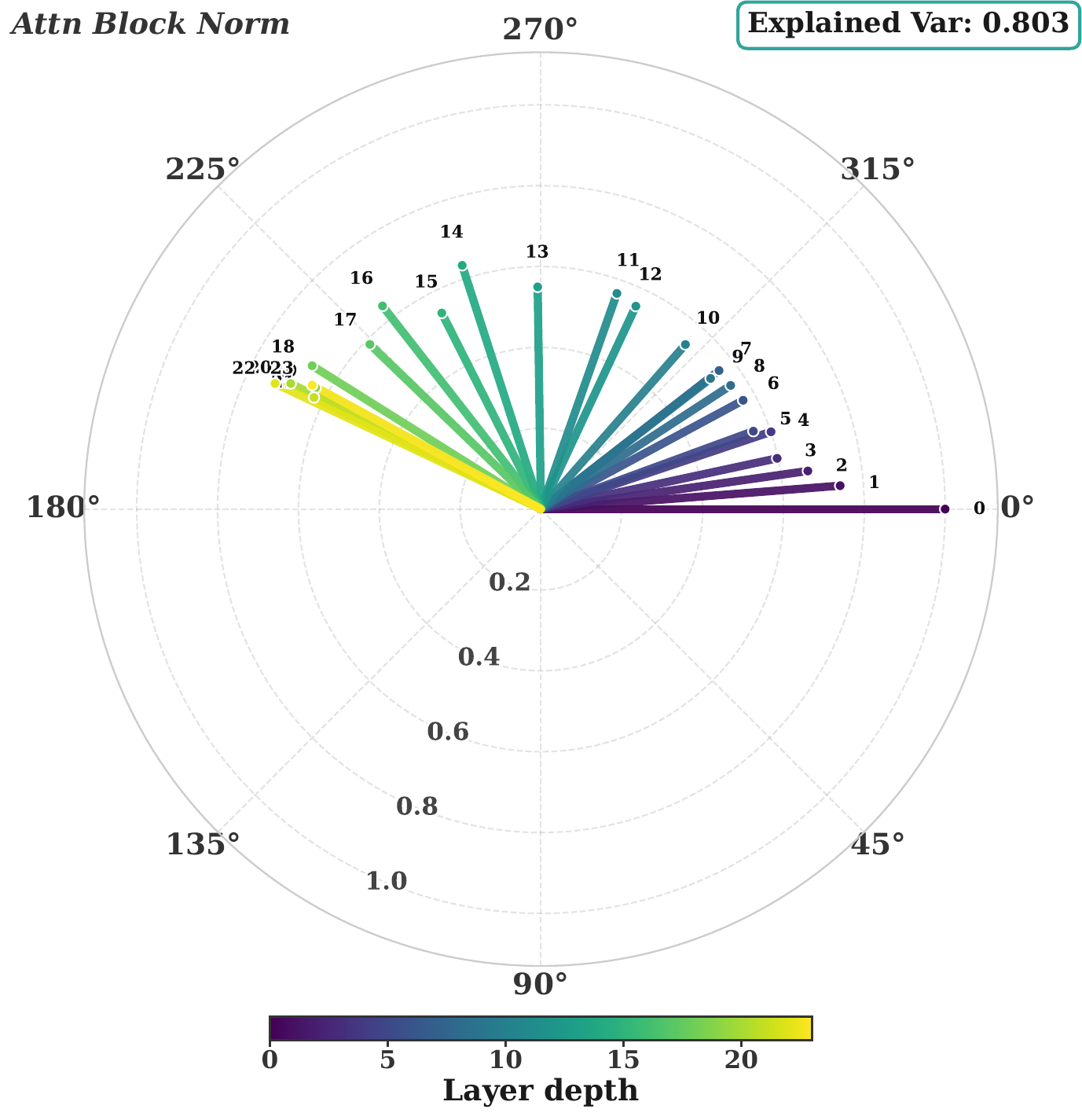}\\[2pt]
        \includegraphics[width=\linewidth]{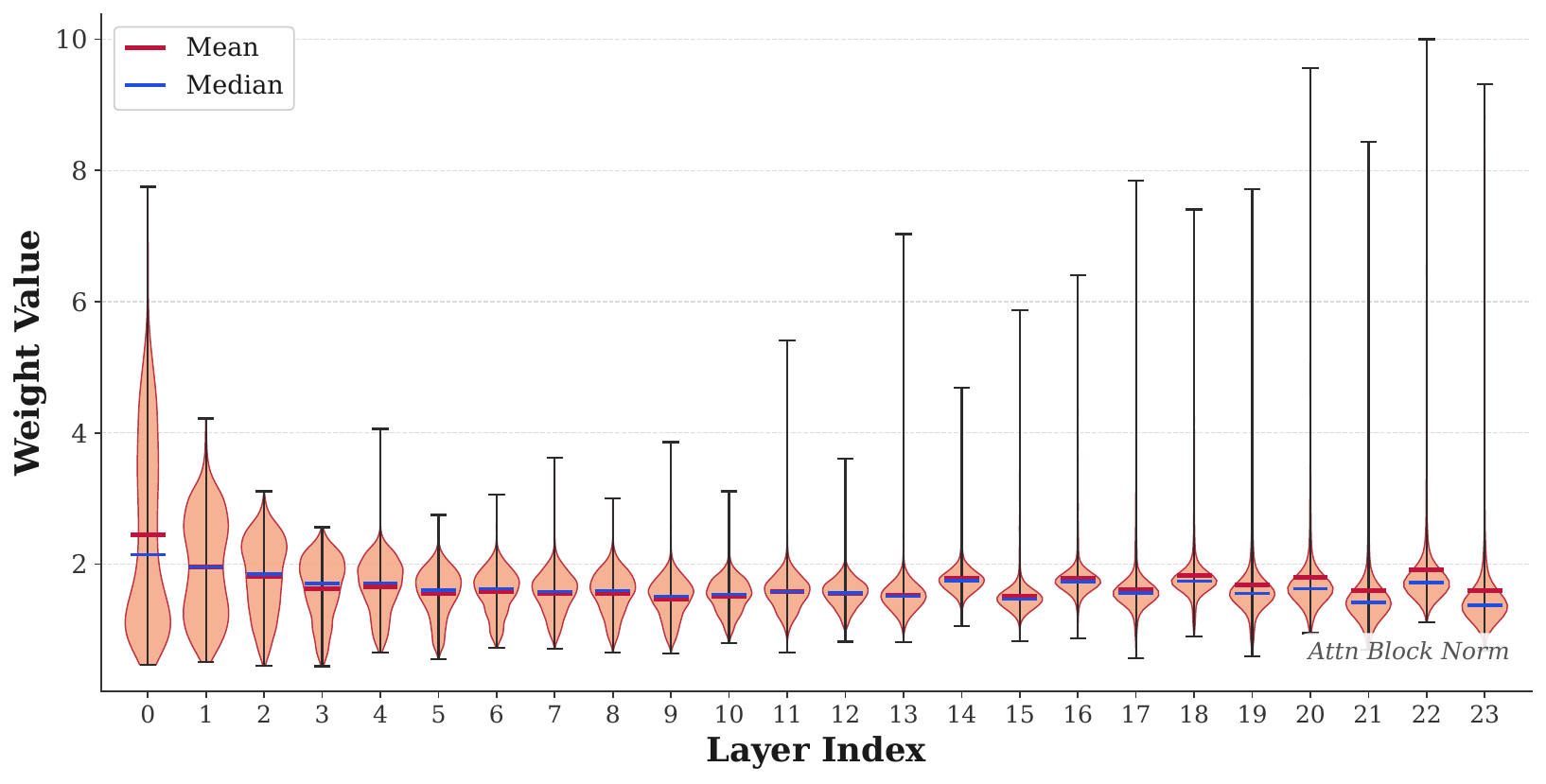}
        \caption{GPT-OSS~20B}
    \end{subfigure}\hspace{0.04\textwidth}%
    \begin{subfigure}[b]{0.44\textwidth}
        \centering
        \includegraphics[width=\linewidth]{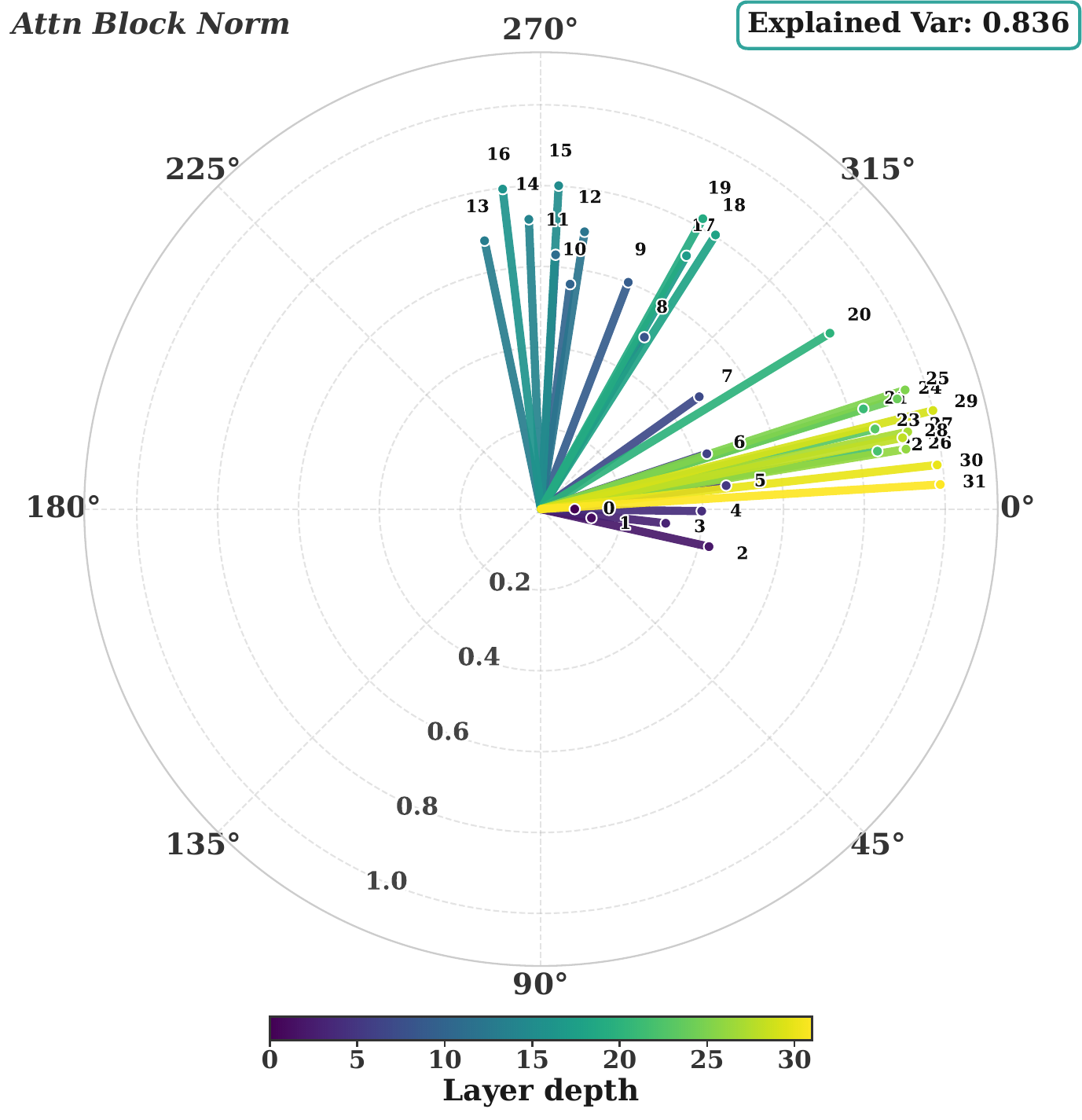}\\[2pt]
        \includegraphics[width=\linewidth]{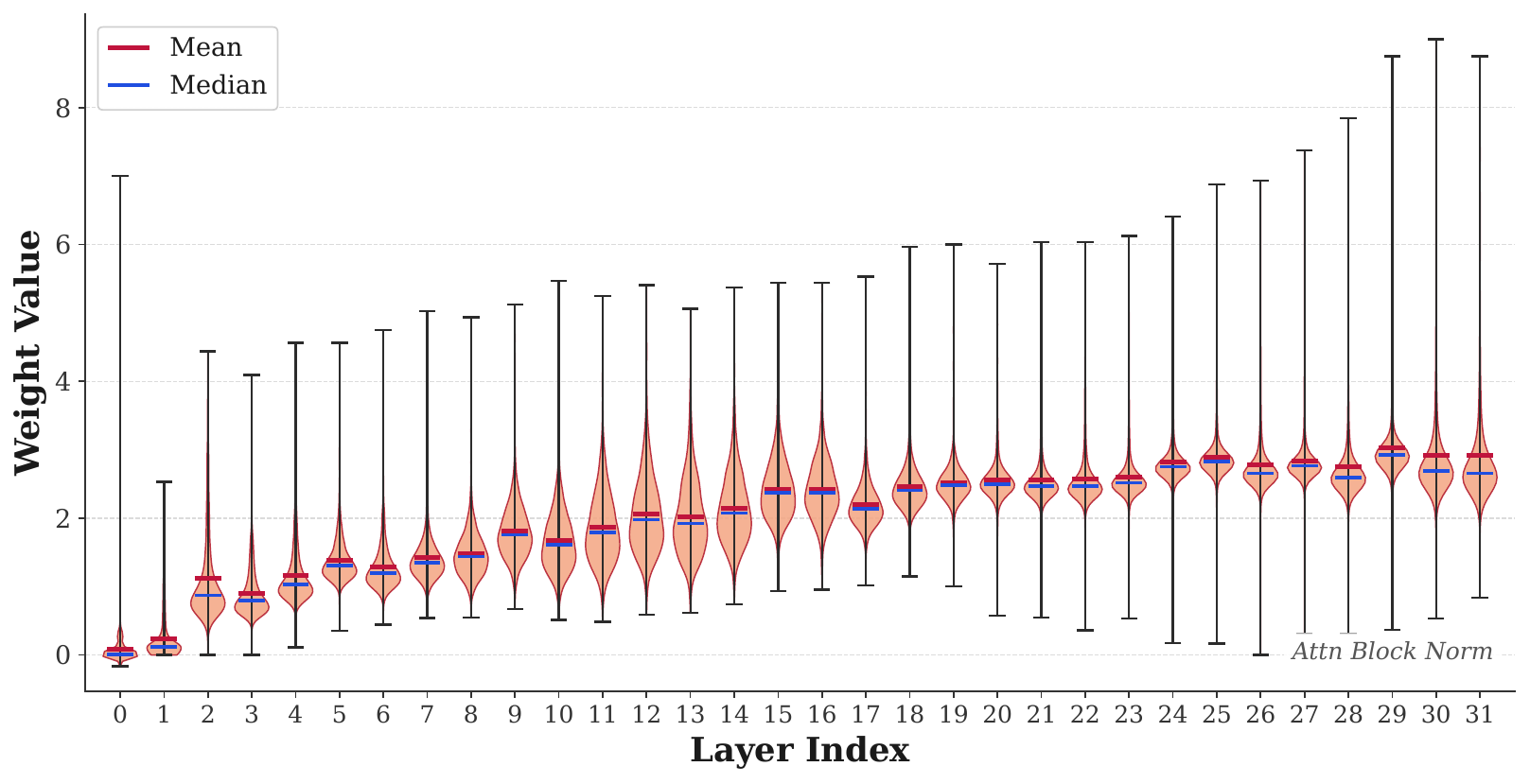}
        \caption{Mistral~7B}
    \end{subfigure}
    \caption{\textbf{Depth encoding in the Attention block norms of Pre-Norm LLMs.} For each model, \emph{top}: MDS projection of the per-layer Attention block $\gamma$ vectors (angle); \emph{bottom}: per-layer distribution of the same $\gamma$ values (magnitude). Compare with the MLP block norms of the same models in Figure~\ref{fig:app_mlp} (LLaMA-2's in Figure~\ref{fig:abu_norm_evidence}).}
    \label{fig:app_attn}
\end{figure}

\textbf{Inferences}: Norm weight analysis plots of MLP blocks are shown in Figure~\ref{fig:app_mlp}, and the Attention blocks are in Figure \ref{fig:app_attn}: 
\begin{itemize}
    \item \textbf{Angular encoding.} Figure~\ref{fig:app_mlp} shows the MDS (Multi-dimensional scaling) plots with successive layers occupying neighboring angles, and showing a clear depth-ordered trajectory. LLaMA-1, GPT-OSS and Phi-2 for instance, trace a single near-monotone arc from the first layer, as LLaMA-2 does in Figure~\ref{fig:abu_norm_evidence}. 
    \item \textbf{Magnitude encoding.} Norm weight distributions exhibit the depth conditioning effect as well, with the LLaMA models, Mistral and even Qwen3 reproduce a near monotonic change with network depth (as in LLAMA-2 in Figure~\ref{fig:abu_norm_evidence}). Other models with more varying depth-wise patterns, though, such as in Phi-2, where magnitude effects are unclear, the angular effects are particular clear, suggesting that angle and magnitude act as complementary carriers of depth. 
    \item \textbf{Outliers.} The effects are not perfect across all Pre-Norm LLMs. Firstly, we can see that in all cases, the norm weights in the Attention Blocks do not show any clear depth conditioning, either angular or magnitude. In the three boxed models Qwen 2.5/3.5 \& Gemma 7B, the MLP block norms also do not clearly encode depth. Additionally, there are some general nuances in the overall trend, with the initial and final layers often breaking from pattern or showing more varied distributions. 
\end{itemize}

\subsection{Peri- and Post-Norm LLMs}
\label{app:norm_depth_peri_post}

\begin{figure}[p]
    \centering
    \begin{subfigure}[b]{0.43\textwidth}
        \centering
        \includegraphics[width=\linewidth]{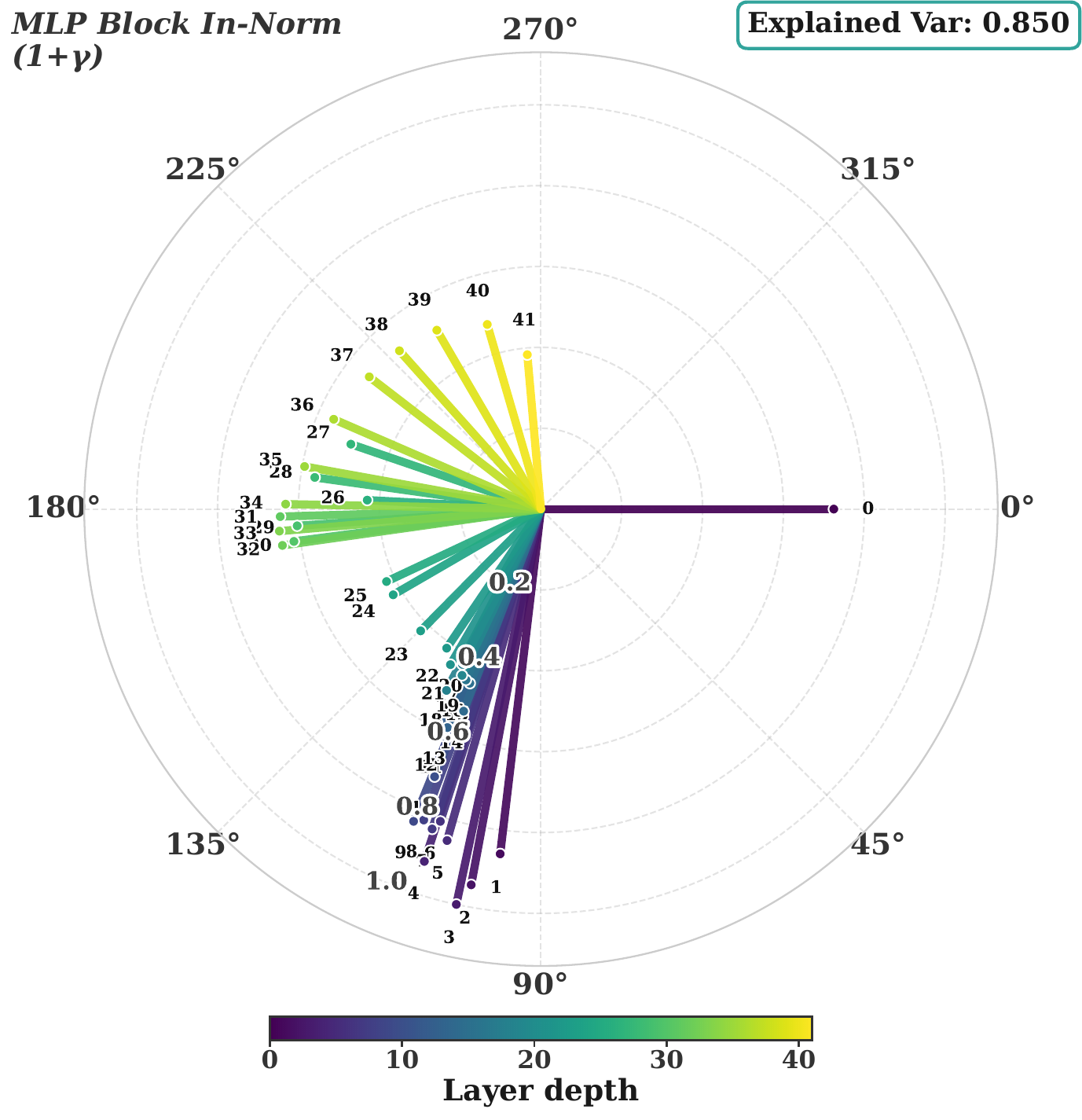}\\[2pt]
        \includegraphics[width=\linewidth]{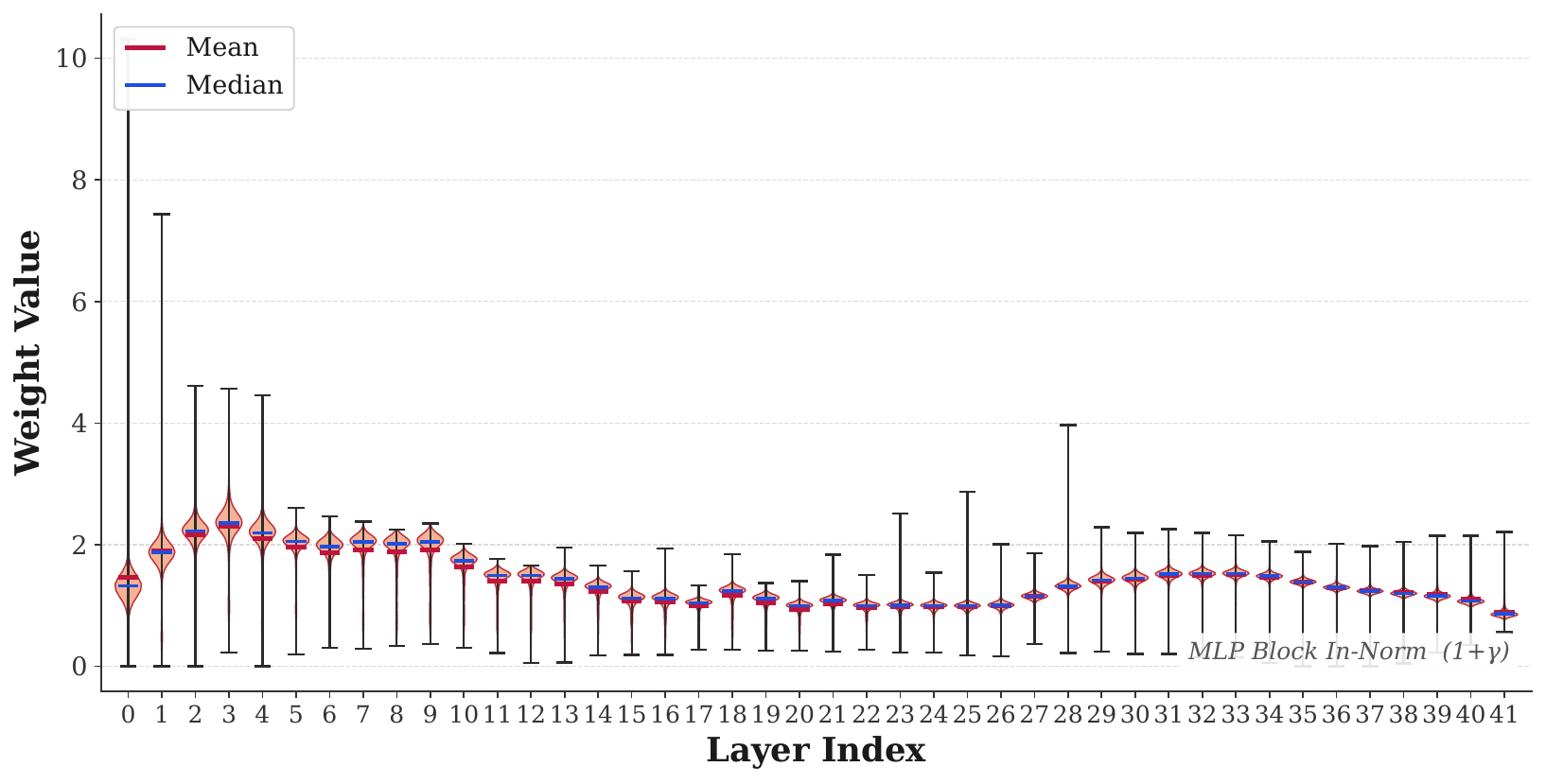}
        \caption{Gemma~2~9B: MLP input norm}
    \end{subfigure}\hspace{0.04\textwidth}%
    \begin{subfigure}[b]{0.43\textwidth}
        \centering
        \includegraphics[width=\linewidth]{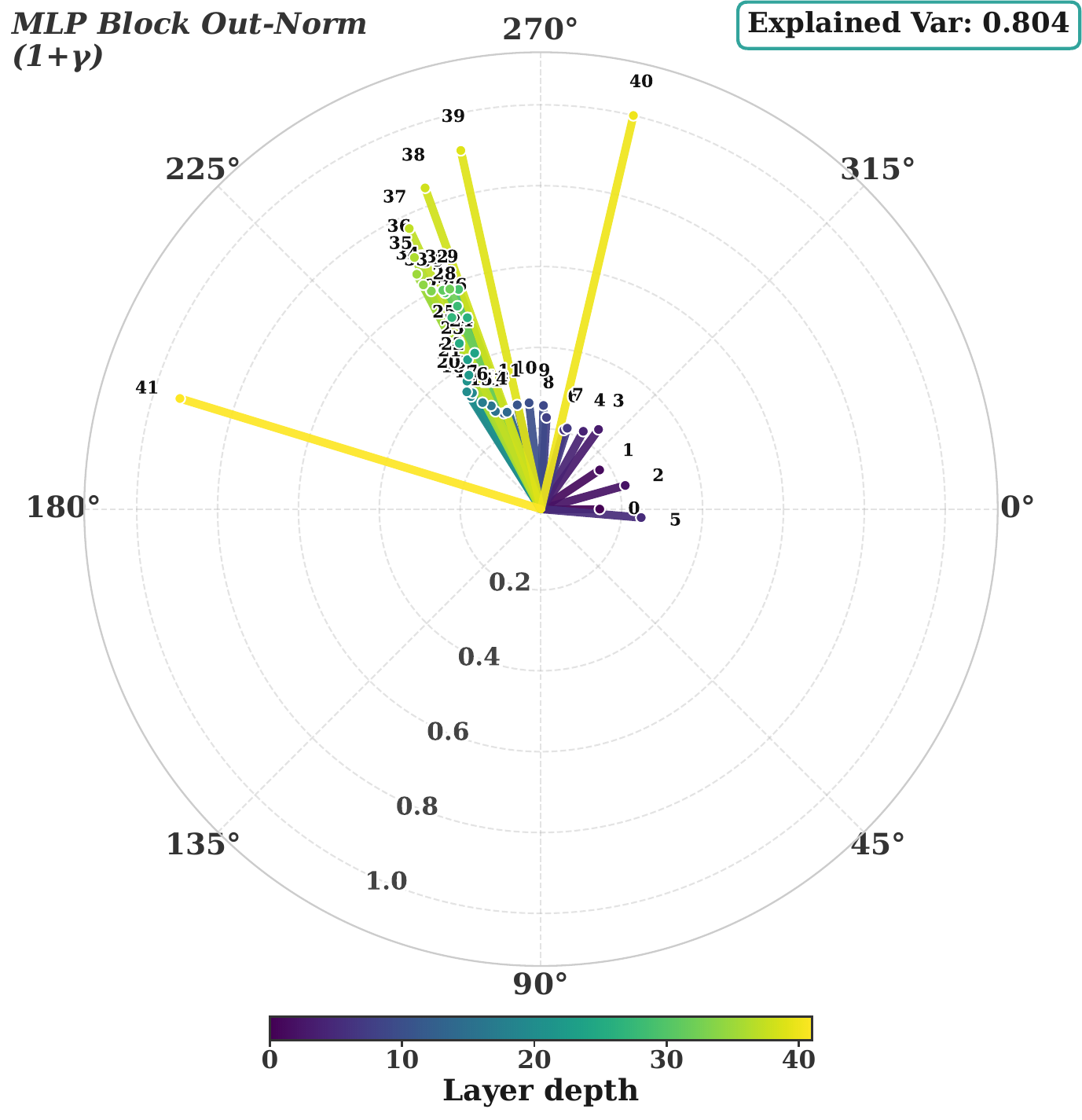}\\[2pt]
        \includegraphics[width=\linewidth]{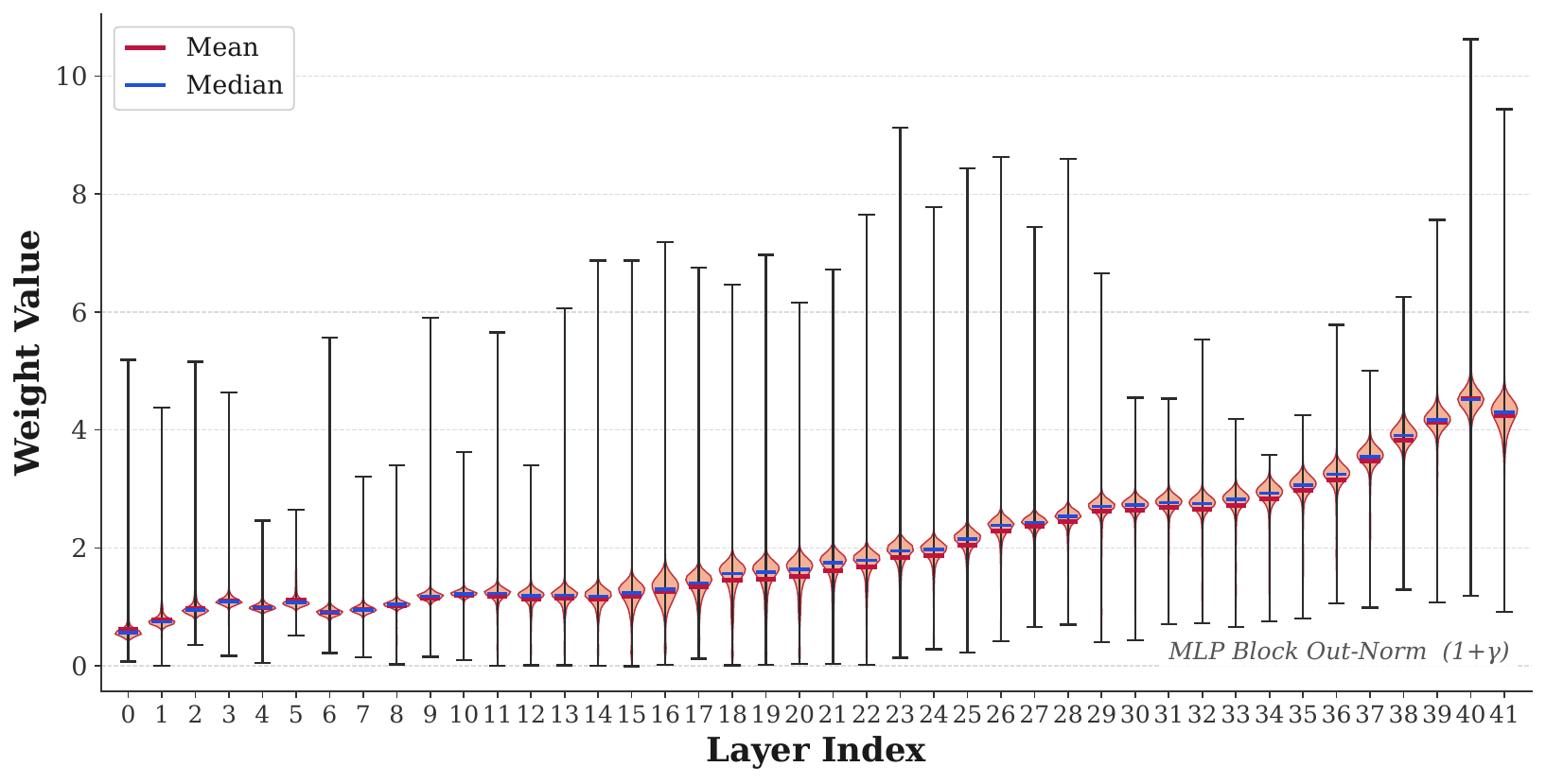}
        \caption{Gemma~2~9B: MLP output norm}
    \end{subfigure}

    \medskip
    \begin{subfigure}[b]{0.43\textwidth}
        \centering
        \includegraphics[width=\linewidth]{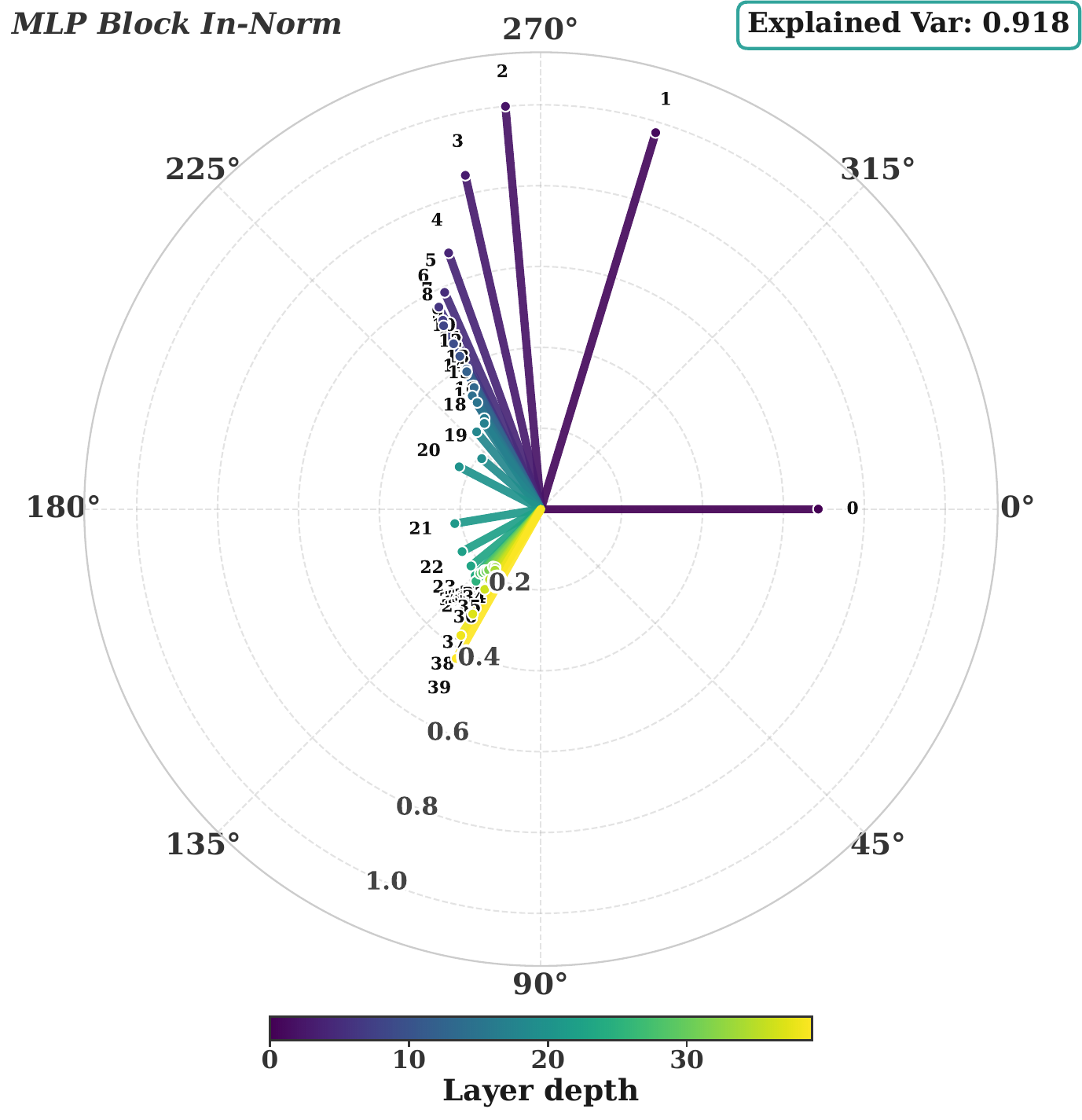}\\[2pt]
        \includegraphics[width=\linewidth]{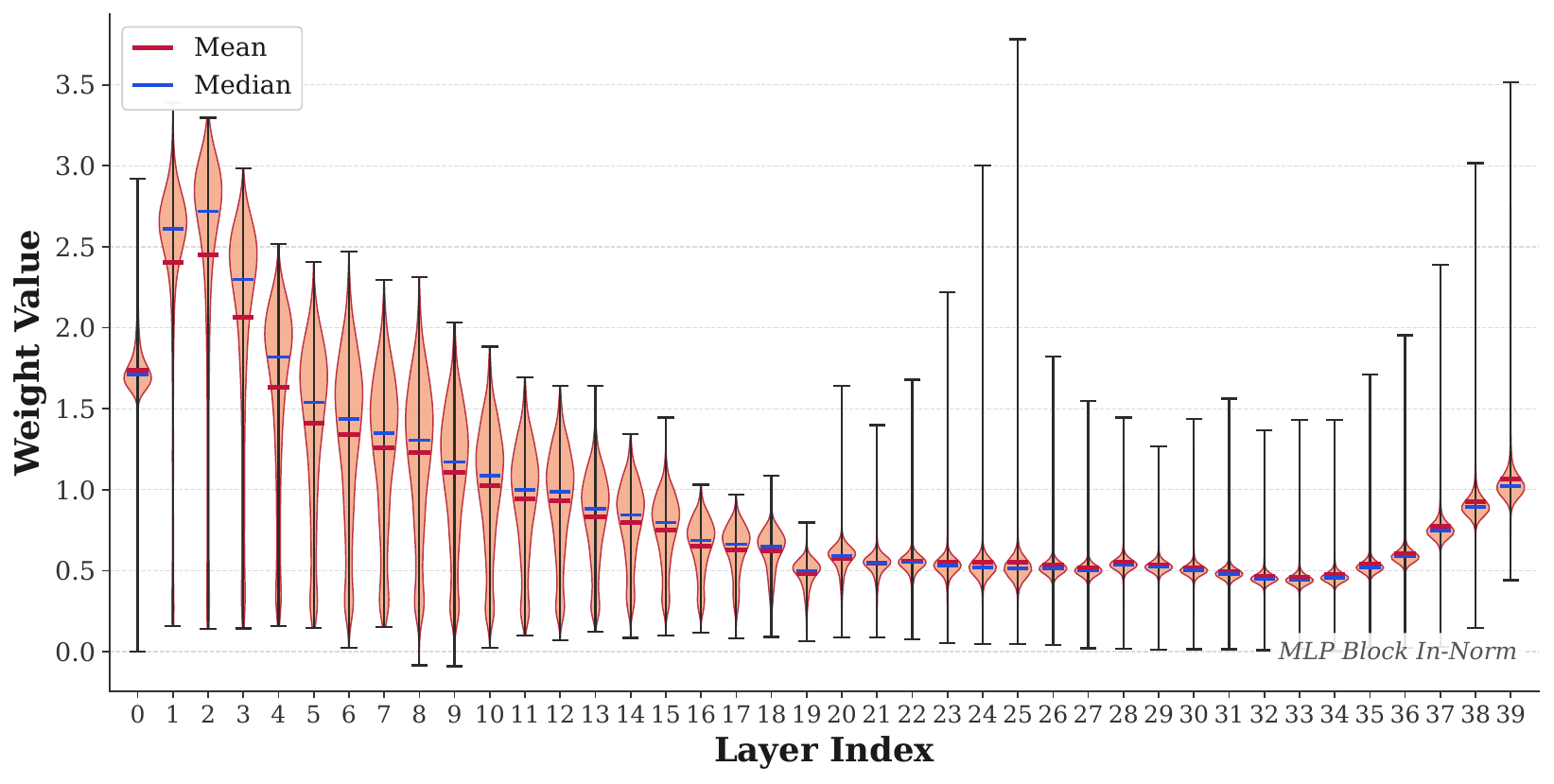}
        \caption{GLM-4~9B: MLP input norm}
    \end{subfigure}\hspace{0.04\textwidth}%
    \begin{subfigure}[b]{0.43\textwidth}
        \centering
        \includegraphics[width=\linewidth]{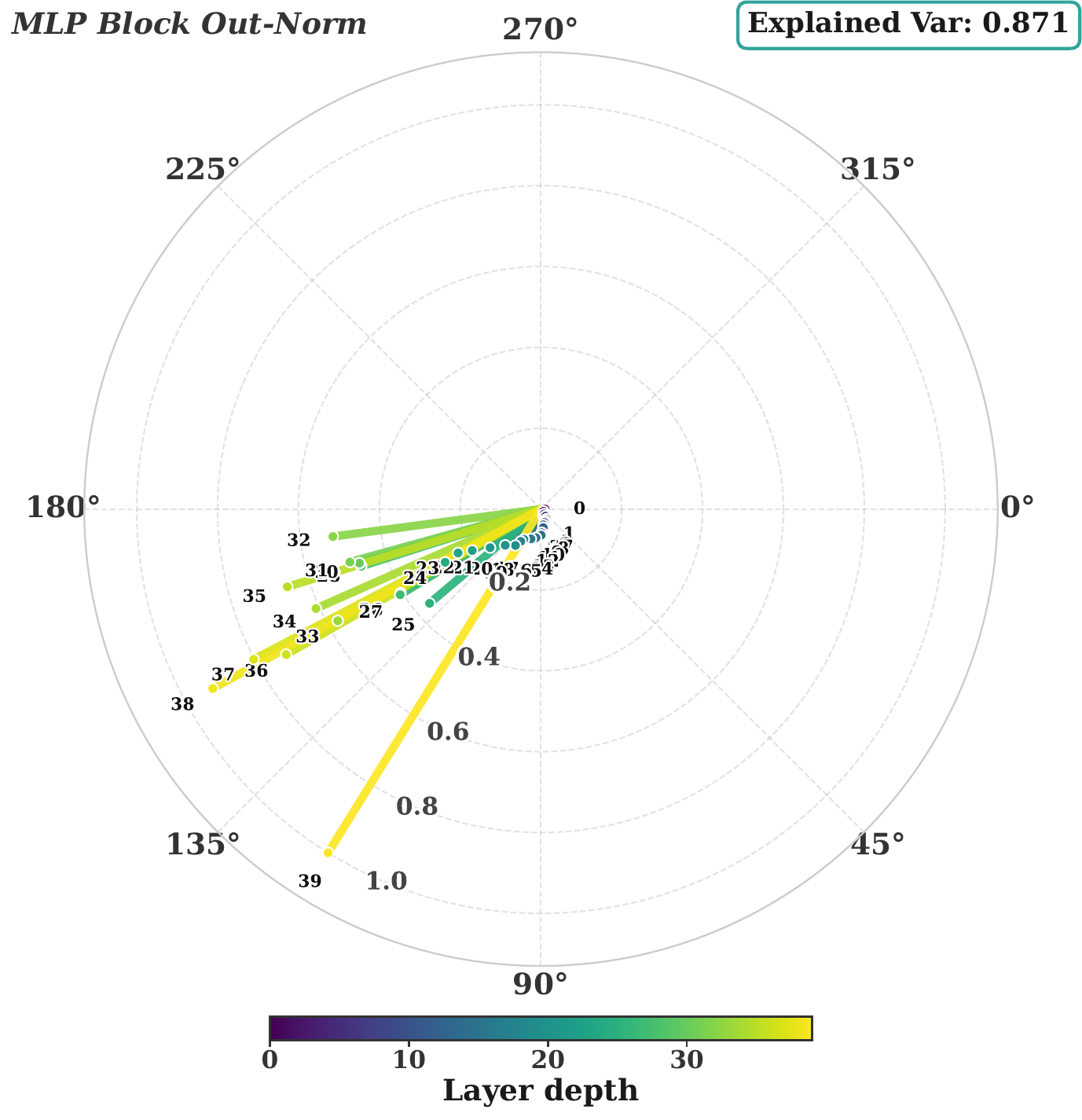}\\[2pt]
        \includegraphics[width=\linewidth]{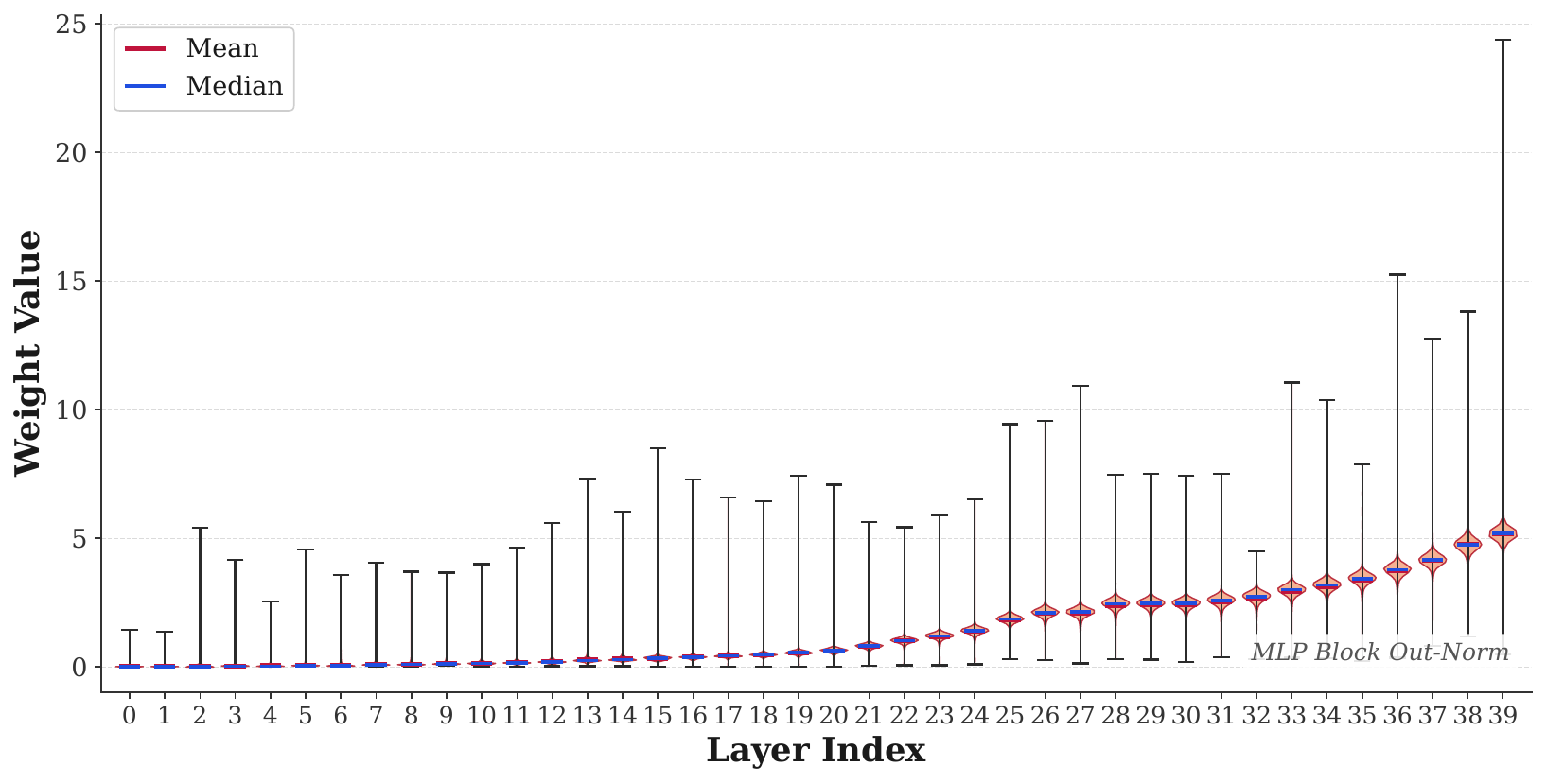}
        \caption{GLM-4~9B: MLP output norm}
    \end{subfigure}
    \caption{\textbf{\textit{(Peri-Norm, MLP Blocks)} Depth-conditioning of Norm Weights.} For each pre-trained model (\textbf{Gemma~2~9B and GLM-4~9B}), the input and output norms of the MLP block are shown: (a, b) Gemma~2~9B and (c, d) GLM-4~9B. In each panel, the MDS projection of the per-layer $\gamma$ vectors (angle) sits above the distribution of $\gamma$ values (magnitude); Gemma~2's gains are shown as the applied $1+\gamma$. We omit the Attention block norms, where as expected, the depth-conditioning is weak}
    \label{fig:app_peri}
\end{figure}

\begin{figure}[p]
    \centering
    \begin{subfigure}[b]{0.44\textwidth}
        \centering
        \includegraphics[width=\linewidth]{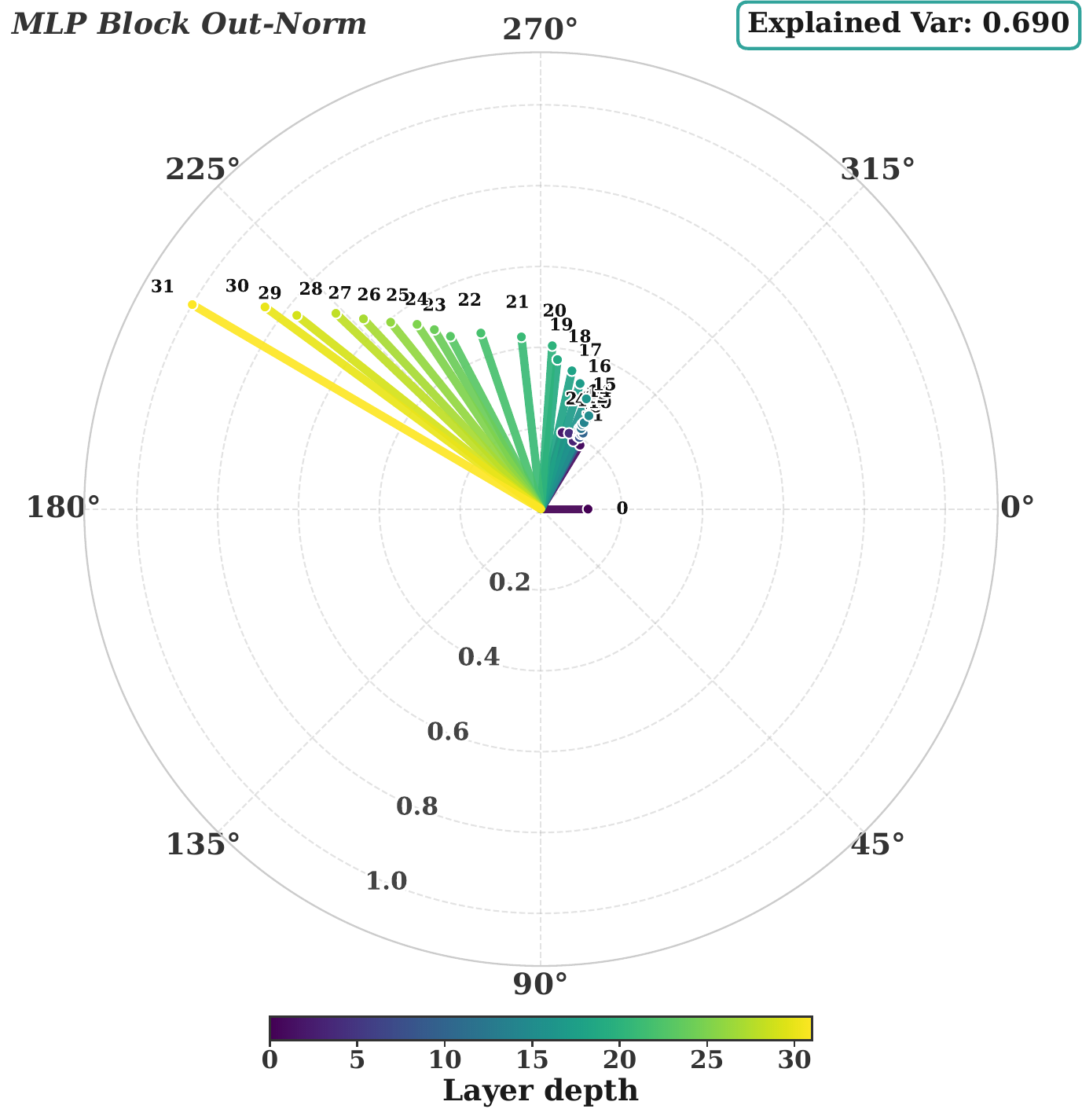}\\[2pt]
        \includegraphics[width=\linewidth]{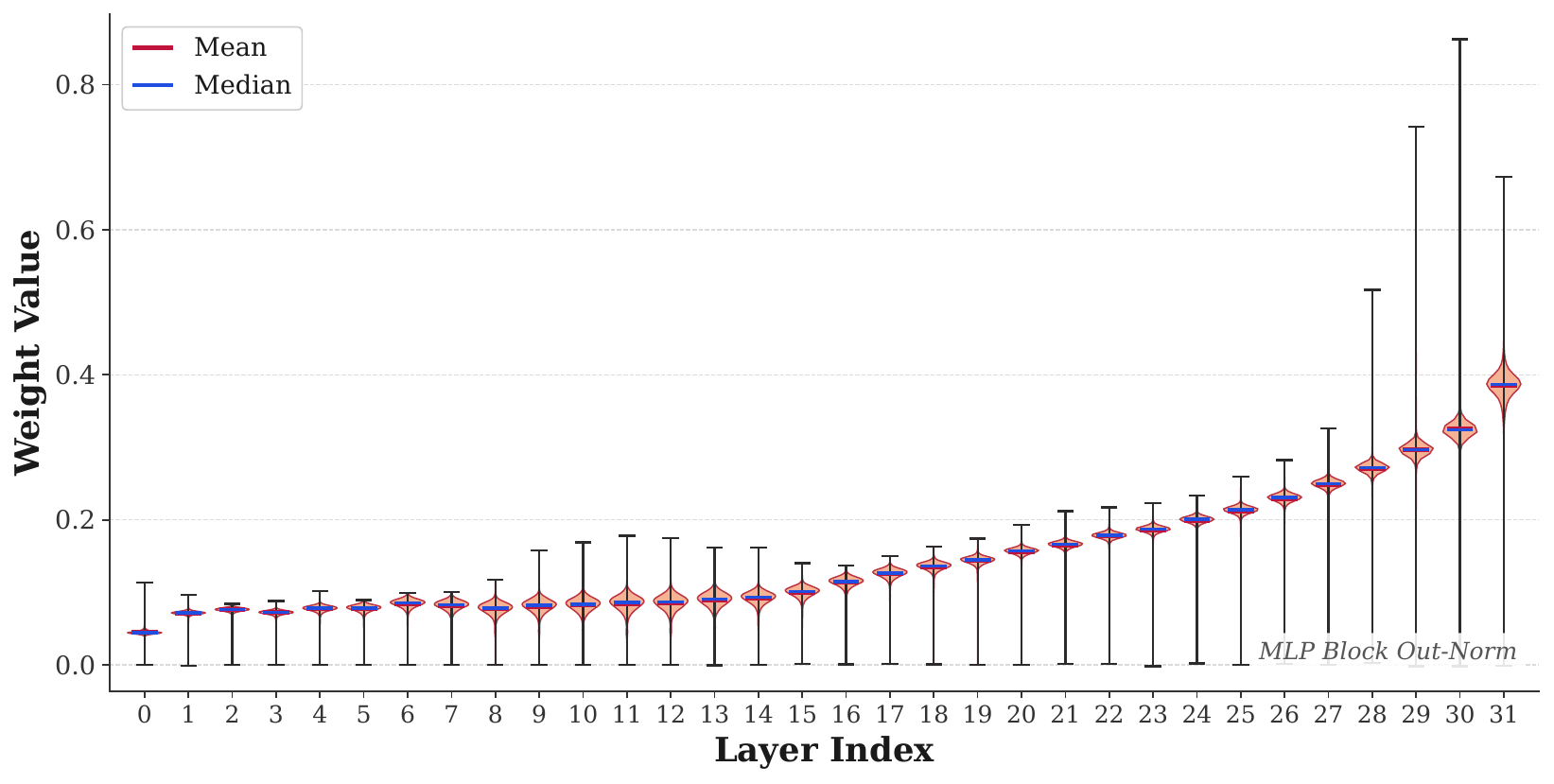}
        \caption{OLMo~2~7B: MLP output norm}
    \end{subfigure}\hspace{0.04\textwidth}%
    \begin{subfigure}[b]{0.44\textwidth}
        \centering
        \includegraphics[width=\linewidth]{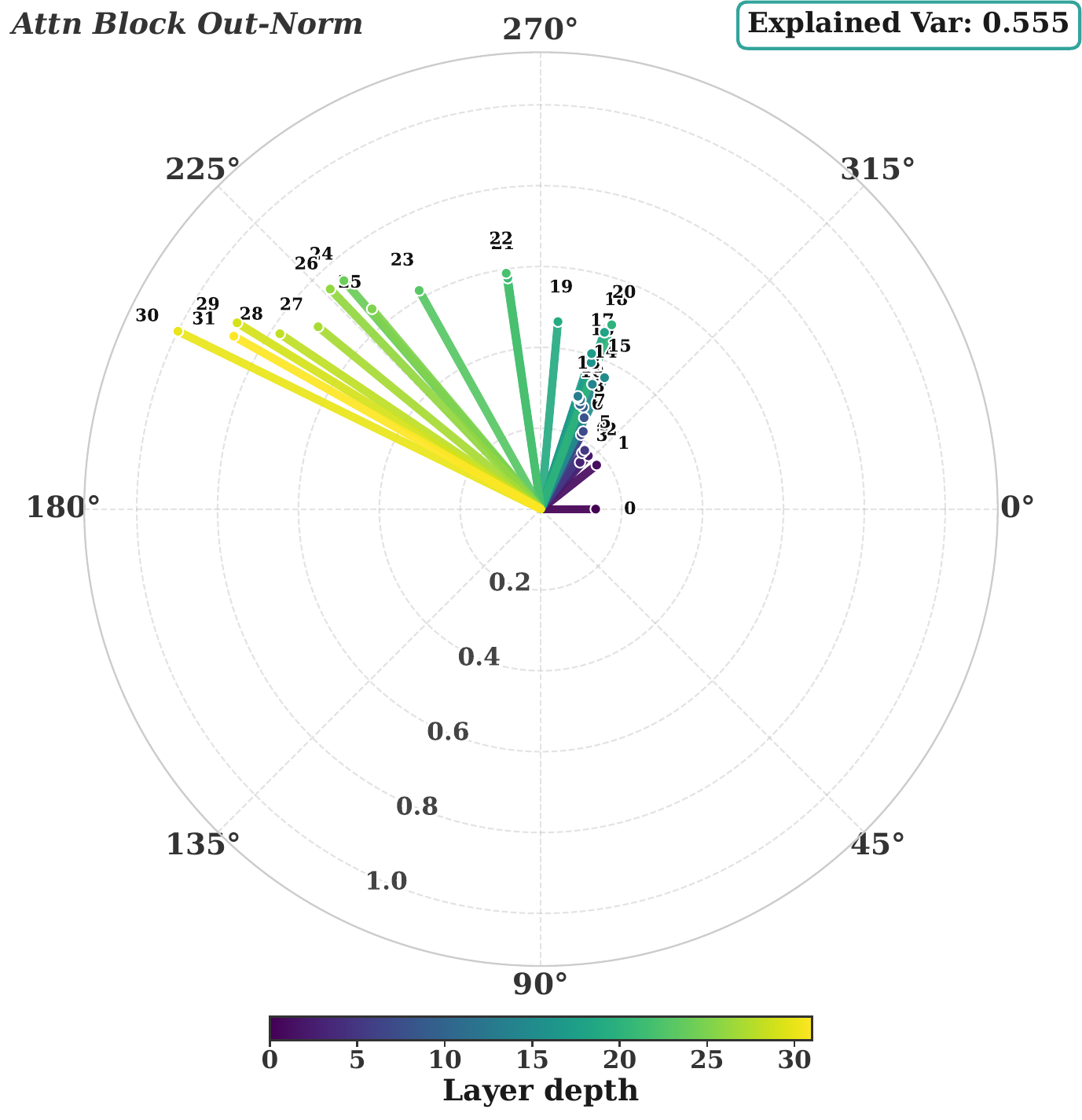}\\[2pt]
        \includegraphics[width=\linewidth]{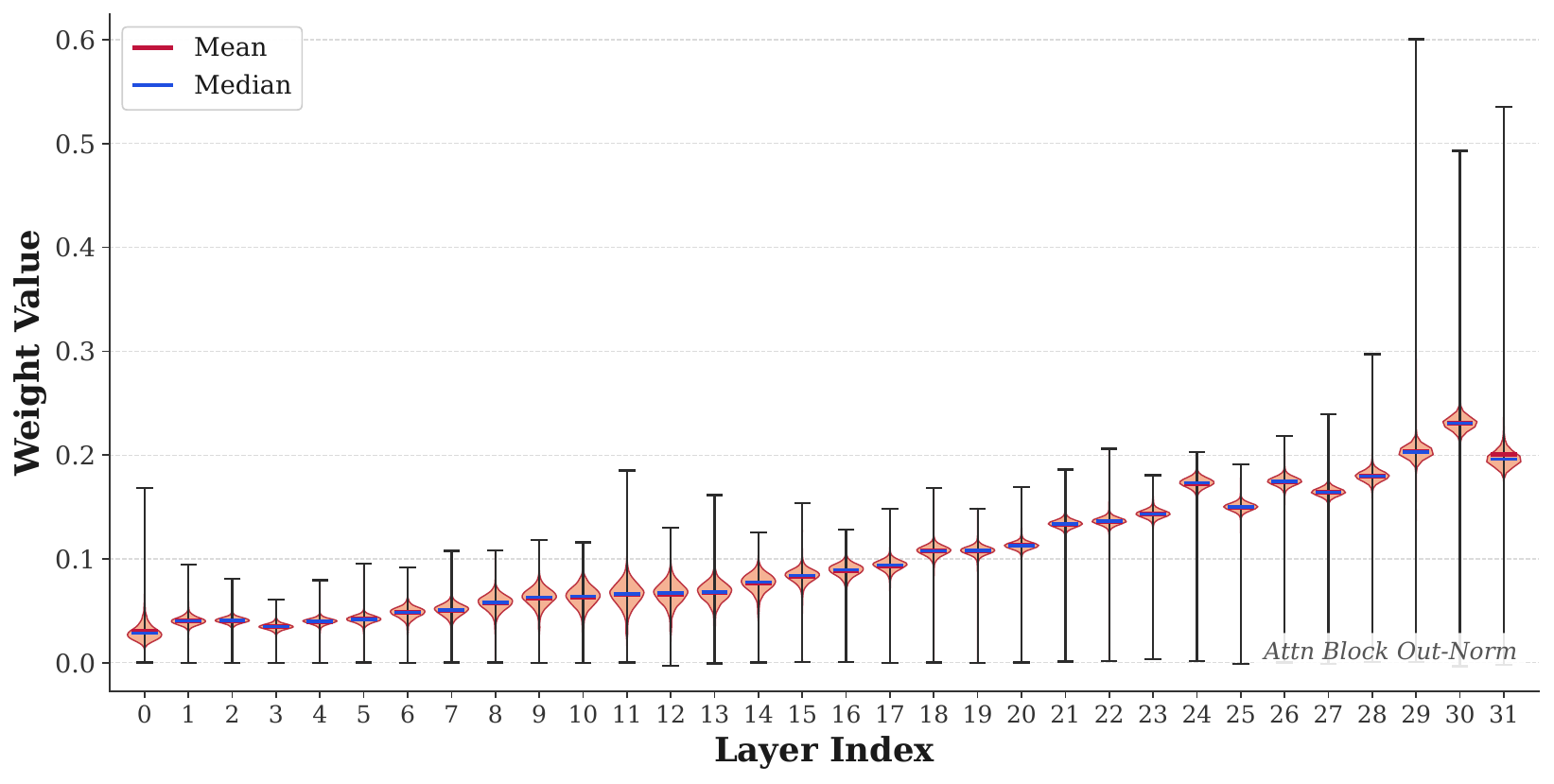}
        \caption{OLMo~2~7B: Attention output norm}
    \end{subfigure}

    \medskip
    \begin{subfigure}[b]{0.44\textwidth}
        \centering
        \includegraphics[width=\linewidth]{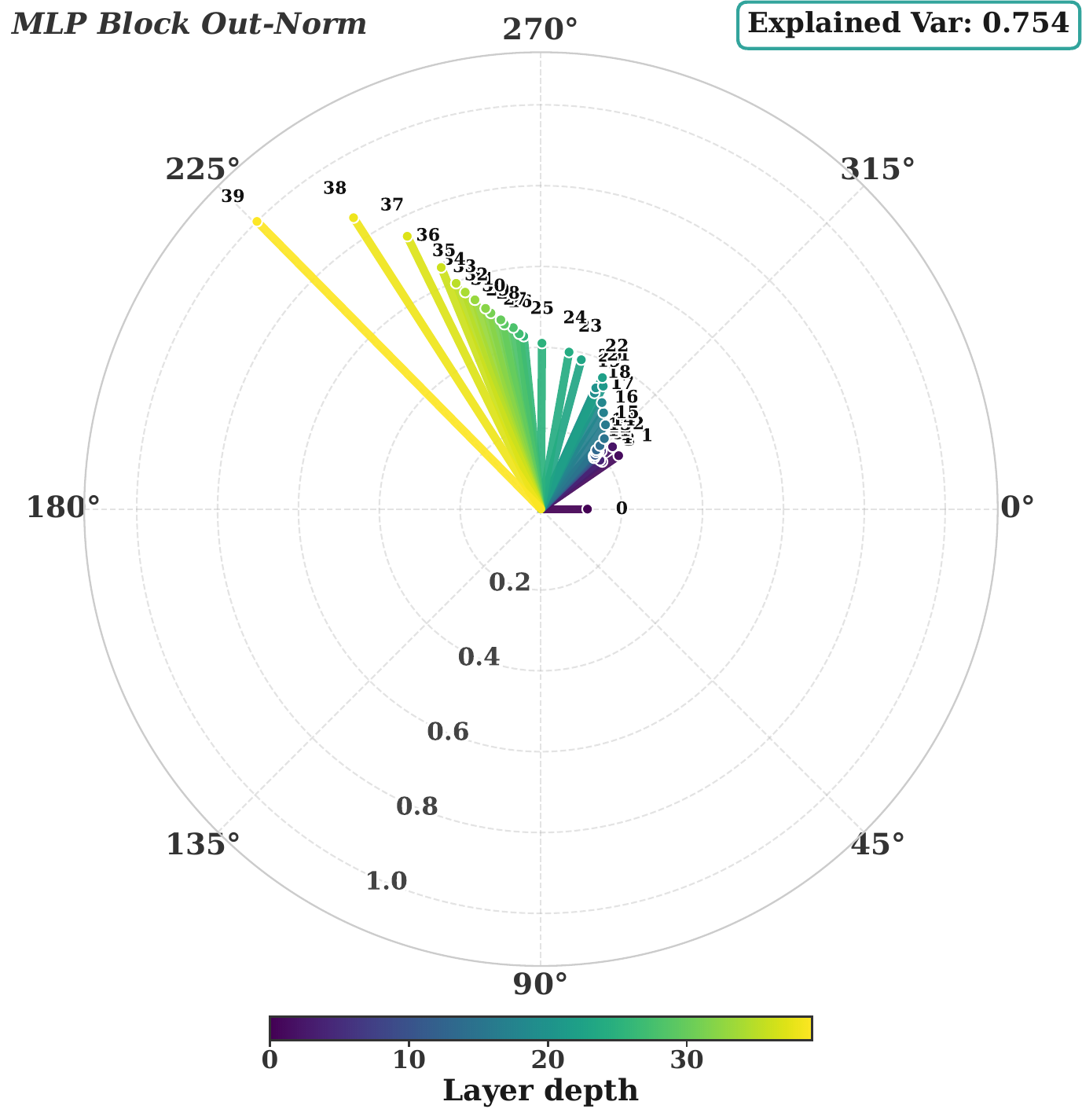}\\[2pt]
        \includegraphics[width=\linewidth]{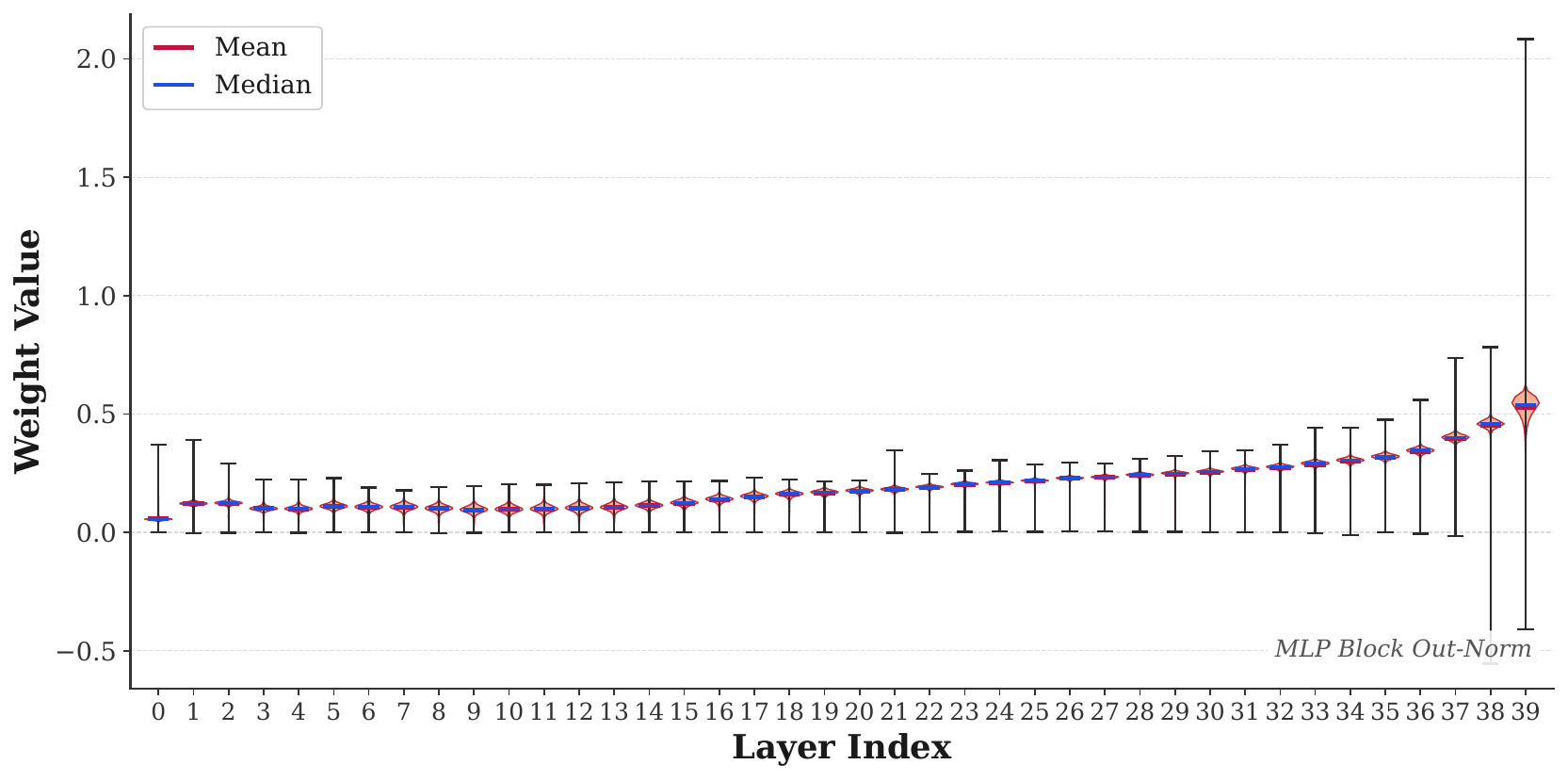}
        \caption{OLMo~2~13B: MLP output norm}
    \end{subfigure}\hspace{0.04\textwidth}%
    \begin{subfigure}[b]{0.44\textwidth}
        \centering
        \includegraphics[width=\linewidth]{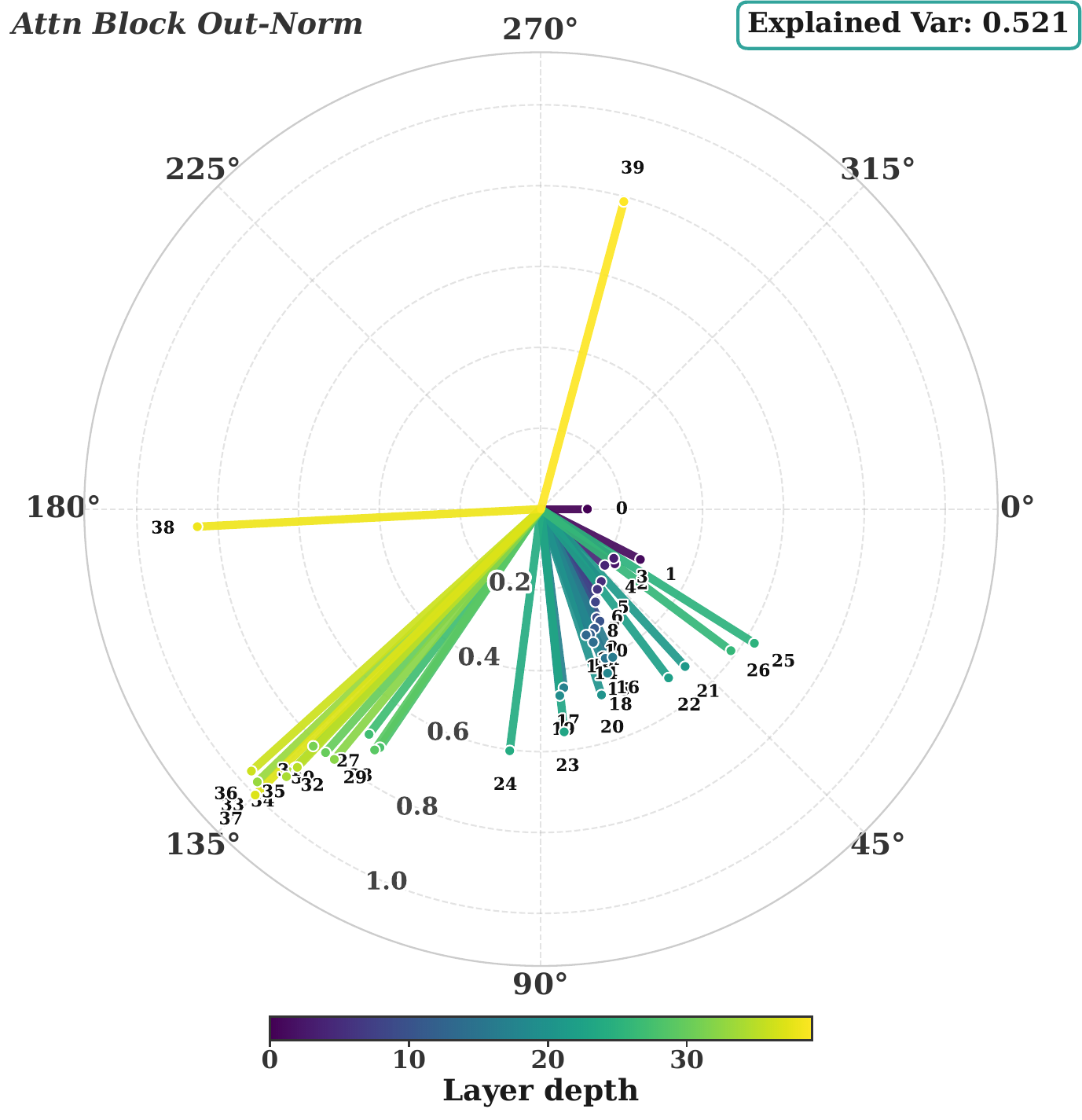}\\[2pt]
        \includegraphics[width=\linewidth]{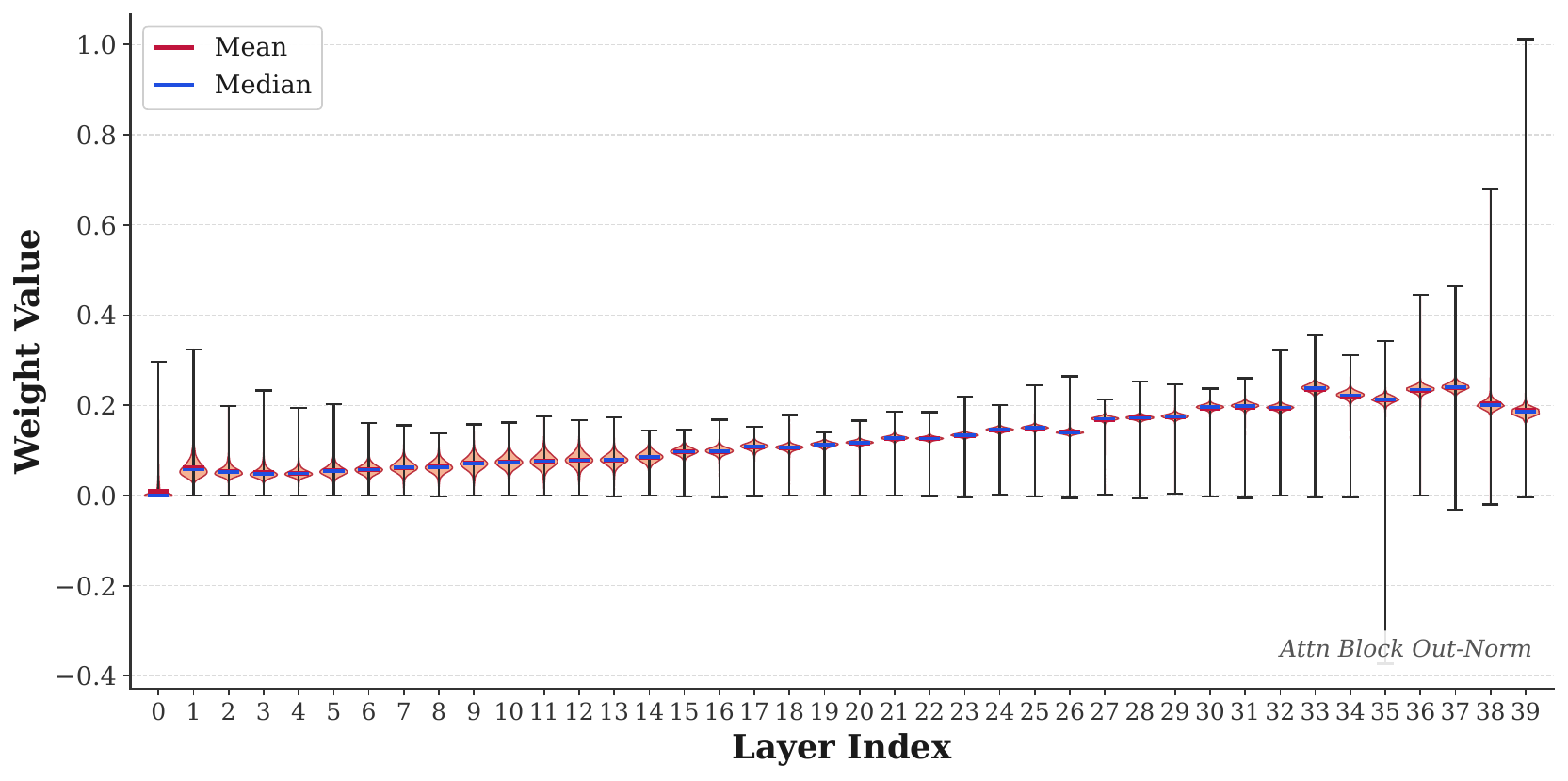}
        \caption{OLMo~2~13B: Attention output norm}
    \end{subfigure}
    \caption{\textbf{\textit{(Post-Norm, MLP and Attention Blocks)} Depth-conditioning of Norm Weights.} For each pre-trained model (\textbf{OLMo~2~7B and OLMo~2~13B}), the MLP and Attention block norms are shown. OLMo~2 normalizes only the output of each block, so each model has an MLP and an Attention output norm. Within each panel, the MDS projection (angle) sits above the distribution of $\gamma$ values (magnitude).}
    \label{fig:app_post_olmo2}
\end{figure}

We further study \textbf{Peri-Norm} LLMs, which normalize both the input and the output of every block (Gemma~2~9B and GLM-4~9B~\citep{glm2024chatglm}), and \textbf{Post-Norm} LLMs, which normalize only the block output (OLMo~2~7B and 13B~\citep{olmo20242}).\footnote{OLMo~2 applies its norm inside the residual branch, $x + \mathrm{Norm}(f(x))$, rather than after the residual addition, $\mathrm{Norm}(x + f(x))$, as in the original Post-Norm Transformer~\citep{vaswani2017attention}.} By and large, Post-Norm LLMs follow the same trends as Pre-Norm LLMs (Figure~\ref{fig:app_post_olmo2}): their MLP block norms carry a clear depth-conditioned code, while their Attention block norms are only weakly or unclearly conditioned on depth. Interestingly, in Peri-Norm LLMs both the input and the output norms of the MLP block are depth-conditioned (Figure~\ref{fig:app_peri}), with the effect stronger on the output norms, whose gains grow near-monotonically with depth.

\noindent\textbf{Summary:} Across 16 pre-trained LLMs from nine model families, spanning Pre-, Peri- and Post-Norm placements and four architecture types (dense, Mixture-of-Experts, parallel-branch and hybrid linear-attention), the MLP block norm weights by and large exhibit strong depth-conditioning in both magnitude and angle, with only the few exceptions noted in Appendix~\ref{app:norm_depth_prenorm_mlp}.

\clearpage

\section{Head-wise Decomposition Analysis of Norm Weights}
\label{app:headwise_decomposition}

We now investigate the underlying structure behind the observed MDS rotation of $\gamma$ vectors in Section~\ref{sec:norm_depth_code}, Figure~\ref{fig:abu_mds_projection}. We do this by decomposing the per-layer $\gamma$ vectors of LLaMA-2 7B head-wise, and then track the per dimension growth of each head, over the layers of the network. We follow the head dimensionality of the attention heads, to split the norm weights $\gamma$ into heads.

\begin{figure}[h]
    \vspace{-5pt}
    \centering
    \begin{subfigure}[b]{\linewidth}
        \centering
        \includegraphics[width=0.8\linewidth]{activation-blowup/images/layerrope/analysis/fig4_head3_norm_growth_llama2_7b.png}
        \caption{Head 3 (Dimensions 384-511 of 4096): Layer-wise growth of Head 3 dimensions of Attention norm weights (left, "Attn Input Norm"), and MLP norm weights (right).}
        \label{fig:abu_heads_head3}
    \end{subfigure}\\[0.4em]
    \begin{subfigure}[b]{\linewidth}
        \centering
        \includegraphics[width=0.8\linewidth]{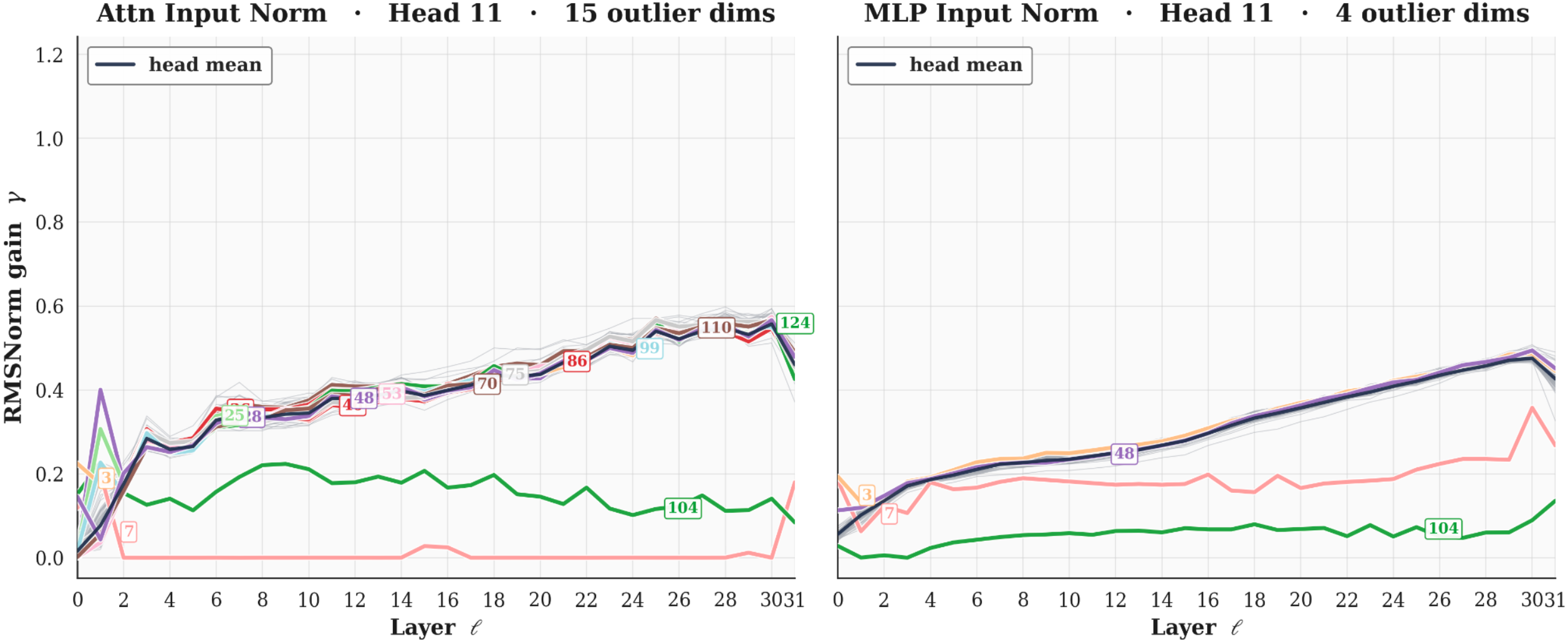}
        \caption{Head 11 (Dimensions 1408-1535 of 4096): Layer-wise growth of Head 11 dimensions of Attention norm weights (left, "Attn Input Norm"), and MLP norm weights (right).}
        \label{fig:abu_heads_head11}
    \end{subfigure}
    \caption{\textbf{Per-dimension RMSNorm gain $\gamma$ across layers} (filtered by specific Heads), for the Attention-input (left panel of each row) and MLP-input (right panel) norms of LLaMA-2 7B.}
    \label{fig:abu_heads}
    \vspace{-7.5pt}
\end{figure}

\noindent We observe that the bulk of dimensions ($\sim\!99\%$) share one approximate trajectory, with little deviation. For every head we examined, almost all dimensions of $\gamma$ follow the same near-monotone curve across layers, well captured by the head-mean (Figure~\ref{fig:abu_heads}). \textit{The bulk of the residual-stream depth code is therefore a single scalar trajectory, repeated across thousands of dimensions.} Interestingly, a small percentage ($\sim\!1\%$) of outliers, break the trend by a large margin, and deviates strongly. We identify two qualitative classes:
\begin{itemize}[leftmargin=*, itemsep=2pt, topsep=2pt]
    \item \emph{\textbf{Anti-symmetric outliers.}} The dimension's trajectory in the Attention-input norm is approximately a mirror image of its trajectory in the MLP-input norm (e.g.\ Head 3 dim 106, Head 11 dim 104). The same scalar dimension is \textit{up-weighted on one branch and down-weighted on the other}, in a layer-dependent way.
    \item \emph{\textbf{Deactivated outliers.}} The dimension's gain is near zero across all layers and both branches (Head 11 dim 7 and dim 104), a consistent null route through the head.
\end{itemize}

\noindent \textbf{Inferences:} This implies that the bulk-mean of the gamma trajectory is essentially a \textit{per-layer scalar magnification of the same trajectory/direction}. The MDS rotation in Section~\ref{sec:norm_depth_code} therefore comes from the outliers: the anti-symmetric dimensions reorient the $\gamma$ vector layer-by-layer, while the deactivated dimensions remain pinned at the origin. This decomposition motivates our LayerRoPE parameterisation: \textit{replace the bulk per-layer $\gamma$ with a single shared vector plus a small set of layer-conditioning scalars, and let an explicit complex factor carry the magnitude and rotation that the network would otherwise distribute across $L$ independent vectors.}

\clearpage

\section{Experimental Details}

\subsection{Model Ladder Scaling}
\label{app:scaling}

We compare methods on a five-model ladder under a common protocol: at every scale, all methods share the data, training recipe and learning-rate tuning procedure, and differ only in normalization. 

\paragraph{Models.}
All models are LLaMA-style decoders (RMSNorm, SwiGLU, RoPE, untied embeddings, head dimension 64), following the setting in ~\citet{sun2025curse}, spanning 58M to 1.34B parameters (Table~\ref{tab:scaling-ladder}).  We compare Pre-Norm, Post-Norm, Peri-Norm, Layer-Norm Scaling and LayerRoPE on the Pre-Norm backbone. LayerRoPE uses one configuration at every scale (depth slopes initialized at $\beta=-0.5$ for magnitude and rotation, rotation base frequency 100); like every baseline, only its learning rate is tuned.

\begin{table}[h]
  \centering
  \small
  \vspace{-5pt}
  \caption{\textbf{Scaling ladder and tuned peak learning rates.} Tokens and $C$ are the total training tokens and compute of each run (Appendix~\ref{app:compute-scaling}). Rates for 60M-500M are measured optima; rates for 1B are single runs at each method's extrapolated optimum (Appendix~\ref{app:lr-sweeps}). $^\dagger$Failed to converge.}
  \label{tab:scaling-ladder}
  \begin{tabular}{lrrrccccc}
    \toprule
     & & & & \multicolumn{5}{c}{Peak learning rate ($\times 10^{-3}$)} \\
    \cmidrule(lr){5-9}
    Model & Params & Tokens & $C$ (FLOPs) & Pre & Post & Peri & \shortstack{LNS} & LayerRoPE \\
    \midrule
    60M  & 58.1M  & 4.6B   & $1.2{\times}10^{18}$ & 16 & 2 & 32 & 16 & 64 \\
    130M & 134.1M & 10.7B  & $7.2{\times}10^{18}$ & 8  & 2 & 8  & 16 & 32 \\
    250M & 250.0M & 20.0B  & $2.7{\times}10^{19}$ & 8  & 1 & 16 & 16 & 64 \\
    500M & 553.8M & 44.3B  & $1.4{\times}10^{20}$ & 4  & 0.0625 & 16 & 8 & 32 \\
    1B   & 1.34B  & 107.1B & $8.3{\times}10^{20}$ & 2  & 0.031$^\dagger$ & 8 & 8 & 32 \\
    \bottomrule
  \end{tabular}
\end{table}

\vspace{-10pt}

\paragraph{Data, budget and training recipe.}
We train on C4 (English) with the LLaMA-2 tokenizer at sequence length 256, padding or truncating each document and masking padding from the loss. This is a single pass over C4 with a batch size of 512, at every scale except 1.3B, which runs for 1.15 epochs to make up the 80 tokens/param. The Adam optimizer is used ($\beta_1{=}0.9$, $\beta_2{=}0.999$, $\epsilon{=}10^{-8}$) \textit{without weight decay or gradient clipping} for fairness. Learning rate with linear warmup over the first 10\% of steps, then cosine decay to $0.1\times$ the peak rate is used. Weights, activations and optimizer states are kept in bf16, except for the normalization gains across all methods, which are kept in fp32 for reliability of the experimentation centered on various normalization formulations. Note, however critically, that in all other experiments for Looped Latent models, ViTs, among others, all parameters are kept in bf16, per the original setting. We report cross-entropy on a fixed 10M-token sample of the C4 validation split. All methods at a scale share one random seed, which fixes both the initialization and data order, so every comparison is paired.

\vspace{-10pt}

\paragraph{The 1B tier.}
We noted that, on training the 1B tier above 100B tokens, since the final normalization gain in our experiments were not weight-decayed, the output-logit scale (across all models) kept growing. At 1B it grows large enough that bf16 rounding of the logits destabilized late training. We therefore continued the Pre-Norm, Peri-Norm, Layer-Norm Scaling and LayerRoPE 1B runs with fp32 output logits from a checkpoint that precedes any instability. Post-Norm nevertheless fails to scale: at 500M every rate above $6.25{\times}10^{-5}$ collapses and the best run plateaus, and at 1B it fails to converge well.

\subsubsection{Variability across seeds.\label{app:multi-seed}}
To measure seed-to-seed variability, we trained every method at 58M and 134M with two further seeds at its tuned rate (Table~\ref{tab:seed-variability}). Standard deviations across seeds are at most 0.016 nats, and LayerRoPE has the lowest loss under every seed at both scales. 

\begin{table}[htb]
  \centering
  \small
  \caption{\textbf{Seed-to-seed variability on the scaling ladder.} Final validation loss of every method at its tuned learning rate (Table~\ref{tab:scaling-ladder}), as mean\,$\pm$\,standard deviation over three seeds. A seed sets both the initialization and the data order and is shared by all methods, so the last column is a paired difference.}
  \label{tab:seed-variability}
  \setlength{\tabcolsep}{4pt}
  \begin{tabular}{lcccccc}
    \toprule
     & \multicolumn{5}{c}{Final validation loss} & Paired gap \\
    \cmidrule(lr){2-6}\cmidrule(lr){7-7}
    Params & Pre & Post & Peri & \shortstack{Layer-Norm\\Scaling} & LayerRoPE & \shortstack{LayerRoPE\\$-$\,Pre} \\
    \midrule
    58M  & 3.172\,{\scriptsize$\pm$\,0.001} & 3.201\,{\scriptsize$\pm$\,0.001} & 3.158\,{\scriptsize$\pm$\,0.014} & 3.147\,{\scriptsize$\pm$\,0.002} & \textbf{3.129}\,{\scriptsize$\pm$\,0.016} & $-0.043$\,{\scriptsize$\pm$\,0.016} \\
    134M & 2.915\,{\scriptsize$\pm$\,0.014} & 2.941\,{\scriptsize$\pm$\,0.003} & 2.884\,{\scriptsize$\pm$\,0.010} & 2.859\,{\scriptsize$\pm$\,0.002} & \textbf{2.831}\,{\scriptsize$\pm$\,0.001} & $-0.085$\,{\scriptsize$\pm$\,0.014} \\
    \bottomrule
  \end{tabular}
\end{table}

\subsubsection{Learning-Rate Sweeps}
\label{app:lr-sweeps}

\begin{figure}[h]
  \centering
  \includegraphics[width=0.6\linewidth]{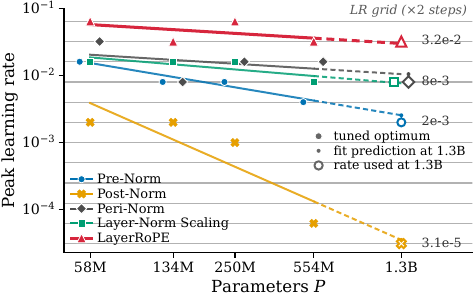}
  \caption{\textbf{Extrapolating the tuned learning rate to 1B.} Filled marks are each method's tuned optima at the four swept scales (Figure~\ref{fig:lr-sweeps}). Lines are the per-method least-squares fits $\log \mathrm{LR}^\star = a_m + b_m \log P$, solid over the fitted range and dashed where extrapolated. At the 1B tier ($P = 1.34$B), each small dot is a fit's prediction and each hollow marker is that prediction rounded to the nearest rate on the $\times2$ grid (horizontal lines); the hollow markers are the rates used for the 1B runs (Table~\ref{tab:scaling-ladder}).}
  \label{fig:lr-extrapolation}
\end{figure}

The peak learning rate is tuned per method and per scale on a shared $\times2$ grid, using the same procedure for every method. At 60M we sweep eight rates ($2.5{\times}10^{-4}$ to $3.2{\times}10^{-2}$). At 130M, 250M and 500M we run brackets of four, three and two rates, centered on the method's previous optimum or on its fitted trend. At every scale, the grid is extended until the optimum is interior. A sweep at 1B was computationally prohibitive, so each method's 1B rate is extrapolated from its four measured optima (Figure~\ref{fig:lr-extrapolation}). We fit $\log \mathrm{LR}^\star = a_m + b_m \log P$ by least squares, evaluate the fit at the 1B model's $P = 1.34$B parameters, and round the prediction to the nearest rate on the $\times2$ grid (rounding moves no prediction by more than 0.39 grid steps), shown in Figure~\ref{fig:lr-extrapolation}. At 1B, each method is trained once, at the rate given by the fit. Table~\ref{tab:scaling-ladder} lists the selected rates.

\subsubsection{Compute Scaling}
\label{app:compute-scaling}
Training compute is $C = D\,F$, where $D$ counts all training tokens and $F$ is the training FLOPs per token from Table~\ref{tab:flop-overhead} (matrix multiplications and attention, backward pass $=2\times$ forward). See Appendix~\ref{app:computational_analysis} for our Analytical FLOP Estimation of LayerRoPE. We fit $L(C) = E + A_m C^{-\alpha_m}$ by nonlinear least squares to the 20 (method, scale) points. $E$ is shared across methods, since all train on the same data, and $(A_m,\alpha_m)$ are per method; Post-Norm is excluded. The fit gives $E = 1.81$ with an RMS residual of 0.005 nats. The $3.4\times$ compute saving is the ratio of Pre-Norm's 1B compute to the compute at which LayerRoPE's fitted curve reaches Pre-Norm's 1B loss. That loss lies between LayerRoPE's measured 500M and 1B losses, so the saving is an interpolation, not an extrapolation. Because every scale uses the same tokens-per-parameter ratio, these curves compare methods at matched budgets rather than along a compute-optimal frontier.

\subsubsection{Scaling to 6.7B Parameters}
\label{app:scaling-7b}
Beyond the ladder, we train every method at 6.7B parameters on a reduced budget. The model follows the LLaMA-7B configuration~\citep{touvron2023llama}, with 32 layers of width 4096, 32 attention heads and a feed-forward width of 11008. A learning-rate sweep is infeasible at this scale, so all methods share a peak rate of $5{\times}10^{-4}$, not tuned for LayerRoPE, and use the ladder's batch size, sequence length, optimizer and precision settings. The learning rate warms up linearly over 15K steps and then decays along a cosine schedule set for 150K steps; all runs end at step 89K (11.7B tokens).

\begin{table}[htb]
  \centering
  \scriptsize
  \caption{\textbf{Scaling to 6.7B parameters.} Final C4 validation loss and perplexity, and zero-shot accuracy on the tasks of Table~\ref{tab:downstream_scaling_layerrope}, after 11.7B training tokens at a shared learning rate. \textbf{Best} and \underline{second-best} are marked per column. $^{\dagger}$Lambada perplexity above $10^{4}$.}
  \label{tab:scaling-7b}
  \setlength{\tabcolsep}{0pt}
  \begin{tabular*}{\textwidth}{@{\extracolsep{\fill}}l*{12}{c}@{}}
    \toprule
     & \multicolumn{2}{c}{\textbf{C4 validation}} & \multicolumn{8}{c}{\textbf{Accuracy} (\%, $\uparrow$)} & & \\
    \cmidrule(lr){2-3}\cmidrule(lr){4-11}
    \textbf{Method} & \textbf{Loss} ($\downarrow$) & \textbf{PPL} ($\downarrow$) & \textbf{ARC-e} & \textbf{HellaS.} & \textbf{PIQA} & \textbf{BoolQ} & \textbf{OBQA} & \textbf{Wino.} & \textbf{LMB.} & \textbf{MMLU} & \textbf{Avg} ($\uparrow$) & \textbf{LMB PPL} ($\downarrow$) \\
    \midrule
    Pre-Norm & 2.545 & 12.74 & 42.7 & 40.5 & 67.8 & \textbf{60.3} & \textbf{31.0} & 51.5 & 29.4 & \underline{23.0} & 46.2 & 37.6 \\
    Post-Norm & 7.172 & 1{,}302 & 26.2 & 26.5 & 49.0 & 37.8 & 24.4 & 48.1 & \phantom{0}0.0 & 22.9 & 30.3 & -$^{\dagger}$ \\
    Peri-Norm & 2.514 & 12.35 & 42.3 & 41.7 & 67.9 & \underline{58.3} & 30.8 & 50.4 & 32.2 & \underline{23.0} & 46.2 & 30.8 \\
    Layer-Norm Scaling & \underline{2.464} & \underline{11.75} & \textbf{45.4} & \underline{45.6} & \textbf{70.9} & 49.9 & 30.8 & \underline{52.6} & \underline{34.2} & \textbf{23.0} & \underline{47.0} & \underline{28.1} \\
    \textbf{LayerRoPE} & \textbf{2.377} & \textbf{10.77} & \underline{44.0} & \textbf{49.6} & \underline{70.5} & 56.5 & \textbf{31.0} & \textbf{56.0} & \textbf{37.6} & 22.9 & \textbf{49.3} & \textbf{19.0} \\
    \bottomrule
  \end{tabular*}
\end{table}

LayerRoPE trains stably and outperforms every baseline (Table~\ref{tab:scaling-7b}). It matches Pre-Norm's final loss after only 41K of the 89K steps, and finishes 0.168 nats below Pre-Norm and 0.087 nats below the strongest baseline, Layer-Norm Scaling. It also attains the best average zero-shot accuracy and the lowest Lambada perplexity, while Post-Norm fails to converge.

\clearpage

\subsection{Model Depth Scaling}
\label{app:depth-scaling}

\paragraph{Training Configuration.}
Similar to the model ladder scaling setup, the depth study uses LLaMA-style decoders, with SwiGLU, RoPE and untied embeddings over the 32K-token LLaMA-2 vocabulary.  
For computational feasibility over 512 layers, we fix the model width to 128 (with SwiGLU feed-forward blocks of width 352 and head size of 64), as we scale depth, from 17.8M parameters at 48 layers to 111.1M parameters at 512 layers. Every model is trained on C4 at sequence length 256 for 20K steps of 512 sequences (2.6B tokens), which is at worst 1x Chinchilla. We follow the optimizer and training recipes from Appendix~\ref{app:scaling}. 

\vspace{-10pt}

\paragraph{LayerRoPE and Baseline Configuration.} LayerRoPE is applied to the Pre-Norm backbone, and using one configuration at every depth. DeepNet~\citep{wang2024deepnet} is applied by scaling the residual stream by a constant $\alpha$ before each post-block normalization and initializes the value, output and feed-forward projections with gain $(8L)^{-1/4}$. The fixed-rate study (Figure~\ref{fig:abu_depth_scaling}) uses $\alpha = (2 \times 32)^{1/4}$, the value DeepNet prescribes for 32 layers, at every depth. For completeness, in the below learning-rate sweep, we use the depth-matched $\alpha = (2L)^{1/4}$. We note that in fact, that at the shared rate of $10^{-3}$, the fixed-$\alpha$ variant reaches the lower loss at every depth, so this choice does not disadvantage DeepNet. The fixed-rate study reports two seeds per method at every depth up to 384 layers and a single seed at 512 layers.

\vspace{-10pt}

\begin{figure}
    \vspace{-30pt}
  \centering
  \includegraphics[width=\linewidth]{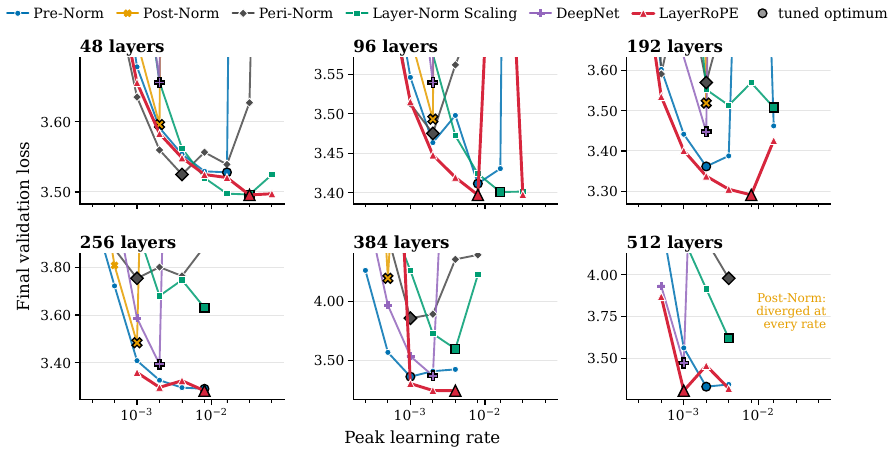}
  \vspace{-10pt}
  \caption{\textbf{Learning-rate sweeps at every Depth.} Final validation loss against peak learning rate for the six methods of Figure~\ref{fig:abu_depth_scaling} at each depth. Outlined markers indicate each method's best rate; loss axes are set per panel.}
  \vspace{-10pt}
  \label{fig:lr-by-depth}
\end{figure}

\begin{figure}
  \centering
  \includegraphics[width=0.59\linewidth]{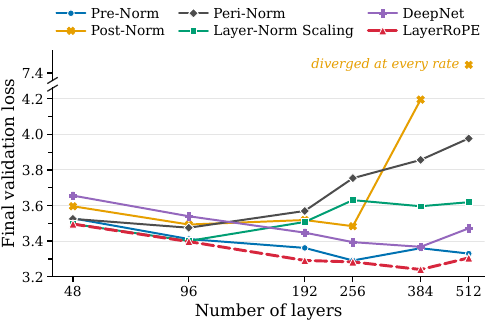}
  \caption{\textbf{Depth scaling at the tuned learning rate.} Final validation loss against the number of layers, with every method at its best rate from Figure~\ref{fig:lr-by-depth} (seed 1). At 512 layers, Post-Norm diverged at every rate (upper strip).}
  \label{fig:depth-best-lr}
\end{figure}

\paragraph{Learning-rate sweep.}
To test whether the fixed-rate results reflect the shared rate rather than the methods, we sweep the peak rate at every depth for all six methods with one seed (Figure~\ref{fig:lr-by-depth}). Figure~\ref{fig:depth-best-lr} shows each method's best loss over the sweep. The sweep confirms our prior observations: LayerRoPE attains the lowest loss at every depth. The baselines' failures persist despite their own best tuned learning rates: Peri-Norm and Layer-Norm Scaling degrade beyond 96 layers, and Post-Norm, which trains at its best rate up to 256 layers, degrades sharply at 384 layers and diverges at every rate at 512. This strongly supports our observation of LayerRoPE's depth-scaling benefits.

\subsection{Learning-Rate Robustness Sweeps}
\label{app:lr-dense}

\paragraph{Setup.}
The sweeps of Section~\ref{sec:lr-robustness} compare Pre-Norm and Peri-Norm, each with and without LayerRoPE, in the configuration of the scaling ladder (Appendix~\ref{app:scaling}), including LayerRoPE's ($\beta=-0.5$, base frequency 100). The 58M runs are the ladder's own (seed 1). The 368M model has 24 layers of width 1024, and the 1.3B model is the ladder's. Both follow the ladder's recipe with shorter budgets: 60K steps with 6K warmup steps at 368M, and a 100K-step schedule with 1K warmup steps at 1.3B. The rate grid doubles from $2.5{\times}10^{-4}$ at 58M; at 368M it doubles from $6.25{\times}10^{-5}$ to $8{\times}10^{-3}$ and continues with $2{\times}10^{-2}$, $5{\times}10^{-2}$ and $10^{-1}$; and at 1.3B it covers rates from $1.25{\times}10^{-4}$ to $8{\times}10^{-3}$.  

\paragraph{Metrics.}
Table~\ref{tab:lr-robustness} summarizes each sweep. Following~\citet{wortsman2024small}, we measure LR sensitivity as the average of each arm's excess loss over the rates swept for both arms, capping diverged runs at the loss of a uniform prediction. We term a run failed if its final loss exceeds the best loss of its arm by more than one nat; above the optimum, every such run diverged or collapsed. We report the same in Table~\ref{tab:lr-robustness}.

\begin{table}[htb]
  \centering
  \small
  \caption{\textbf{Learning-rate robustness in the dense sweeps} (Figure~\ref{fig:lr-robustness}), without ($-$) and with ($+$) LayerRoPE: the tuned rate $\eta^\star$ and loss $L^\star$, the largest rate that trained stably, and LR sensitivity $S$~\citep{wortsman2024small}, the mean of $\min(L(\eta), L_0) - \min_{\eta} L(\eta)$ over the rates swept for both arms, with $L_0 = \ln 32000$. Rates are in units of $10^{-3}$. $^\ddagger$Largest rate swept.}
  \label{tab:lr-robustness}
  \setlength{\tabcolsep}{5pt}
  \begin{tabular}{llrrrrrrrr}
    \toprule
     & & \multicolumn{2}{c}{$\eta^\star$} & \multicolumn{2}{c}{$L^\star$} & \multicolumn{2}{c}{Largest stable rate} & \multicolumn{2}{c}{LR sensitivity $S$} \\
    \cmidrule(lr){3-4}\cmidrule(lr){5-6}\cmidrule(lr){7-8}\cmidrule(lr){9-10}
    Backbone & Params & $-$ & $+$ & $-$ & $+$ & $-$ & $+$ & $-$ & $+$ \\
    \midrule
    Pre-Norm  & 58M  & 16  & 64 & 3.172 & 3.118 & 16  & 128$^{\ddagger}$      & 0.458 & 0.119 \\
              & 368M & 4   & 20 & 2.759 & 2.660 & 8   & 50                    & 1.069 & 0.360 \\
              & 1.3B & 0.5 & 1  & 2.648 & 2.505 & 0.5 & 8$^{\ddagger}$        & 1.162 & 0.112 \\
    \midrule
    Peri-Norm & 58M  & 32  & 64 & 3.152 & 3.120 & 32  & 128$^{\ddagger}$      & 0.440 & 0.087 \\
              & 368M & 8   & 8  & 2.702 & 2.671 & 8   & 100$^{\ddagger}$      & 1.310 & 0.263 \\
              & 1.3B & 0.5 & 1  & 2.584 & 2.511 & 1   & 8$^{\ddagger}$        & 0.506 & 0.127 \\
    \bottomrule
  \end{tabular}
\end{table}

\paragraph{Results.} We can see from Table~\ref{tab:lr-robustness}, that adding LayerRoPE ($+$), greatly reduces LR Sensitivity across model scales, and for both Pre- and Peri-Norm, by large margins. This we take to imply a flattening of the loss basin near the optima, and the tolerance to divergence on deviation from the optima as well.

\subsection{Looped Latent Models}
\label{app:looped}

\paragraph{Setup.}
We follow the official Parcae recipe~\citep{prairie2026parcae} at every scale. At 140M, 370M and 770M parameters, the prelude, recurrent block and coda each have 2, 4 and 6 Transformer layers, of width 768, 1024 and 1280, respectively.  The GPT baselines are non-looped Transformers of the same width with 6, 12 and 18 layers, identical to Parcae, except the recurrent block iterations. All models here are trained on FineWeb-Edu~\citep{penedo2024fineweb} with Parcae's 32K-token BPE tokenizer, 2{,}048-token sequences and batches of 256 sequences, for 11.2B, 29.6B and 61.6B tokens, using Muon and AdamW with a trapezoidal learning-rate schedule, gradient clipping at 1.0 and bf16 mixed precision. During training, authors in \citet{{prairie2026parcae}} sample the number of loop iterations per sequence. All evaluations are at $T{=}8$ loop iterations, unless stated otherwise. In Parcae~$+$~LayerRoPE, LayerRoPE naively generates the RMSNorm gains of every layer from shared gains through its depth-conditioned magnitude and rotation (Section~\ref{sec:abu_layerrope}). As mentioned in main text, we naively apply LayerRoPE by using each layer's depth index is its position in the network (independent of the looped iteration). LayerRoPE's magnitude depth slopes are initialized at $0$, with rotation base frequency 100. No part of the Parcae recipe is changed or tuned. Downstream, we evaluate on the large tasks (at least 1{,}000 examples), on which every model performs above chance: ARC-Easy, HellaSwag (zero- and 10-shot), PIQA, Lambada, SQuAD, CoQA and BIG-bench QA WikiData, averaging over three few-shot sampling seeds. The full results are provided here in Table~\ref{tab:parcae_layerrope_filtered_pertask}.

\vspace{-10pt}

\begin{wrapfigure}{r}{0.55\columnwidth}
  \centering
  \vspace{-20pt}
  \includegraphics[
    width=\linewidth,
    trim=0 0 0 0,
    clip
  ]{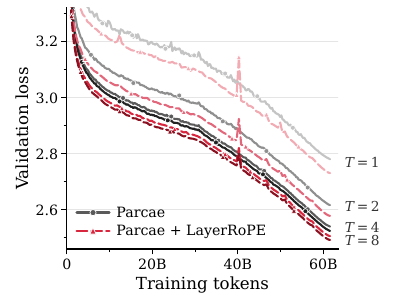}
  \vspace{-25pt}
  \caption{
    \textbf{Every loop depth at 770M.}
    Validation loss throughout training for Parcae (solid) and
    Parcae~$+$~LayerRoPE ({\color{red}dashed}), evaluated at
    $T \in \{1,2,4,8\}$ loop iterations, from lighter to darker.
    The loss axis omits the first ${\sim}1$B tokens.
  }
  \label{fig:loss_770m_tsweep}
  \vspace{-30pt}
\end{wrapfigure}

\paragraph{Loop iterations.}
Figure~\ref{fig:loss_770m_tsweep} evaluates both 770M models at every number of loop iterations. LayerRoPE improves Parcae's convergence at every $T$, albeit with diminishing benefit as loops are added.

\vspace{-10pt}
\begin{figure}[htbp]
  \centering
  \includegraphics[width=\linewidth]{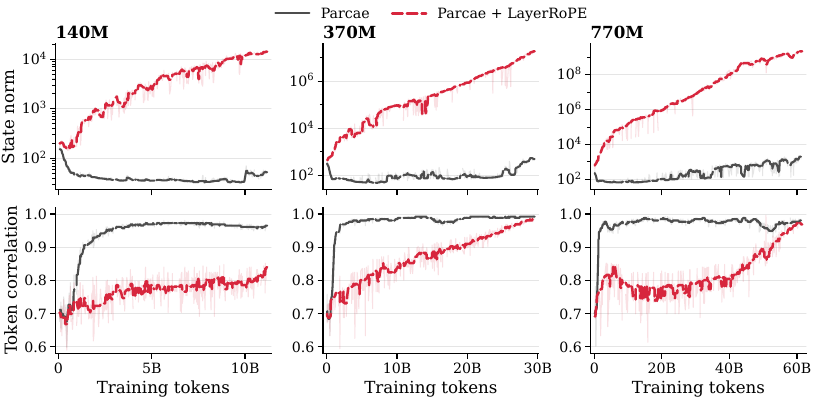}
  \caption{\textbf{Recurrent state during training.} Mean norm of the recurrent state leaving the recurrent block (top, log scale) and mean pairwise cosine similarity between its centred token states (bottom), for Parcae and Parcae~$+$~LayerRoPE. Lines are rolling medians over logged steps, with raw values shown faintly.}
  \label{fig:parcae-recurrent-state}
\end{figure}

\paragraph{Recurrent state.}
Figure~\ref{fig:parcae-recurrent-state} shows that LayerRoPE lets the recurrent state grow by several orders of magnitude during training, reaching norms of ${\sim}10^{9}$ at 770M against ${\sim}10^{3}$ for Parcae. Yet for most of training its token states stay markedly less correlated than Parcae's, which are nearly collinear (mean pairwise cosine similarity 0.96-0.99). From the markedly less token correlation in the recurrent states, we infer that LayerRoPE encourages better utilization of the recurrent block's capacity during looping. The fact that this occurs while the recurrent state norms are increasing orders of magnitude over vanilla Parcae, is indicative of our finding that controlled norm/variance increase may infact be beneficial to the utilization of depth in Transformers.

\begin{table}[h]
  \centering
  \small
  \setlength{\tabcolsep}{4pt}
  \resizebox{\textwidth}{!}{%
  \begin{tabular}{c l c cc cccccccc c}
    \toprule
    & & & & & \multicolumn{8}{c}{\textbf{Accuracy} (\%, $\uparrow$)} & \\
    \cmidrule(lr){6-13}
    & \textbf{Model} & \shortstack[c]{\textbf{Looped}\\\textbf{Iters.\ (T)}} & \shortstack[c]{\textbf{Val.}\\\textbf{PPL} ($\downarrow$)} & \shortstack[c]{\textbf{Lambada}\\\textbf{PPL} ($\downarrow$)} & \textbf{ARC-e} & \textbf{HellaS.-0} & \textbf{HellaS.-10} & \textbf{PIQA} & \textbf{LMB.} & \textbf{SQuAD} & \textbf{CoQA} & \textbf{QA-WD} & \textbf{Avg.} ($\uparrow$) \\
    \midrule
    \multirow{3}{*}{\rotatebox[origin=c]{90}{\shortstack[c]{\textbf{140M}\\[2pt]{\scriptsize 11.2\,B}}}} & GPT & - & 20.28 & 105.61 & 51.0 & 31.6 & 31.7 & 62.2 & 24.3 & 5.0 & 9.8 & 23.3 & 29.9 \\
     & Parcae & 8 & 18.66 & 81.17 & 53.6 & 33.7 & 33.7 & 63.1 & 26.4 & 7.5 & 12.8 & 33.1 & 33.0 \\
     & \textbf{Parcae + LayerRoPE} & 8 & \textbf{18.36} & \textbf{67.32} & \textbf{54.0} & \textbf{34.3} & \textbf{34.5} & \textbf{63.9} & \textbf{27.1} & \textbf{14.0} & \textbf{15.6} & \textbf{35.7} & \textbf{34.9} \\
    \midrule
    \multirow{3}{*}{\rotatebox[origin=c]{90}{\shortstack[c]{\textbf{370M}\\[2pt]{\scriptsize 29.6\,B}}}} & GPT & - & 15.01 & 37.16 & 59.5 & 40.3 & 40.1 & 67.0 & 32.9 & 20.2 & 20.1 & 44.4 & 40.6 \\
     & Parcae & 8 & 14.22 & 32.46 & \textbf{61.4} & 43.6 & 43.5 & 68.5 & 35.0 & 28.0 & 20.6 & 46.8 & 43.4 \\
     & \textbf{Parcae + LayerRoPE} & 8 & \textbf{14.05} & \textbf{27.98} & 61.2 & \textbf{43.8} & \textbf{44.3} & \textbf{68.8} & \textbf{37.2} & \textbf{28.5} & \textbf{21.0} & \textbf{47.4} & \textbf{44.0} \\
    \midrule
    \multirow{3}{*}{\rotatebox[origin=c]{90}{\shortstack[c]{\textbf{770M}\\[2pt]{\scriptsize 61.6\,B}}}} & GPT & - & 12.80 & 20.80 & 64.7 & 47.8 & 47.6 & 69.4 & 40.5 & 32.1 & 24.0 & 50.3 & 47.0 \\
     & Parcae & 8 & 12.49 & 18.56 & \textbf{67.4} & 51.2 & 51.9 & \textbf{71.2} & 42.2 & 38.0 & 26.0 & 52.6 & 50.1 \\
     & \textbf{Parcae + LayerRoPE} & 8 & \textbf{12.09} & \textbf{18.44} & 65.7 & \textbf{51.8} & \textbf{52.9} & \textbf{71.2} & \textbf{42.9} & \textbf{40.9} & \textbf{26.2} & \textbf{53.1} & \textbf{50.6} \\
    \bottomrule
  \end{tabular}
  }
  \caption{\textbf{Per-task downstream results of the looped latent models.} Validation perplexity on FineWeb-Edu, Lambada perplexity and accuracy (\%) on each downstream task, averaged over three few-shot sampling seeds, with looped models at $T{=}8$; training tokens under each size, best per size in \textbf{bold}.}
  \label{tab:parcae_layerrope_filtered_pertask}
\end{table}

\section{Learned Depth-Conditioning in LayerRoPE}
\label{app:depth-conditioning}\label{sec:activation-growth}

\paragraph{Setup.}
We analyze the 368M, 1.3B scale Pre- and Per-Norm models of Section~\ref{sec:lr-robustness}, with and without LayerRoPE, as well as a Layer-Norm Scaling model trained identically. We pass 512 samples from the C4 validation set, and measure the variance of the residual stream after each layer's last residual addition (critically, excluding the first (BOS) position whose massive activations may otherwise dominate the early layers). For the 1.3B scale, we also read out LayerRoPE's learned depth code (Section~\ref{sec:abu_layerrope}) at its four sites, the input (read) and residual write (out) of the attention and MLP blocks: the magnitude multiplier $e^{r(\ell)}$ and the rotation angle $\theta(\ell)$, whose slopes $\beta^{\text{mag}}$ and $\beta^{\text{rot}}$ are both initialized at $-0.5$.

\begin{figure}[htbp]
  \centering
  \vspace{-10pt}
  \includegraphics[width=\linewidth]{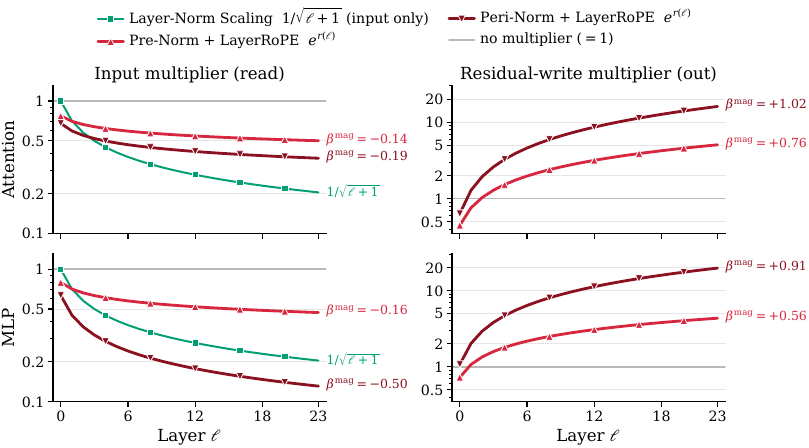}
  \caption{\textbf{Learned depth multipliers at 1.3B.} Learned magnitude multipliers $e^{r(\ell)}$ of LayerRoPE against the fixed $1/\sqrt{\ell+1}$ of Layer-Norm Scaling.}
  \label{fig:depth-multiplier}
\end{figure}

\vspace{-20pt}

\begin{figure}[htbp]
  \centering
  \includegraphics[width=\linewidth]{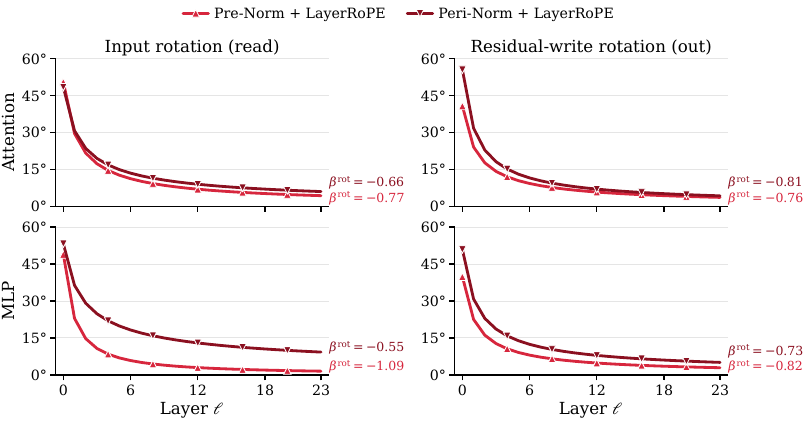}
  \caption{\textbf{Learned rotation angles at 1.3B.} Rotation $\theta(\ell)$ of the highest-frequency channel pair at each LayerRoPE site. Labels give the learned slopes $\beta^{\text{rot}}$.}
  \label{fig:depth-angle}
  \vspace{-20pt}
\end{figure}

\paragraph{Learned multipliers and rotations.}
Figure~\ref{fig:depth-multiplier} contrasts the magnitude multipliers that LayerRoPE learns at 1.3B with the fixed $1/\sqrt{\ell+1}$ input multiplier of Layer-Norm Scaling, and Figure~\ref{fig:depth-angle} shows its learned rotations. In both backbones we notice that LayerRoPE keeps attenuating what each block reads, but also learns to amplify what it writes. The residual-write slopes, initialized at $-0.5$ like the input ones, turn positive during training, so that the \textbf{\textit{write multiplier grows $6$-$25\times$ from the first layer to the last}}. 

\vspace{-10pt}

\begin{figure}[htbp]
 
  \centering
  \includegraphics[width=\linewidth]{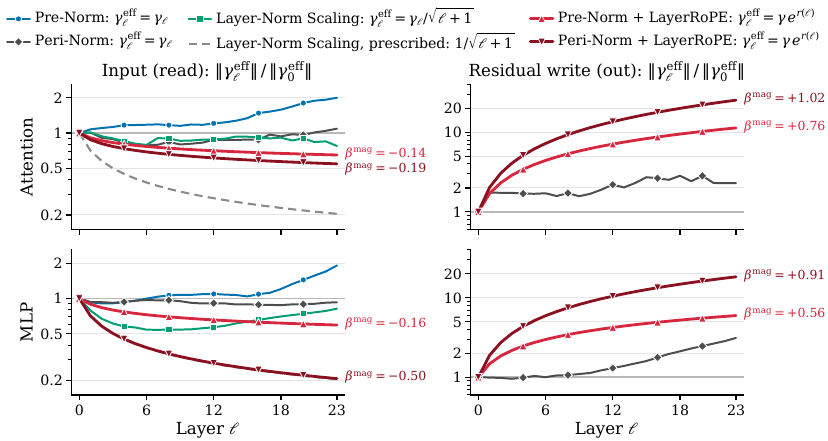}
  \caption{\textbf{Effective depth gain at 1.3B.} Magnitude ratio $\|\gamma^{\text{eff}}_\ell\|/\|\gamma^{\text{eff}}_0\|$ of each layer's effective normalization weights to the first layer's, with every depth-dependent factor folded into $\gamma^{\text{eff}}_\ell$ (legend). For Peri-Norm, the right column shows its output normalization; Pre-Norm and Layer-Norm Scaling have no residual-write weights. LayerRoPE's weights $\gamma$ are shared across layers, so its ratio is $(\ell+1)^{\beta^{\text{mag}}}$; labels give the learned slopes $\beta^{\text{mag}}$.}
  \label{fig:depth-gain}

\end{figure}

\paragraph{Effective depth gain.}
Per-layer normalization weights can encode depth as well, so Figure~\ref{fig:depth-gain} compares the effective normalization weights $\gamma^{\text{eff}}_\ell$ of every method, with its depth-dependent factors folded in, as the magnitude ratio $\|\gamma^{\text{eff}}_\ell\|/\|\gamma^{\text{eff}}_0\|$ to the first layer's. With free per-layer weights, Pre-Norm grows its input weights to about $2\times$ the first layer's by the last layer, while Peri-Norm keeps its input weights within the same range, and grows those of its output normalization to upto $\sim 3\times$. Strikingly, we observe that the learned weights of Layer-Norm Scaling largely compensate for their prescribed down-scaling! Note that the ratio never falls below $\sim 0.75$ at the attention input, and traces a U-shape at the MLP input. LayerRoPE, whose weights are shared across layers, in contrast to all, instead retains a smooth, monotone ratio of $(\ell+1)^{\beta^{\text{mag}}}$. It is unclear though, to what degree this smoothness is beneficial, and we leave this for future work.

\vspace{-10pt}

\clearpage

\section{Related Work\label{sec:related_work}}

\textbf{Controlling residual growth.}
Hidden-state norms/variance grow with depth is a well known phenomena in Pre-Norm Transformers~\citep{sun2025curse}. Existing work to contain this growth, can be seen to be along two lines: The first \emph{attenuates} what enters or exits the residual stream: Layer-Norm Scaling~\citep{sun2025curse} scales each block's input (the output of its normalization) by $1/\sqrt{\ell+1}$, Peri-Norm~\citep{kim2025peri} re-normalizes each block's output, while DeepNet~\citep{wang2024deepnet} and Depth-$\mu$P~\citep{yang2024tensor} bound each block's update to train at extreme depth. Others like ReZero~\citep{bachlechner2021rezero} and LayerScale~\citep{touvron2021going} gate each residual branch with learned multipliers initialized near zero. The Second line of work \emph{restructures} the residual pathway, by circumventing the single stream problem entirely: Attention Residuals~\citep{team2026attention} for instance, replaces the residual with a learned and expensive bounded aggregation of streams over depth, and Hyper-Connections~\citep{zhu2025hyper,xie2025mhc} widen the single stream into several parallel streams, where constrained mixing can preserve signal magnitude. Concurrently, some works like \citet{wang2026negligible}  have shown that the scale vectors of normalization layers, though negligible in size, provides useful preconditioning of future layers, and \citet{wang2026residual} derive a formulaic depth- and loop-dependent residual scale for looped Transformers.  Both treat these gains as fixed or as free per-layer parameters. LayerRoPE is orthogonal to both lines: it directly tackles the singular additive stream problem, by counterintuiviely widening the stream further, generating per-layer gains that regulate each computational block's I/O using one depth-conditioned shared vector.

\textbf{Outliers and norm weights.}
A complementary line of work focuses on understanding \emph{where} these activations blow up (and aim to identify \emph{why}). Works on massive activations~\citep{sun2024massive}, attention sinks~\citep{xiao2023efficient} and high-norm register tokens~\citep{darcet2024vision}, show that extreme values concentrate in a few tokens and channels. Many have pointed to normalization gains as having a role: \citet{wei2022outlier} suggest they amplify outlier channels. \citet{kovaleva2021bert} show that norm gains carry high-magnitude dimensions at the same position in every layer, and \citet{dong2026dissecting} show they vary in magnitude with depth and attention type. These works isolate individual outliers or exploit cross-layer redundancy after training. We instead study the systemic trajectory of \emph{all} dimensions of $\gamma$ across depth: we show that its magnitude and direction evolve jointly, that ${\sim}99\%$ of each head's dimensions share a single depth-wise trajectory, and that a sparse set of outlier dimensions cause this rotation effect (Section~\ref{sec:norm_depth_code}). Finally, LayerRoPE directly builds this structure into the architecture rather than attenuating it.

\textbf{Positional encodings along sequence and depth.}
Self-attention is permutation-equivariant, so Transformers encode token order explicitly, originally with sinusoidal or learned vectors added to the input~\citep{vaswani2017attention}. Rotary position embeddings~\citep[RoPE;][]{su2024roformer} instead rotate each pair of query and key features by an angle proportional to the token's position, with frequencies on a geometric spectrum, so that attention scores depend on relative position alone. Depth, by contrast, is rarely encoded explicitly: Universal Transformers add a timestep embedding at each recurrent step~\citep{dehghani2018universal}, and looped Transformers benefit from per-loop scales conditioned on a timestep encoding~\citep{xu2024expressive}. We find that standard LLMs learn such a depth code implicitly, in the magnitude and direction of their per-layer gains. LayerRoPE makes it explicit, applying RoPE's rotation, together with a learned magnitude, along depth to a shared gain vector rather than along the sequence to queries and keys.
\clearpage


\end{document}